\documentclass{IEEEtran}
\usepackage[utf8]{inputenc}
\pdfoutput=1

\title{
Comparison of Anomaly Detectors: 
Context Matters
\\ {\large IEEE TRANSACTIONS ON
NEURAL NETWORKS AND LEARNING SYSTEMS \\
Special Issue on Deep Learning for Anomaly Detection}
}
\author{Vít Škvára, Jan Franců, Matěj Zorek, Tomáš Pevný,~\IEEEmembership{Member,~IEEE,} Václav Šmídl,~\IEEEmembership{Member,~IEEE}
\thanks{All authors are with the Department
of Computer Science, Faculty of Electrical Engineering, Czech Technical University, Prague, Czech Republic}}%

\usepackage[numbers]{natbib}
\usepackage{graphicx}
\usepackage{url}
\usepackage{amsmath}
\usepackage{amssymb}
\usepackage{booktabs}
\usepackage{pgfplots}
\usepackage{multirow}
\usepackage{subcaption}
\usetikzlibrary{positioning}

\usepackage{chngcntr}

\pgfplotsset{compat=1.15}
\usepgfplotslibrary{statistics}
\usepgfplotslibrary{groupplots}

\renewcommand{\vec}[1]{\boldsymbol{#1}}

\newcommand*\rot{\rotatebox{90}}

\begin{document}

\maketitle

\begin{abstract}
    Deep generative models are challenging the classical methods in the field of anomaly detection nowadays. Every new method provides evidence of outperforming its predecessors, often with contradictory results. The objective of this comparison is twofold: to compare anomaly detection methods of various paradigms with focus on deep generative models, and identification of sources of variability that can yield different results. The methods were compared on popular tabular and image datasets. We identified the main sources of variability to be experimental conditions: i) the type data set (tabular or image) and the nature of anomalies (statistical or semantic), and ii) strategy of selection of hyperparameters, especially the number of available anomalies in the validation set. Different methods perform the best in different contexts, i.e. combination of experimental conditions together with computational time. This explains the variability of the previous results and highlights the importance of careful specification of the context in the publication of a new method. All our code and results are available for download.
\end{abstract}

\section{Introduction}
\IEEEPARstart{D}{eep} generative models are gaining popularity in anomaly detection since the introduction of the Variational Autoencoder (VAE)~\citep{kingma2013auto}. The number of modifications and extensions of VAE or generative adversarial networks (GAN)~\citep{goodfellow2014generative} is sharply increasing, each claiming superiority over the prior art. This raises a suspicion that some of the methods are overspecialized or poorly tested. This work, inspired by the paper "Do we need hundreds of classifiers to solve real-world classification problems?"~\citep{fernandez2014we}, strives to compare under "fair" conditions anomaly detectors to observe how the field has evolved in the last twenty years (the oldest compared detector~\citep{ramaswamy2000efficient} was published in 2000). Specifically, it investigates if methods based on \textit{deep} generative models offer a benefit over methods based on alternative paradigms, either the \textit{classical} methods based on distances, or deep architectures without the capability of generating samples.

Surely there already exist comparisons of anomaly detectors. Earlier surveys~\citep{pimentel2014review, campos2016evaluation, goldstein2016comparative, pevny2016loda} do not compare to deep generative methods, because they were not developed or sufficiently popular at that time. Contrary to that, the study in~\citep{kiran2018overview} contains a detailed description of deep models, but provides experiments only with the basic VAE and only on specialized video datasets. Ref.~\citep{chalapathy2019deep} introduces a taxonomy of deep anomaly detection models but does not compare them experimentally. Other recent surveys~\citep{moustafa2019holistic, kwon2019survey, fernandes2019comprehensive, wang2019progress, pang2020deep} either ignore deep generative models altogether or describe them only theoretically, without making any experimental comparison. The most relevant prior art is~\citep{ruff2020unifying}, which tries to theoretically link deep and shallow techniques\footnote{The \textit{shallow} techniques corresponds to those we call \textit{classical}. We prefer the later terminology, as models based on random forests are in their essence deep, although they cannot capture semantic structure --- a touted feature of deep models}. But again, an extensive experimental comparison of different generative models is missing. One would also expect papers introducing new methods to contain such a comparison. Some of them do~\citep{pevny2016loda}, but generally, we have found comparisons limited (e.g. using a small number of datasets or methods) or flawed, which is elaborated below.

How does this paper avoid the aforementioned deficiencies? First, eight classical methods in comparison serve as a baseline, over which we expect the state-of-the-art deep methods should improve upon (latest compared method~\citep{wang2020advae} was published in 2020). Second, the comparison uses a large number of tabular (40) and image (6) datasets popular in the evaluation of deep models. Third, all methods have been given the same conditions, which primarily means optimization of hyperparameters, as~\citep{vskvara2018generative} has shown this to have a significant impact.

The list of contributions contains:
\begin{enumerate}
    \item Experimental comparison of classical and deep anomaly detectors on a large number of datasets.
    \item Identification of the dataset type, hyper-parameter selection strategy, and computational cost as major factors in the selection of the most suitable method.
    \item We publish codes of the evaluation pipeline and compared methods, including automatic download of datasets, splitting them into training, validation, and testing, and calculating the performance metrics.
\end{enumerate}

The paper is organized as follows. First, the anomaly detection contexts that had the greatest influence on the outcome of our experiments are defined. Second, there is a brief theoretical overview of the tested generative deep models and other methods. Sec.~\ref{sec:experimentalsetup} details the datasets,  different approaches to the selection of hyper-parameters, and other design decisions in the experimental setup. Next, the experimental section discusses the experimental results and lessons we have learned. We summarize the paper with a recommendation to practitioners and our suggestions for future work.

\section{Anomaly Detection Contexts}
\label{sec:contexts}
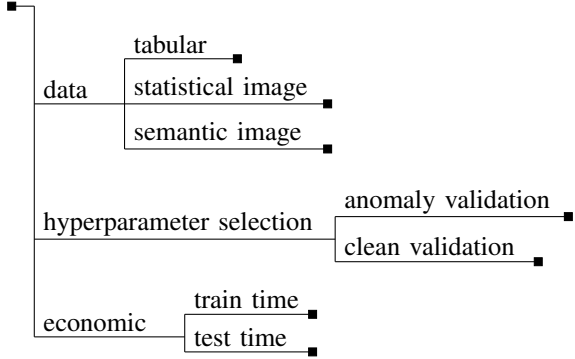
\begin{figure}
    \centering
    \begin{tikzpicture}
          \draw (-0.3,1.3) -- (0,1.3);
          \draw (0,1.3) -- (0,-3.1);
          
          \filldraw ([xshift=-1.5pt,yshift=-1.5pt]-0.3,1.3) rectangle ++(3pt,3pt);

          \draw (0,0) -- (1.2,0);
          \node[anchor=west] at (0,0.2) {data};
          \draw (1.2,-0.6) -- (1.2,0.6);
          \node[anchor=west] at (1.2,0.8) {tabular};
          \filldraw ([xshift=-1.5pt,yshift=-1.5pt]2.7,0.6) rectangle ++(3pt,3pt);
          \draw (1.2,0.6) -- (2.7,0.6);
          \node[anchor=west] at (1.2,0.2) {statistical image};
          \filldraw ([xshift=-1.5pt,yshift=-1.5pt]3.9,0) rectangle ++(3pt,3pt);
          \draw (1.2,0) -- (3.9,0);
          \node[anchor=west] at (1.2,-0.4) {semantic image};
          \filldraw ([xshift=-1.5pt,yshift=-1.5pt]3.9,-0.6) rectangle ++(3pt,3pt);
          \draw (1.2,-0.6) -- (3.9,-0.6);
          
          \draw (0,-1.8) -- (4,-1.8);
          \node[anchor=west] at (0,-1.6) {hyperparameter selection};
          \draw (4,-1.5) -- (4,-2.1);
          \filldraw ([xshift=-1.5pt,yshift=-1.5pt]7.1,-1.5) rectangle ++(3pt,3pt);
          \draw (4,-1.5) -- (7.1,-1.5);
          \node[anchor=west] at (4.0,-1.3) {anomaly validation};
          \filldraw ([xshift=-1.5pt,yshift=-1.5pt]6.7,-2.1) rectangle ++(3pt,3pt);
          \draw (4,-2.1) -- (6.7,-2.1);
          \node[anchor=west] at (4.0,-1.9) {clean validation};
          
          \draw (0,-3.1) -- (2,-3.1);
          \node[anchor=west] at (0,-2.9) {economic};
          \draw (2,-2.8) -- (2,-3.3);
          \filldraw ([xshift=-1.5pt,yshift=-1.5pt]3.7,-2.8) rectangle ++(3pt,3pt);
          \draw (2,-2.8) -- (3.7,-2.8);
          \node[anchor=west] at (2,-2.6) {train time};
          \filldraw ([xshift=-1.5pt,yshift=-1.5pt]3.7,-3.3) rectangle ++(3pt,3pt);
          \draw (2,-3.3) -- (3.7,-3.3);
          \node[anchor=west] at (2,-3.1) {test time};

    \end{tikzpicture}
    
    \caption{Various aspects of anomaly detection comparison forming the \textit{context} of the experiment.}
    \label{fig:context}
\end{figure}

While many practitioners are eager to see which method is best for their application, the specifics of the application may differ. We have conducted a large number of experiments to identify the main sources of variability influencing the performance of anomaly detection methods. The number of combinations of these aspects is huge, therefore we identified the key axes of variability: datasets, hyperparameter selection strategy, and economic point of view. From these axes, we select a few discrete points, on which we will provide a comparison. The particular combination of the selected aspect will be called \emph{context}, see Fig.~\ref{fig:context} for illustration. 

The first axis is the target data domain. Our experiments used two types of datasets: \textit{tabular} and \textit{image}. This is the most obvious split and indeed most authors of prior art test their methods on either choice of data. Another possible way to look at data is whether they contain \textit{statistical} or \textit{semantic} anomalies. Statistical anomalies should be located in areas of low likelihood of the normal class, while semantic~\citep{ahmed2020detecting} anomalies cannot be differentiated from normal data in a statistical way. This is because they appear in datasets with multiple sources of variations, where only some of them are considered anomalous. Such types of anomalies are most common in image datasets.  Imagine a detector that aims to learn a representation of birds from images without preprocessing. Most of the bird pictures are going to have sky in the background. Since the background occupies most of a picture and therefore has a strong signal, a bird on grass is going to be a statistical anomaly, while an airplane with sky in the background is an example of a semantic anomaly in case the original goal was to identify pictures that do not contain a bird. The suitability of the tested methods for the dataset context axis is studied in Sec.~\ref{sec:dataset_context}.

The second axis of variability is the hyperparameter selection strategy. It should be a gold standard that the experiments are repeated on different splits of data to training, validation, and testing subsets, especially if the datasets are small. However, in most of the reviewed recent papers~\citep{liu2019generative,wang2020advae,schlegl2017unsupervised, akcay2018ganomaly, perera2019ocgan}, this procedure was not mentioned with the exception of~\citep{ruff2018deep}. Therefore, our comparison fills this gap. Also, it is important to define the nature of information available for selection of the hyperparameters: it is indeed a very different task if there is some (often small) number of known anomalies in the validation dataset that can be used to choose hyperparameters by cross-validation or if the validation dataset is clean. In our experience, the former case is more common. Our observations are summarised in Sec.~\ref{sec:hyperparameter_context}.

The third axis is the economic aspect of a problem. There might be serious computational restrictions present in solving real-life problems. One might then not opt for a method that promises state-of-the-art performance, but for another that reaches slightly worse performance but can be trained economically and its performance is robust with regards to hyperparameter optimization. More details on this can be found in Sec.~\ref{sec:economic_context}.

Finally, Sec.~\ref{sec:other_context} contains other influences that we have originally considered to be important, but in the end did not prove to make a significant difference in comparison of multiple methods. These include the use of performance measures other than traditional AUC, the use of Bayesian optimization and others. 

\section{Compared methods}
\label{sec:comparedmethods}
This section briefly reviews deep generative models in the order of exactness of calculation of likelihood. Therefore, it starts with flow models, continues with probabilistic (variational) autoencoders, where the prior art on the application in anomaly detection is rich, and finishes with generative adversarial networks where the calculation of any score related to likelihood is dubious at best. We specifically focus on issues that affect the performance of the method for anomaly detection, most often the anomaly score, if it is not rigorously defined. We also briefly review other examples of deep methods that are relevant for comparison such as two-stage models and distance-based models that are not generative but can be used in anomaly detection. We do not review classical methods here, as this has been done many times elsewhere, but we list them in the relevant experimental section.

Before the description, we introduce a notation. Training samples, $\vec{x},$ are assumed to be i.i.d from the underlying probability distribution $p({\vec{x}})$ defined on the input space $\mathcal{X}$. Following the conventional definition of an anomaly~\citep{barnett1974outliers}, each anomaly detection method is expected to provide a quantity (called score and denoted $s(\vec{x}')$) related to the probability of a sample $\vec{x}'$ being generated from $p(\vec{x}).$ The score does not need to be a normalized distribution, as the threshold is typically determined as an empirical estimate of the quantile. Most functions in this section are assumed to have parameters optimized during training. 

\subsection{Normalizing flows}
The name normalizing flows refers to methods relying on the change of variables formula
\begin{equation}
    p\left(\vec{x}\right) = p\left(\vec{z}\right)\!\left\vert \text{det} J_f\!\left(\vec{z}\right) \right\vert^{-1}, \, \vec{z} = f^{-1}\!\left(\vec{x}\right),
\label{eq:rv_transformation}
\end{equation}
where $J_f\!\left(\vec{z}\right)$ is Jacobi matrix of function $f$ evaluated at $\vec{z}$. $p(\vec{z})$ is a known distribution of the latent variable $\vec{z}$ from space $\mathcal{Z}$ of the same dimension as $\mathcal{X}$.

Theoretical reviews~\citep{papamakariosNormalizingFlowsProbabilistic2019, kobyzevNormalizingFlowsIntroduction2020} require $f$ to be invertible and both $f$ and $f^{-1}$ to be differentiable. Flow models therefore primarily differ in how they define the class of functions $f$, which ranges from simple affine transformations to solutions of ordinary differential equations. The expressive power comes from their composition as is usual in neural networks. In the comparison, we consider flows on tabular data only, for which we have implemented well-known RealNVP~\citep{dinh2016density} and MAF~\citep{papamakariosMaskedAutoregressiveFlow2018} flows alongside with a promising class of Sum Product Transform networks --- SPTN~\citep{pevny2020sum} combining normalizing flows with a graphical model. The likelihood is used as a natural anomaly score.

Flow models have not yet enjoyed a lot of popularity in anomaly detection~\citep{yamaguchi2019adaflow, schmidtNormalizingFlowsNovelty2019, diasAnomalyDetectionTrajectory2020a, pevny2020sum} in comparison to auto-encoders reviewed below. To us, this is surprising, since these methods can exactly calculate likelihood functions, which under a good fit are the ideal anomaly score. Meanwhile, the focus of the surrounding community is on the topic of \textit{out of distribution detection} (OOD)\footnote{Out of distribution detection means identifying samples coming from a different dataset. For example, a model trained on MNIST / CIFAR10 should assign a low likelihood to samples from Fashion MNIST / SVHN respectively.}~\citep{nalisnickDeepGenerativeModels2019}, which is very related to anomaly detection if not being equal. Ref.~\citep{choiWAICWhyGenerative2019} suggests to use ensembles, while~\citep{renLikelihoodRatiosOutofDistribution2019} recommends to convert the single-class problem to classification problems in the spirit of \citep{steinwart2005a}. A deep investigation of OOD in~\citep{kirichenkoWhyNormalizingFlows2020}, shows that with low-level features such as pixel intensities, flows tend to learn local models, i.e. according to taxonomy in~\citep{ruff2020unifying} they fail to detect semantic anomalies.

\subsection{Autoencoder-based models}
\label{sec:ae_theory}
Autoencoder-based models differ from flows by relaxing the exact mapping between $\vec{x}=f(\vec{z})$ \eqref{eq:rv_transformation} into a probability distribution $p_{\vec{\theta}} (\vec{x}|\vec{z})=\mathcal{N}(\vec{\mu}_{\vec{\theta}}(\vec{z}), \mathrm{diag}(\vec{\sigma}_{\vec{\theta}}(\vec{z})))$,\footnote{Other forms of the distribution are possible, e.g. Bernoulli for scaled pixel intensities.} called \emph{decoder}. The symbol $\vec{\theta}$ denotes the trainable parameters of the decoder, e.g. weights of a neural network. The marginal likelihood is computed as
\begin{equation}
    p(\vec{x})=\int p_{\vec{\theta}}(\vec{x}|\vec{z})p(\vec{z}) d\vec{z},
    \label{eq:vae-px}
\end{equation}
where $p(\vec{z})$  is a chosen prior probability distribution of the latent variable. This relaxation allows for more flexible models, e.g. using different dimension of $\vec{x}$ and $\vec{z}$. However, training and evaluation of the model is more demanding since the marginal likelihood \eqref{eq:vae-px} is not available in closed form. Therefore~\citep{kingma2013auto} introduces \emph{encoder} distribution $q_{\vec{\phi}}(\vec{z}|\vec{x})=\mathcal{N}(\vec{\mu}_{\vec{\phi}}(\vec{x}), \mathrm{diag}(\vec{\sigma}_{\vec{\phi}}(\vec{x})))$  with parameters $\vec{\phi}$ allowing approximation of Equation~\eqref{eq:vae-px} as described below in Equation~\ref{eq:vae_loss}.

Various modifications of the original formulation have been proposed, giving rise to many specialized methods. Below we describe extensions in three blocks according to i) approximation of the likelihood \eqref{eq:vae-px} used for training, ii) prior model, iii) approximations used for evaluating the anomaly score, and iv) various modifications of the original concept.

\paragraph{Training loss}
The original Variational Autoencoder~\citep{kingma2013auto} (VAE) proposes to replace \eqref{eq:vae-px} by the evidence lower bound (ELBO)
\begin{equation}
    \mathcal{L}_{\text{VAE}} (\vec{\theta}, \vec{\phi}) = - \mathbb{E}_{q_{\vec{\phi}}(\vec{z}|\vec{x})} \left[ \log p_{\vec{\theta}}(\vec{x}|\vec{z}) \right] + D_{KL} \left( q_{\vec{\phi}}(\vec{z}|\vec{x}) || p(\vec{z}) \right),
\label{eq:vae_loss}
\end{equation}
which combines reconstruction error with regularization term is form of the Kullback-Leibler divergence (KLD) between the encoder distribution and the prior. Models based on~\eqref{eq:vae_loss} will be referred to as the VAE family.

Asymmetry of the KL divergence motivated search for a more accurate metric. Ref.~\citep{tolstikhin2017wasserstein} proposes to replace KL by a Wasserstein divergence, yielding training loss function in the form:
\begin{equation}
    \mathcal{L}_{\text{WAE}} (\vec{\theta}, \vec{\phi}) = - \mathbb{E}_{q_{\vec{\phi}}} \left[ \log p_{\vec{\theta}}(\vec{x}|\vec{z}) \right] + \lambda D \left( q_{\vec{\phi}}(\vec{z}|\vec{x}) || p(\vec{z}) \right),
\label{eq:wae_loss}
\end{equation}
where $\lambda >0$  is a scalar hyperparametr, an $D$  is an arbitrary divergence. The most commonly used divergence is the  kernelized maximum-mean-discrepancy (MMD) with kernel  $k$  which was reported to performs well in matching high dimensional distributions~\citep{zhao2017infovae}. Models based on~\eqref{eq:wae_loss} will be referred to as the WAE  family.

An alternative choice of the divergence $D$ in~\eqref{eq:wae_loss} proposed in~\citep{tolstikhin2017wasserstein} is the adversarial loss, which in combination with the Gaussian decoder coincides with the adversarial autoencoder~\citep{makhzani2015adversarial}. This divergence introduces a third network $d_{\vec{\psi}}(\vec{z}):\mathcal{Z} \rightarrow \left[ 0,1 \right]$,  called discriminator, trained to distinguish between samples from the prior $p(\vec{z})$ and samples $x$ projected by the encoder $q(z|x)$. Every step of optimization separately updates the autoencoder and discriminator parts to minimize the loss functions
\begin{align}
\label{eq:aae_loss}
\mathcal{L}_{\text{AE}}(\vec{\theta}, \vec{\phi}) & = - \mathbb{E}_{q_{\vec{\phi}}(\vec{z}|\vec{x})} \left[ \log p_{\vec{\theta}}(\vec{x}|\vec{z}) \right] - \lambda \log d_{\vec{\psi}}(\vec{z}^q), \\
\mathcal{L}_{\text{D}}(\vec{\psi}) & = \log d_{\vec{\psi}}(\vec{z}^p) + \log \left(1-d_{\vec{\psi}}(\vec{z}^q)\right),
\end{align}
respectively, where $\vec{z}^p \sim p(\vec{z})$, $\vec{z}^q \sim q_{\vec{\phi}}(\vec{z}|\vec{x})$. Models trained with loss function~(\ref{eq:aae_loss}) will be denoted as AAE.

\paragraph{Prior model}
A common criticism of the VAE model is its use of the standard Gaussian prior $p(z),$ which stimulates the distribution $q(z|x)p(x)$ to have a single mode, and therefore it is hard to fit data with a multi-modal latent distribution. Ref.~\citep{tomczak2018vae} proposes a learnable multimodal \emph{Vamp} prior realized as a mixture of $K$ independent Gaussian components. Vamp prior is compatible with AAE and WAE models since it does not have an analytical expression of KLD in~\eqref{eq:vae_loss}. The mean values of components of the mixture are learned together with the parameters of the autoencoder. In the model selection below, the Vamp prior is considered as a binary hyperparameter with an additional parameter, $K,$ specifying the number of components.

\paragraph{Anomaly Score}
The likelihood function \eqref{eq:vae-px} also constitutes the ideal anomaly score. Some training losses such as ELBO \eqref{eq:vae_loss} were designed as approximations of the likelihood and can thus be used as anomaly scores. However, this interpretation is not so clear for other  training losses, i.e. \eqref{eq:wae_loss}, \eqref{eq:aae_loss}, hence their authors propose anomaly scores as part of the method. Nevertheless, many scores are interchangeable, giving rise to another degree of freedom (hyperparameter) for the use of autoencoders in anomaly detection. A common score is based on the first term in the loss i.e. a Monte Carlo estimate of the expectation of conditional log-likelihood over the encoder, yielding \begin{align}
s_{\text{rs}}(\vec{x}) = & - \frac{1}{L}\sum_{l=1}^L \log p_{\vec{\theta}}(\vec{x}| \vec{z_l}),   \vec{z}_l \sim q_{\vec{\phi}}(\vec{z}|\vec{x}). \label{eq:score_sample}
\end{align}
This score, called sampled reconstruction error (abbreviated as rs), was shown in~\citep{xu2018unsupervised} to be more accurate than evaluating~\eqref{eq:vae-px} by sampling $z$ from the prior $p(\vec{z})$. Further simplification is based on replacing samples from the encoder  by its mean, yielding the common reconstruction error score (abbreviated as rm)
\begin{align}
s_{\text{rm}}(\vec{x}) = & - \log p_{\vec{\theta}}(\vec{x}| \vec{\mu}_{\vec{\phi}}(\vec{x})) \label{eq:score_mean}
\end{align}
The usage of~\eqref{eq:score_mean} is justified by the assumption that taking the mean at the encoder should approximate~\eqref{eq:score_sample} while having lower computational demands. For adversarial autoencoders, these simplifications can be combined with the discriminator score~\citep{schlegl2017unsupervised, zenatiEfficientGANBasedAnomaly2018},
\begin{equation}
    s_{\text{a}}(\vec{x}) = \alpha s_{\text{rm}}(\vec{x}) + (1-\alpha) d_{\vec{\psi}}(\vec{\mu}_{\vec{\phi}}(\vec{x})), \alpha \in \left[ 0, 1 \right].
\label{eq:aae_score}
\end{equation}

The reconstruction-error based anomaly scores were criticized in~\citep{pidhorskyi2018generative}  for not capturing the true data density $p(\vec{x}).$ The proposed replacement is based on the orthogonal decomposition of the data into $x=x^\bot +x^{\parallel} $ where the $x^\parallel$  lies in the tangent space of to the manifold defined by the decoder. This allows to decompose the marginal likelihood into a product of two orthogonal parts
\begin{equation}
    p(\vec{x}) \approx p(\vec{x}^{||})p(\vec{x}^{\perp}),
\label{eq:jacodeco}
\end{equation}
where $p(\vec{x}^\bot)$ is the reconstruction error, and $p(\vec{x}^\parallel)$ is obtained by transformation of variables~\eqref{eq:rv_transformation}. This score is abbreviated as \textit{jc} in the following text. The calculation of~\eqref{eq:jacodeco} is expensive, as it needs to compute the singular value decomposition of the Jacobian. For implementation details, see~\citep{pidhorskyi2018generative} or~\citep{vsmidl2019anomaly}.

\paragraph{Other models and techniques}
 A plethora of models based on probabilistic autoencoders and specialized for anomaly detection was introduced in recent years, such as~\citep{zong2018deep, pereira2018unsupervised, xu2018unsupervised, principi2017acoustic, chen2018unsupervised, chalapathyGroupAnomalyDetection2018}. Below, we list models included in the comparison and not described above.

The self-adversarial Variational Autoencoder (adVAE)~\citep{wang2020advae} was included because it claims superiority over the state-of-the-art methods, such as VAE, DAGMM~\citep{zong2018deep}, WGAN-GP~\citep{gulrajani2017improved} or MO-GAAL~\citep{liu2019generative} on tabular datasets. It augments the usual encoder-decoder pair with a transformer, whose goal is to simulate anomalies during training. The seeming flaw of the model is that it is trained only on normal data and there is no link between real and simulated anomalies. The sampled reconstruction is used as an anomaly score.

Despite its name, GANomaly~\citep{akcay2018ganomaly, ahnDeepGenerativeModelsBased2020} is more related to adversarial autoencoders than to GANs. It consists of encoder-decoder-encoder with a discriminator, similar to an AAE. The anomaly score is the difference between latent representations of a sample after the first and second encoding. An upgrade to this model, skip-GANomaly~\citep{akcay2019skip}, uses skip connections in a U-Net type architecture. Here, the anomaly score is a combination of reconstruction error and feature-matching loss (see the next section on fmGAN). Although originally proposed only for use in images, we have implemented a variant for tabular data as well.  

\subsection{Generative adversarial networks} \label{sec:gan}
Generative adversarial networks~\citep{goodfellow2014generative} construct and train two networks: a generator $g_{\vec{\phi}}(\vec{z}):\mathcal{Z} \rightarrow \mathcal{X}$ and a discriminator $d_{\vec{\psi}}(\vec{x}) :\mathcal{X} \rightarrow \left[ 0,1 \right]$ approximating the probability
of $\vec{x}$ being a sample from the data distribution rather than the generator. The generator aims to transforms samples from $p(\vec{z}) = \mathcal{N}(\vec{0}, \vec{I})$ to $\mathcal{X}$ such that they are indistinguishable from the real data. Formally, the  optimization objectives are
\begin{align}
\mathcal{L}_{\text{d}}(\vec{\psi}) = & \log d_{\vec{\psi}}(\vec{x}) + \log \left( 1 - d_{\vec{\psi}}(g_{\vec{\phi}}(\vec{z})) \right) , \\
    \mathcal{L}_{\text{g}}(\vec{\phi}) = &- \log d_{\vec{\psi}}(g_{\vec{\phi}}(\vec{z})),\label{eq:gan-Lg}
\end{align}
where $\mathcal{L}_d$ is maximized while $\mathcal{L}_g$  is minimized, $\vec{z}$ is sampled from the prior and $\vec{x}$ from the training set. The optimization searches the saddle point of the two losses, which is difficult and notoriously unstable. Therefore a long series of work, e.g.~\citep{hong2019generative}, proposes improvements over the basic approach~\citep{goodfellow2014generative}. One of the approaches is based on the introduction of  feature-matching loss~\citep{salimans2016improved}. We will denote the model trained with this loss as feature-matching GAN  (fmGAN). In fmGAN training, the cost function of the generator~\ref{eq:gan-Lg} is augmented with output of some intermediate layer of the discriminator. Specifically, the generator is optimized as 
\begin{equation}
    \mathcal{L}_{\text{fm}}(\vec{\phi}) = \alpha \mathcal{L}_{\text{g}}(\vec{\phi}) + || h_{\vec{\psi}}(\vec{x}) - h_{\vec{\psi}}(g_{\vec{\phi}}(\vec{z})) ||^2,
    \label{eq:fmloss}
\end{equation}
where $h_{\vec{\psi}}$ is the output of the intermediate layer of the discriminator and $\vec{z}$ is a sample from $p(\vec{z})$. This feature-matching loss is used in AnoGAN~\citep{schlegl2017unsupervised} for detection of anomalous objects in images, with hyperparameter $\alpha$, which was zero in the original publication \cite{salimans2016improved}. 

GANs are frequently augmented with a third model $q(z|x)$~\citep{donahue2016adversarial} which makes the distinction between GANs and VAEs blurred, as demonstrated by using min-max (GAN-like) approximation of Wasserstein divergence in Wasserstein autoencoders, Sec.~\ref{sec:ae_theory}.  This makes it sometimes hard to assign a model to some class (for example GANomaly belongs according to us to probabilistic autoencoders).

Recall the role of the discriminator is to discriminate \textit{generated} samples from \textit{real} ones. Since the generator is trained to generate samples with a high discriminator score, it seems logical to use the discriminator to score anomalies
\begin{equation}
     s_{\text{GAN}}(\vec{x}) = 1 - d_{\vec{\psi}}(\vec{x}),
     \label{eq:disc_score}
\end{equation}
which is used e.g. in~\citep{liu2019generative}. The common critique is that the discriminator was not trained to recognize an arbitrary distribution of the anomalies, but only that of the latent transformed by the generator. Thus it may fail to recognize anomalous samples of interest. 
AnoGAN recognizes this flaw and proposes to augment the discriminator loss~\eqref{eq:fmloss} for training and an iterative procedure that searches for the latent code $z$ most likely to generate the tested sample to identify anomalous images. However, this procedure is computationally expensive. Its sequel, f(ast)AnoGAN~\citep{schleglFAnoGANFastUnsupervised2019}, uses Wasserstein GAN with gradient penalization~\citep{gulrajani2017improved} to improve stability and adds $q(z|x)$ to find $z$ closest to given $x$ in $q(x|z)$ faster. The anomaly score of fAnoGAN is a combination of discriminator score and feature-matching loss.

Multiple-Objective Generative Adversarial Active Learning (MOGAAL)~\citep{liu2019generative}, train $k$ generators against a single discriminator on input data divided into $k$ subsets. The usual discriminator score in Equation~\eqref{eq:disc_score} is used to test new samples.
Other anomaly detection models derived from GAN include~\citep{zenatiEfficientGANBasedAnomaly2018, kliger2018novelty, perera2019ocgan}, however, it seems that autoencoder-based models are more popular for anomaly detection and our selected candidates are representative.

\subsection{Two-stage models}
A recurring idea~\citep{ergen2017unsupervised, yaoUnsupervisedAnomalyDetection2019, ruff2018deep, vskvara2020detection} is to combine autoencoders with a secondary model acting on the latent space defined by the encoder. The motivation behind it is that the encoder should preserve the semantic information of the sample and remove noise (e.g. background in images). Additionally, reducing the size mitigates the curse of dimensionality,  as high dimensions can be problematic for some models.

We are not aware of a general term for this approach. We use the term \textit{two-stage models}, following~\citep{dai2019diagnosing}, although~\citep{chalapathy2019deep} uses the term \textit{deep hybrid models}. In~\citep{ruff2018deep}, the model optimizes the projection of data (by virtue of NNs) to a new space, where they can be easily enclosed in a sphere of minimum radius. The approach presented in~\citep{vskvara2020detection, yaoUnsupervisedAnomalyDetection2019} explicitly splits the creation of the detector into two parts. It first trains a VAE (and its variants)  and then it fixes the encoder. The anomaly score is calculated by a kNN \citep{vskvara2020detection} or by OC-SVM \citep{yaoUnsupervisedAnomalyDetection2019} detectors in the latent space, obtained by projecting the sample by the fixed encoder. The two-stage models can be also viewed as a kNN with a trained metric or OC-SVM with a trained kernel. The embedding can be optimized differently, for example by enforcing the margin between anomaly candidates and normal data as done in the REPEN~\citep{pangLearningRepresentationsUltrahighdimensional2018} method, which uses an ensemble of 1NN detectors as the second stage.

\section{Experimental setup}
\label{sec:experimentalsetup}
\subsection{Datasets}
\label{sec:datasets}
The choice of datasets (mainly the tabular ones), was guided by two criteria: first, they ought to be publicly available and second, they should appear in surveys or in articles presenting new methods.  In total, we have collected $40$ tabular datasets, the majority of which came from the UCI repository~\citep{Dua:2019}. With the exception of ANNThyroid, Arrhytmia, HAR, HTRU2, KDD Cup 99 (small), Spambase, Mammography, and Seismic, where the anomaly class has a clear meaning (security incident or disease), we have followed the technique of~\citep{emmott2013systematic} for creating artificial datasets for anomaly detection tasks from classification datasets. More precisely, we have used only "easy" and "medium" anomalies, as "hard" and "very hard" are not truly anomalous in the sense of being clearly statistically distinct from the normal class. Prior to model training, features on tabular datasets were normalized to have zero mean and unit variance. Further details of the datasets are provided in Tab.~\ref{tab:tabular_datasets}.

The number of image datasets used for the evaluation of deep models is limited, as there are very few datasets designed purely for anomaly detection - MNIST-C~\citep{muMNISTCRobustnessBenchmark2019} and MVTec-AD~\citep{bergmannMVTecADComprehensive2019}. Therefore, we have decided to extend these with artificially created anomaly datasets based on common image datasets MNIST~\citep{lecun-mnisthandwrittendigit-2010}, FashionMNIST~\citep{xiao2017fashion}, CIFAR10~\citep{krizhevsky2009learning}, and SVHN2~\citep{netzer2011reading}. These are also used for anomaly detection tasks in the prior art~\citep{perera2019ocgan, pidhorskyi2018generative, ruff2018deep}. Most often a single class (of digits/objects) is considered normal and the rest anomalous, which is the primary setting of our experiments. Only three subsets \textit{wood, grid, transistor} of the MVTec-AD dataset were used due to our computational constraints. These were chosen since they represent problems of various degrees of difficulty. Downsampling to $128 \times 128$ was required to fit the computational envelope for most methods on the real-world high-resolution images in MVTec-AD.  We have linearly extrapolated the image data when training GANomaly so that the input dimensions were a multiple of 16 (this is a result of the model's fixed architecture). No other preprocessing has been applied prior to training since the source data already have all channels scaled to [0,1].  Basic statistics on image datasets are shown in table Tab.~\ref{tab:image_datasets}. 

We have performed a visual inspection of the nature of anomalies in the image datasets, see Supplementary~\ref{sec:appendix_extending_image_results}. Since most of the datasets that were manually processed (MNIST, FashionMNIST, MVTec-AD and MNISTC) have a rich and consistent number of samples in the normal class and clear anomalies, we consider them to be statistical anomalies. On the other hand, images in the majority of classes in  CIFAR10 and SVHN2 have strong background and are thus considered to contain semantic anomalies. This prior division is also supported by a different behavior of different methods as reported in the Supplementary, Figure~\ref{fig:image_knowledge_rank_pat_auc}. 

\subsection{Data splits and experiment repetitions}
\label{sec:repetitions}
The experiments on tabular data and MNIST-C/MVTec-AD image datasets were repeated five times with different random cross-validation splits. Specifically, in each repetition (five in total) of an experiment with the same model hyper-parameters, the \textit{normal data} in each dataset was randomly split in 60\%/20\%/20\% ratios to train/validation/test subsets, respectively. \textit{Anomalous data} were split such that 50\% were in the validation part and 50\% in the testing part, which means the training subset has not contained anomalous samples. \footnote{A training set without any anomalies is in practice very optimistic, but this decision removes another degree of freedom from the evaluation for the sake of clarity of results.} The proportion of anomalies that were used in the validation phase varied from zero to the selected 50\%. 

On the rest of the image datasets, due to the already substantial computational requirements, we have not trained models with the same hyperparameters on repeated random cross-validation splits. In our experience, the results on different splits of these datasets are almost the same, since the number of samples is large and our trial experiments (see Supplementary Tab.~\ref{tab:images_seed_consistency}) have not exhibited a large variation between the different random cross-validation experiment repetitions. 

\begin{table}
\centering
\tabcolsep=0.1cm
\begin{tabular}{llrrr}
\toprule
\textbf{dataset} & \textbf{alias} & \textbf{dim} & \textbf{anom} & \textbf{normal}   \\\midrule
ANNthyroid & ann  & 21 & 534 & 6665 \\
Arrhythmia & arr  & 275 & 206 & 245 \\
HAR & har & 561 & 1944 & 8355  \\
HTRU2 & htr & 8 & 1638 & 16257  \\
KDD99 (10\%) & kdd & 118 & 396742 & 97276  \\
Mammography & mam & 6 & 260 & 10921  \\
Seismic & sei  & 24 & 170 & 2412  \\
Spambase & spm & 57 & 1812 & 2786  \\
\midrule
Abalone & aba & 10 & 50 & 2151  \\
Blood Transfusion & blt & 4 & 16 & 382  \\
Breast Cancer Wisconsin & bcw  & 30 & 206 & 356 \\
Breast Tissue & bts & 9 & 22 & 65 \\
Cardiotocography & crd & 27 & 228 & 1830  \\
Ecoli & eco & 7 & 108 & 205  \\
Glass & gls & 10 & 94 & 112  \\
Haberman & hab & 3 & 14 & 225  \\
Ionosphere & ion & 33 & 122 & 225  \\
Iris & irs & 4 & 46 & 100  \\
Isolet & iso & 617 & 3300 & 4496  \\
Letter Recognition & ltr & 617 & 3600 & 4196  \\
Libras & lbr & 90 & 142 & 215  \\
Magic Telescope & mgc & 10 & 3882 & 12331  \\
Miniboone & mnb & 50 & 23922 & 93565  \\
Multiple Features & mlt & 649 & 800 & 1200  \\
PageBlocks & pgb & 10 & 384 & 4911  \\
Parkinsons & prk & 22 & 44 & 146  \\
Pendigits & pen & 16 & 5384 & 5537  \\
Pima Indians & pim & 8 & 176 & 500  \\
Sonar & snr & 60 & 96 & 110  \\
Spect Heart & sph & 44 & 52 & 211  \\
Statlog Satimage & sat & 36 & 2630 & 3592  \\
Statlog Segment & seg & 18 & 938 & 1320  \\
Statlog Shuttle & sht & 8 & 28 & 57767  \\
Statlog Vehicle & vhc & 18 & 132 & 627  \\
Synthetic Control Chart & scc & 60 & 200 & 400  \\
Wall Following Robot & wrb & 24 & 2220 & 2921  \\
Waveform-1 & wf1 & 21 & 1482 & 3302  \\
Waveform-2 & wf2 & 21 & 1472 & 3302  \\
Wine & wne & 13 & 70 & 106  \\
Yeast & yst & 8 & 390 & 751   \\\bottomrule
\end{tabular}
\vspace*{0.15cm}
\caption{Basic statistics of the tabular dataset designed for anomaly detection  (above split) and multi-class datasets  (bellow split).}
\label{tab:tabular_datasets}
\end{table}

\begin{table}
    \centering
    \tabcolsep=0.1cm
    \begin{tabular}{lllrr}
    \toprule
    \textbf{dataset} & \textbf{alias} & \textbf{dim} & \textbf{anom} & \textbf{normal} \\
    \midrule
    MNIST-C & mnistc & 28x28x1 & 70000 & 70000 \\
    MVTec-AD - wood & wood & 1024x1024x3 & 60 & 266 \\
    MVTec-AD - grid & grid & 1024x1024x3 & 57 & 285 \\
    MVTec-AD - transistor & transistor & 1024x1024x3 & 40 & 273 \\
    \midrule
    CIFAR10 & cifar10 & 32x32x3 & 54000 & 6000  \\
    FashionMNIST & fmnist & 28x28x1 & 63000 & 7000   \\
    MNIST & mnist & 28x28x1 & 63686 & 6312  \\
    SVHN2 & svhn2 & 32x32x3 & 80327 & 18960  \\\bottomrule
    \end{tabular}
    \vspace*{0.15cm}
    \caption{Basic statistics of image datasets designed for anomaly detection (above split) and multi-class datasets (below split).}
    \label{tab:image_datasets}
\end{table}

\subsection{Notes on implementation of models}
As mentioned in the introduction, we have compared various types of \textit{deep} methods to \emph{classical} ones serving as an etalon. Classical methods included ABOD~\citep{kriegel2008angle}, HBOS~\citep{goldstein2012histogram}, LODA~\citep{pevny2016loda}, LOF~\citep{breunig2000lof}, IsolationForest~\citep{liu2008isolation}, OC-SVM~\citep{scholkopf2001estimating}, PIDForest~\citep{gopalanPIDForestAnomalyDetection2019}, and kNN~\citep{ramaswamy2000efficient}. For ABOD, HBOS, and LODA, we have used pyOD library~\citep{zhao2019pyod} implementation, for LOF, IsolationForest, and OC-SVM we used scikit-learn~\citep{scikit-learn} implementation, and last but not least we have used our own implementation of kNN. The acronyms used in the result section are summarized in Tab.~\ref{tab:model_acronyms_2col} together with the classification of the deep methods as described in Sec.~\ref{sec:comparedmethods}.

Since image datasets are typically much larger than tabular, the OC-SVM was implemented as an ensemble of 10 OC-SVM models trained disjoint subsets of data, see Sec.~\ref{sec:appendix_implementation}, because it has at best $O(n^2)$ scaling in the number of samples. 

To ensure consistency among deep models, we implemented all methods except the MOGAAL\footnote{The pyOD implementation was used.} ourselves using the Flux~\citep{innes:2018} framework in Julia~\citep{Julia-2017} language. Apart from the models mentioned in Sec.~\ref{sec:comparedmethods}, we have also implemented DeepSVDD~\citep{ruff2018deep}, DAGMM~\citep{zong2018deep} and REPEN~\cite{pangLearningRepresentationsUltrahighdimensional2018}, which have been included in multiple comparisons~\citep{ruff2020unifying, wang2020advae, chalapathy2018anomaly}.

\begin{table}
    \centering
    \tabcolsep=0.1cm
    
    \begin{tabular}{cll|cll}
    \toprule
    \textbf{class} & \textbf{model} & \textbf{acronym} & \textbf{class} & \textbf{model} & \textbf{acronym}  \\\midrule
    
    \parbox[t]{2mm}{\multirow{3}{*}{\rotatebox[origin=c]{90}{flows}}} & MAF & maf & \parbox[t]{2mm}{\multirow{5}{*}{\rotatebox[origin=c]{90}{two-stage}}} & DAGMM & dgmm \\
        & RealNVP & rnvp & & DeepSVDD & dsvd \\
        & SPTN & sptn & & REPEN & rpn  \\
        & & & & VAE-kNN & vaek \\

    \parbox[t]{2mm}{\multirow{5}{*}{\rotatebox[origin=c]{90}{autoencoders}}} 
        & AAE & aae & & VAE-OC-SVM & vaeo \\
        & adVAE & avae & & & \\

        & GANomaly & gano & \parbox[t]{2mm}{\multirow{8}{*}{\rotatebox[origin=c]{90}{classical}}} & ABOD & abod \\

        & skipGANomaly & skip & & HBOS & hbos \\
        & VAE & vae & & IsolationForest & if \\
        & WAE & wae & & kNN & knn \\
        & & & & LODA & loda \\
    \parbox[t]{2mm}{\multirow{3}{*}{\rotatebox[origin=c]{90}{gans}}}
        & fAnoGAN & fano & & LOF & lof \\ 
        & fmGAN & fmgn & & OC-SVM & osvm \\
        & GAN & gan & & PidForest & pidf \\
        & MOGAAL & mgal & \\
    
    \bottomrule
    \end{tabular}

    \vspace*{0.15cm}
    \caption{Overview of the main classes of compared methods and the acronyms used in the text.}
    \label{tab:model_acronyms_2col}
\end{table}

We emphasize that while implementing models, we have carefully compared our implementations to the reference where possible and (or) verified that our experimental results are similar to those provided in the corresponding publication.

Neural networks were trained with the ADAM~\citep{kingma2014adam} optimizer with early stopping measuring the continued decrease of loss on the validation dataset. All deep models trained on all image datasets used convolutional layers. More implementation details are in the Supplementary materials Sec.~\ref{sec:appendix_implementation}. The implementation code in the form of a Julia package can be found at \textbf{https://github.com/aicenter/GenerativeAD.jl}.

\subsection{Hyperparameters and their optimization}
\label{sub:hyperparameteroptimization}
Properly exploring the space of hyperparameters of all models is paramount to achieving fair and comparable experimental comparison, yet this is often superficially treated. Researchers often use \textit{default} or \textit{recommended} values ignoring that they are sub-optimal on datasets they use in their comparison. A nice demonstration of this are conflicting results of the MOGAAL method in the original publication~\citep{liu2019generative} and in~\citep{wang2020advae}. A prototypical example in the classical methods is OC-SVM, which is typically used with Gaussian kernel and with $\nu$ set to some default value, for example 0.05~\citep{pevny2016loda}, but can achieve better results with different kernels. The choice of hyperparameters in anomaly detection is everything but easy. But this means that the experimental settings should be set up such that all methods have been optimized equally. We conjecture that recommended and default values of hyperparameters are strongly correlated with the choice of evaluation datasets in the publications that recommend them.

\textit{Random grid search:} In order to explore the hyperparameter space of each method properly, we have employed a random search over a predefined grid for each method. This allows the construction of sections through the space for sensitivity studies. Moreover, it is frequently more efficient than grid search~\citep{bergstra2012random} and more flexible. For each model, dataset, and repetition, we have sampled 100 configurations from corresponding sets (see Tab.~\ref{tab:classical_hyperparameters}--\ref{tab:flow_hyperparameters}) and trained the models with them. In order to keep the neural-network-based models fixed across the repetitions on a single dataset, for each hyperparameter configuration we have also sampled a random seed that was used to initialize the network weights. To prevent running the training of expensive methods forever, there was a hard deadline of 24 hours in which the training of a single configuration for a given number of repetitions/classes should be finished. This automatically penalizes complicated models. 

Encoders for the two-stage models were selected from models performing best in terms of validation AUC or reconstruction error on the validation set. The second-stage models (kNN and OC-SVM) used hyperparameters sampled from Tab.~\ref{tab:classical_hyperparameters}. 

\textit{Bayesian optimization:} To overcome the limitation of random sampling in larger configuration spaces we have also trained models whose hyperparameter choice was guided by Bayesian optimization of an evaluation metric on validation data. We followed 50/50 strategy, where the underlying Gaussian process has been fitted with 50 randomly sampled configurations and another 50 have been sampled based on the acquisition function. We have relied on the off-the-shelf implementation from the scikit-optimize framework~\citep{skopt} with default parameters.

\textit{Number of anomalies in the validation set:} A thorough exploration of the hyperparameter selection context also requires changing the criteria of model selection. When anomalies are available for validation, we select hyper-parameters maximizing the AUC on the validation set. For experiments with no available anomalies, we have decided on the following hyperparameter selection mechanism. For \textit{classical} methods, we have used default hyperparameter values from literature - either they were recommended by authors of the method, were used in a survey, or are default in a given implementation. Their overview is in Tab.~\ref{tab:default_hyperparameters}. For \textit{deep} methods, this is unfortunately impossible, since their hyperparameter space is much larger and the values are usually tuned to a specific dataset. Therefore, in order to have a universal solution, we have selected the already trained and evaluated models based on the lowest anomaly score on validation normal data. This approach is theoretically justified for models with proper likelihood. 

\textit{Ensembles:} Some results on ensembles of anomaly detectors were already reported in~\citep{choiWAICWhyGenerative2019, nalisnickDeepGenerativeModels2019}. Since the other experiments required training a large number of models, it was decided to test whether even a naive approach to this problem brings some improvements. In order to mitigate some uncertainty given by different hyperparameter values, ensembles of detectors were constructed by averaging scores of some number of best-performing detectors. We have experimented with different fixed sizes of ensembles --- either top 10 or top 5. In such a case, the anomaly score is the average of the anomaly scores of ensemble members.

\textit{Performance criteria:} While the area under the ROC curve (AUC) is the most common criteria in anomaly detection, some authors use different metrics, such as partial AUC~\citep{dodd2003partial} or the true positive rate at a chosen false negative rate (TPR@). We have re-evaluated all results obtained for the random grid search on the TPR@5\%.

We have kept track of the time spent on training of individual models and also on the time needed to evaluate them on validation and test sets. In total, we have trained 1,364,989 model instances in 9619 CPU days, evaluated 4,256,470 different scoring functions in 2704 CPU days, and created 10.2TB of experimental data.

\section{Experimental results}
Before starting with the description of the experimental results, here we summarize conventions that are used unless said otherwise. The performance results are estimates of the AUC on the testing set and are averaged over five random cross-validation repetitions in the case of tabular and MvTec-AD/MNISTC image datasets. When ranks are reported, they are calculated by ordering methods on each dataset and calculating the average across them (as recommended in~\citep{demvsar2006statistical}). Hyperparameters are selected using the best average performance over 5 seeds, or individually over 10 anomaly classes because the individual class splits constitute different anomaly detection problems. 

\begin{figure*}[h]
    \begin{tabular}{c c}
    \resizebox{\columnwidth}{!}{\begin{tikzpicture}[scale=0.45] 
  \draw (2.0,0) -- (23.0,0); 
  \foreach \x in {2,...,23} \draw (\x,0.10) -- (\x,-0.10) node[anchor=north]{$\x$}; 
  \draw (2.9,0) -- (2.9,0.5) -- (1.9, 0.5) node[anchor=east] {\textcolor{gray}{osvm}}; 
  \draw (6.5,0) -- (6.5,1.0999999999999999) -- (1.9, 1.0999999999999999) node[anchor=east] {vae}; 
  \draw (6.8,0) -- (6.8,1.6999999999999997) -- (1.9, 1.6999999999999997) node[anchor=east] {vaeo}; 
  \draw (6.9,0) -- (6.9,2.3) -- (1.9, 2.3) node[anchor=east] {aae}; 
  \draw (7.0,0) -- (7.0,2.9) -- (1.9, 2.9) node[anchor=east] {wae}; 
  \draw (8.4,0) -- (8.4,3.4999999999999996) -- (1.9, 3.4999999999999996) node[anchor=east] {\textcolor{gray}{knn}}; 
  \draw (8.9,0) -- (8.9,4.1000000000000005) -- (1.9, 4.1000000000000005) node[anchor=east] {maf}; 
  \draw (9.0,0) -- (9.0,4.7) -- (1.9, 4.7) node[anchor=east] {rnvp}; 
  \draw (10.3,0) -- (10.3,5.3) -- (1.9, 5.3) node[anchor=east] {gano}; 
  \draw (10.7,0) -- (10.7,5.9) -- (1.9, 5.9) node[anchor=east] {sptn}; 
  \draw (11.0,0) -- (11.0,6.5) -- (1.9, 6.5) node[anchor=east] {vaek}; 
  \draw (11.4,0) -- (11.4,7.1) -- (1.9, 7.1) node[anchor=east] {\textcolor{gray}{abod}}; 
  \draw (11.4,0) -- (11.4,6.8) -- (23.1, 6.8) node[anchor=west] {\textcolor{gray}{lof}}; 
  \draw (11.4,0) -- (11.4,6.2) -- (23.1, 6.2) node[anchor=west] {fmgn}; 
  \draw (12.0,0) -- (12.0,5.6) -- (23.1, 5.6) node[anchor=west] {gan}; 
  \draw (12.1,0) -- (12.1,5.0) -- (23.1, 5.0) node[anchor=west] {rpn}; 
  \draw (13.4,0) -- (13.4,4.4) -- (23.1, 4.4) node[anchor=west] {avae}; 
  \draw (14.3,0) -- (14.3,3.8) -- (23.1, 3.8) node[anchor=west] {\textcolor{gray}{pidf}}; 
  \draw (14.5,0) -- (14.5,3.2) -- (23.1, 3.2) node[anchor=west] {\textcolor{gray}{if}}; 
  \draw (14.9,0) -- (14.9,2.6) -- (23.1, 2.6) node[anchor=west] {\textcolor{gray}{hbos}}; 
  \draw (16.1,0) -- (16.1,1.9999999999999998) -- (23.1, 1.9999999999999998) node[anchor=west] {\textcolor{gray}{loda}}; 
  \draw (16.2,0) -- (16.2,1.4) -- (23.1, 1.4) node[anchor=west] {dsvd}; 
  \draw (19.8,0) -- (19.8,0.8) -- (23.1, 0.8) node[anchor=west] {dagm}; 
  \draw (22.3,0) -- (22.3,0.2) -- (23.1, 0.2) node[anchor=west] {mgal}; 
  \draw[line width=0.03cm,color=black,draw opacity=1.0] (2.87,0.15) -- (7.03,0.15); 
  \draw[line width=0.03cm,color=black,draw opacity=1.0] (6.47,0.3) -- (11.43,0.3); 
  \draw[line width=0.03cm,color=black,draw opacity=1.0] (6.77,0.44999999999999996) -- (12.129999999999999,0.44999999999999996); 
  \draw[line width=0.03cm,color=black,draw opacity=1.0] (8.370000000000001,0.6) -- (13.43,0.6); 
  \draw[line width=0.03cm,color=black,draw opacity=1.0] (8.870000000000001,0.75) -- (14.33,0.75); 
  \draw[line width=0.03cm,color=black,draw opacity=1.0] (10.270000000000001,0.9) -- (14.93,0.9); 
  \draw[line width=0.03cm,color=black,draw opacity=1.0] (10.67,1.05) -- (16.130000000000003,1.05); 
  \draw[line width=0.03cm,color=black,draw opacity=1.0] (10.97,1.2) -- (16.23,1.2); 
  \draw[line width=0.03cm,color=black,draw opacity=1.0] (14.47,0.3) -- (19.830000000000002,0.3); 
  \draw[line width=0.03cm,color=black,draw opacity=1.0] (19.77,0.15) -- (22.330000000000002,0.15); 
 \end{tikzpicture} } & \resizebox{\columnwidth}{!}{\begin{tikzpicture}[scale=0.65] 
  \draw (6.0,0) -- (21.0,0); 
  \foreach \x in {6,...,21} \draw (\x,0.10) -- (\x,-0.10) node[anchor=north]{$\x$}; 
  \draw (6.0,0) -- (6.0,0.30000000000000004) -- (5.9, 0.30000000000000004) node[anchor=east] {\textcolor{gray}{knn}}; 
  \draw (6.0,0) -- (6.0,0.7000000000000001) -- (5.9, 0.7000000000000001) node[anchor=east] {\textcolor{gray}{osvm}}; 
  \draw (7.1,0) -- (7.1,1.1) -- (5.9, 1.1) node[anchor=east] {rnvp}; 
  \draw (7.2,0) -- (7.2,1.5) -- (5.9, 1.5) node[anchor=east] {maf}; 
  \draw (7.4,0) -- (7.4,1.9) -- (5.9, 1.9) node[anchor=east] {sptn}; 
  \draw (8.0,0) -- (8.0,2.3000000000000003) -- (5.9, 2.3000000000000003) node[anchor=east] {\textcolor{gray}{lof}}; 
  \draw (8.6,0) -- (8.6,2.7) -- (5.9, 2.7) node[anchor=east] {\textcolor{gray}{abod}}; 
  \draw (8.6,0) -- (8.6,3.1) -- (5.9, 3.1) node[anchor=east] {\textcolor{gray}{if}}; 
  \draw (9.0,0) -- (9.0,3.5) -- (5.9, 3.5) node[anchor=east] {wae}; 
  \draw (10.4,0) -- (10.4,3.9) -- (5.9, 3.9) node[anchor=east] {vae}; 
  \draw (10.7,0) -- (10.7,4.300000000000001) -- (5.9, 4.300000000000001) node[anchor=east] {vaek}; 
  \draw (11.3,0) -- (11.3,4.700000000000001) -- (5.9, 4.700000000000001) node[anchor=east] {\textcolor{gray}{hbos}}; 
  \draw (11.6,0) -- (11.6,4.6000000000000005) -- (21.1, 4.6000000000000005) node[anchor=west] {gano}; 
  \draw (11.6,0) -- (11.6,4.2) -- (21.1, 4.2) node[anchor=west] {vaeo}; 
  \draw (12.5,0) -- (12.5,3.8000000000000003) -- (21.1, 3.8000000000000003) node[anchor=west] {aae}; 
  \draw (12.5,0) -- (12.5,3.4000000000000004) -- (21.1, 3.4000000000000004) node[anchor=west] {\textcolor{gray}{pidf}}; 
  \draw (14.8,0) -- (14.8,3.0000000000000004) -- (21.1, 3.0000000000000004) node[anchor=west] {fmgn}; 
  \draw (15.4,0) -- (15.4,2.6000000000000005) -- (21.1, 2.6000000000000005) node[anchor=west] {gan}; 
  \draw (16.2,0) -- (16.2,2.2) -- (21.1, 2.2) node[anchor=west] {avae}; 
  \draw (16.6,0) -- (16.6,1.8) -- (21.1, 1.8) node[anchor=west] {dsvd}; 
  \draw (18.3,0) -- (18.3,1.4000000000000001) -- (21.1, 1.4000000000000001) node[anchor=west] {\textcolor{gray}{loda}}; 
  \draw (19.5,0) -- (19.5,1.0) -- (21.1, 1.0) node[anchor=west] {rpn}; 
  \draw (19.7,0) -- (19.7,0.6000000000000001) -- (21.1, 0.6000000000000001) node[anchor=west] {dagm}; 
  \draw (20.6,0) -- (20.6,0.2) -- (21.1, 0.2) node[anchor=west] {mgal}; 
  \draw[line width=0.03cm,color=black,draw opacity=1.0] (5.97,0.13) -- (12.53,0.13); 
  \draw[line width=0.03cm,color=black,draw opacity=1.0] (7.17,0.26) -- (14.83,0.26); 
  \draw[line width=0.03cm,color=black,draw opacity=1.0] (7.97,0.39) -- (15.43,0.39); 
  \draw[line width=0.03cm,color=black,draw opacity=1.0] (8.57,0.52) -- (16.23,0.52); 
  \draw[line width=0.03cm,color=black,draw opacity=1.0] (8.97,0.65) -- (16.630000000000003,0.65); 
  \draw[line width=0.03cm,color=black,draw opacity=1.0] (10.67,0.78) -- (18.330000000000002,0.78); 
  \draw[line width=0.03cm,color=black,draw opacity=1.0] (12.47,0.91) -- (19.73,0.91); 
  \draw[line width=0.03cm,color=black,draw opacity=1.0] (14.770000000000001,0.13) -- (20.630000000000003,0.13); 
 \end{tikzpicture} } \\
    a) tabular datasets, anomaly validation & b) tabular datasets, clean validation \\  
    \resizebox{\columnwidth}{!}{\begin{tikzpicture}[scale=1.0] 
  \draw (2.0,0) -- (10.0,0); 
  \foreach \x in {2,...,10} \draw (\x,0.10) -- (\x,-0.10) node[anchor=north]{$\x$}; 
  \draw (2.9,0) -- (2.9,0.19999999999999998) -- (1.9, 0.19999999999999998) node[anchor=east] {wae}; 
  \draw (3.1,0) -- (3.1,0.5) -- (1.9, 0.5) node[anchor=east] {vae}; 
  \draw (3.4,0) -- (3.4,0.7999999999999999) -- (1.9, 0.7999999999999999) node[anchor=east] {fano}; 
  \draw (3.6,0) -- (3.6,1.0999999999999999) -- (1.9, 1.0999999999999999) node[anchor=east] {aae}; 
  \draw (4.0,0) -- (4.0,1.4) -- (1.9, 1.4) node[anchor=east] {dsvd}; 
  \draw (4.9,0) -- (4.9,1.6999999999999997) -- (1.9, 1.6999999999999997) node[anchor=east] {vaeo}; 
  \draw (5.1,0) -- (5.1,1.7) -- (10.1, 1.7) node[anchor=west] {gano}; 
  \draw (5.4,0) -- (5.4,1.4) -- (10.1, 1.4) node[anchor=west] {\textcolor{gray}{knn}}; 
  \draw (5.7,0) -- (5.7,1.0999999999999999) -- (10.1, 1.0999999999999999) node[anchor=west] {vaek}; 
  \draw (6.4,0) -- (6.4,0.8) -- (10.1, 0.8) node[anchor=west] {\textcolor{gray}{osvm}}; 
  \draw (7.5,0) -- (7.5,0.5) -- (10.1, 0.5) node[anchor=west] {fmgn}; 
  \draw (9.1,0) -- (9.1,0.2) -- (10.1, 0.2) node[anchor=west] {skip}; 
  \draw[line width=0.06cm,color=black,draw opacity=1.0] (2.87,0.1) -- (5.430000000000001,0.1); 
  \draw[line width=0.06cm,color=black,draw opacity=1.0] (3.37,0.2) -- (5.73,0.2); 
  \draw[line width=0.06cm,color=black,draw opacity=1.0] (3.97,0.30000000000000004) -- (6.430000000000001,0.30000000000000004); 
  \draw[line width=0.06cm,color=black,draw opacity=1.0] (5.069999999999999,0.4) -- (7.53,0.4); 
  \draw[line width=0.06cm,color=black,draw opacity=1.0] (7.47,0.1) -- (9.129999999999999,0.1); 
 \end{tikzpicture} } & \resizebox{\columnwidth}{!}{\begin{tikzpicture}[scale=1.0] 
  \draw (3.0,0) -- (11.0,0); 
  \foreach \x in {3,...,11} \draw (\x,0.10) -- (\x,-0.10) node[anchor=north]{$\x$}; 
  \draw (3.2,0) -- (3.2,0.19999999999999998) -- (2.9, 0.19999999999999998) node[anchor=east] {aae}; 
  \draw (3.6,0) -- (3.6,0.5) -- (2.9, 0.5) node[anchor=east] {vae}; 
  \draw (3.8,0) -- (3.8,0.7999999999999999) -- (2.9, 0.7999999999999999) node[anchor=east] {\textcolor{gray}{knn}}; 
  \draw (4.1,0) -- (4.1,1.0999999999999999) -- (2.9, 1.0999999999999999) node[anchor=east] {wae}; 
  \draw (4.7,0) -- (4.7,1.4) -- (2.9, 1.4) node[anchor=east] {vaek}; 
  \draw (5.2,0) -- (5.2,1.6999999999999997) -- (2.9, 1.6999999999999997) node[anchor=east] {gano}; 
  \draw (5.2,0) -- (5.2,1.7) -- (11.1, 1.7) node[anchor=west] {vaeo}; 
  \draw (5.7,0) -- (5.7,1.4) -- (11.1, 1.4) node[anchor=west] {\textcolor{gray}{osvm}}; 
  \draw (6.0,0) -- (6.0,1.0999999999999999) -- (11.1, 1.0999999999999999) node[anchor=west] {fano}; 
  \draw (7.5,0) -- (7.5,0.8) -- (11.1, 0.8) node[anchor=west] {dsvd}; 
  \draw (9.1,0) -- (9.1,0.5) -- (11.1, 0.5) node[anchor=west] {skip}; 
  \draw (10.1,0) -- (10.1,0.2) -- (11.1, 0.2) node[anchor=west] {fmgn}; 
  \draw[line width=0.06cm,color=black,draw opacity=1.0] (3.1700000000000004,0.1) -- (5.73,0.1); 
  \draw[line width=0.06cm,color=black,draw opacity=1.0] (3.5700000000000003,0.2) -- (6.03,0.2); 
  \draw[line width=0.06cm,color=black,draw opacity=1.0] (5.17,0.30000000000000004) -- (7.53,0.30000000000000004); 
  \draw[line width=0.06cm,color=black,draw opacity=1.0] (7.47,0.2) -- (9.129999999999999,0.2); 
  \draw[line width=0.06cm,color=black,draw opacity=1.0] (9.07,0.1) -- (10.129999999999999,0.1); 
 \end{tikzpicture} }\\
    c) statistical image datasets, anomaly validation & d) statistical image  datasets, clean validation \\ 
     \resizebox{\columnwidth}{!}{\begin{tikzpicture}[scale=1.0] 
  \draw (2.0,0) -- (9.0,0); 
  \foreach \x in {2,...,9} \draw (\x,0.10) -- (\x,-0.10) node[anchor=north]{$\x$}; 
  \draw (2.4,0) -- (2.4,0.19999999999999998) -- (1.9, 0.19999999999999998) node[anchor=east] {fmgn}; 
  \draw (3.8,0) -- (3.8,0.5) -- (1.9, 0.5) node[anchor=east] {vae}; 
  \draw (4.8,0) -- (4.8,0.7999999999999999) -- (1.9, 0.7999999999999999) node[anchor=east] {fano}; 
  \draw (4.9,0) -- (4.9,1.0999999999999999) -- (1.9, 1.0999999999999999) node[anchor=east] {wae}; 
  \draw (5.0,0) -- (5.0,1.4) -- (1.9, 1.4) node[anchor=east] {aae}; 
  \draw (5.6,0) -- (5.6,1.6999999999999997) -- (1.9, 1.6999999999999997) node[anchor=east] {vaek}; 
  \draw (6.2,0) -- (6.2,1.7) -- (9.1, 1.7) node[anchor=west] {dsvd}; 
  \draw (6.6,0) -- (6.6,1.4) -- (9.1, 1.4) node[anchor=west] {vaeo}; 
  \draw (7.8,0) -- (7.8,1.0999999999999999) -- (9.1, 1.0999999999999999) node[anchor=west] {skip}; 
  \draw (7.8,0) -- (7.8,0.8) -- (9.1, 0.8) node[anchor=west] {\textcolor{gray}{knn}}; 
  \draw (8.1,0) -- (8.1,0.5) -- (9.1, 0.5) node[anchor=west] {gano}; 
  \draw (8.3,0) -- (8.3,0.2) -- (9.1, 0.2) node[anchor=west] {\textcolor{gray}{osvm}}; 
  \draw[line width=0.06cm,color=black,draw opacity=1.0] (2.37,0.1) -- (5.63,0.1); 
  \draw[line width=0.06cm,color=black,draw opacity=1.0] (3.77,0.3) -- (6.63,0.3); 
  \draw[line width=0.06cm,color=black,draw opacity=1.0] (4.77,0.4) -- (8.129999999999999,0.4); 
  \draw[line width=0.06cm,color=black,draw opacity=1.0] (4.87,0.2) -- (8.33,0.2); 
  \node[anchor=center] at (5.5,-0.7) {average model rank};
 \end{tikzpicture} } & \resizebox{\columnwidth}{!}{\begin{tikzpicture}[scale=1.0] 
  \draw (3.0,0) -- (10.0,0); 
  \foreach \x in {3,...,10} \draw (\x,0.10) -- (\x,-0.10) node[anchor=north]{$\x$}; 
  \draw (3.4,0) -- (3.4,0.19999999999999998) -- (2.9, 0.19999999999999998) node[anchor=east] {dsvd}; 
  \draw (4.6,0) -- (4.6,0.5) -- (2.9, 0.5) node[anchor=east] {\textcolor{gray}{osvm}}; 
  \draw (4.7,0) -- (4.7,0.7999999999999999) -- (2.9, 0.7999999999999999) node[anchor=east] {vaek}; 
  \draw (4.8,0) -- (4.8,1.0999999999999999) -- (2.9, 1.0999999999999999) node[anchor=east] {\textcolor{gray}{knn}}; 
  \draw (4.8,0) -- (4.8,1.4) -- (2.9, 1.4) node[anchor=east] {vaeo}; 
  \draw (5.2,0) -- (5.2,1.6999999999999997) -- (2.9, 1.6999999999999997) node[anchor=east] {gano}; 
  \draw (5.4,0) -- (5.4,1.7) -- (10.1, 1.7) node[anchor=west] {aae}; 
  \draw (6.6,0) -- (6.6,1.4) -- (10.1, 1.4) node[anchor=west] {fano}; 
  \draw (7.0,0) -- (7.0,1.0999999999999999) -- (10.1, 1.0999999999999999) node[anchor=west] {vae}; 
  \draw (7.2,0) -- (7.2,0.8) -- (10.1, 0.8) node[anchor=west] {wae}; 
  \draw (8.8,0) -- (8.8,0.5) -- (10.1, 0.5) node[anchor=west] {skip}; 
  \draw (9.4,0) -- (9.4,0.2) -- (10.1, 0.2) node[anchor=west] {fmgn}; 
  \draw[line width=0.06cm,color=black,draw opacity=1.0] (3.37,0.1) -- (6.63,0.1); 
  \draw[line width=0.06cm,color=black,draw opacity=1.0] (4.569999999999999,0.3) -- (7.23,0.3); 
  \draw[line width=0.06cm,color=black,draw opacity=1.0] (5.37,0.4) -- (8.83,0.4); 
  \draw[line width=0.06cm,color=black,draw opacity=1.0] (6.569999999999999,0.2) -- (9.43,0.2); 
  \node[anchor=center] at (6.5,-0.7) {average model rank};
 \end{tikzpicture} } \\
    e) semantic image  datasets, anomaly validation & f) semantic image datasets, clean validation \\
    \end{tabular}
 \caption{Critical difference diagram of models ranked via the test AUC. Models whose performance is statistically indistinguishable have difference of ranks under the critical value of the Nemenyi test $CD_{0.1}$ and are joined by a horizontal band. Results are presented for different types of datasets: tabular (Top row), image datasets with statistical anomalies (Middle row), and  image datasets with semantic  anomalies (Bottom row); and two different hyperparameter selectios cases: using anomalies in validation (left) and using clean validation (right).
 }
 \label{fig:critical_diag}
\end{figure*}
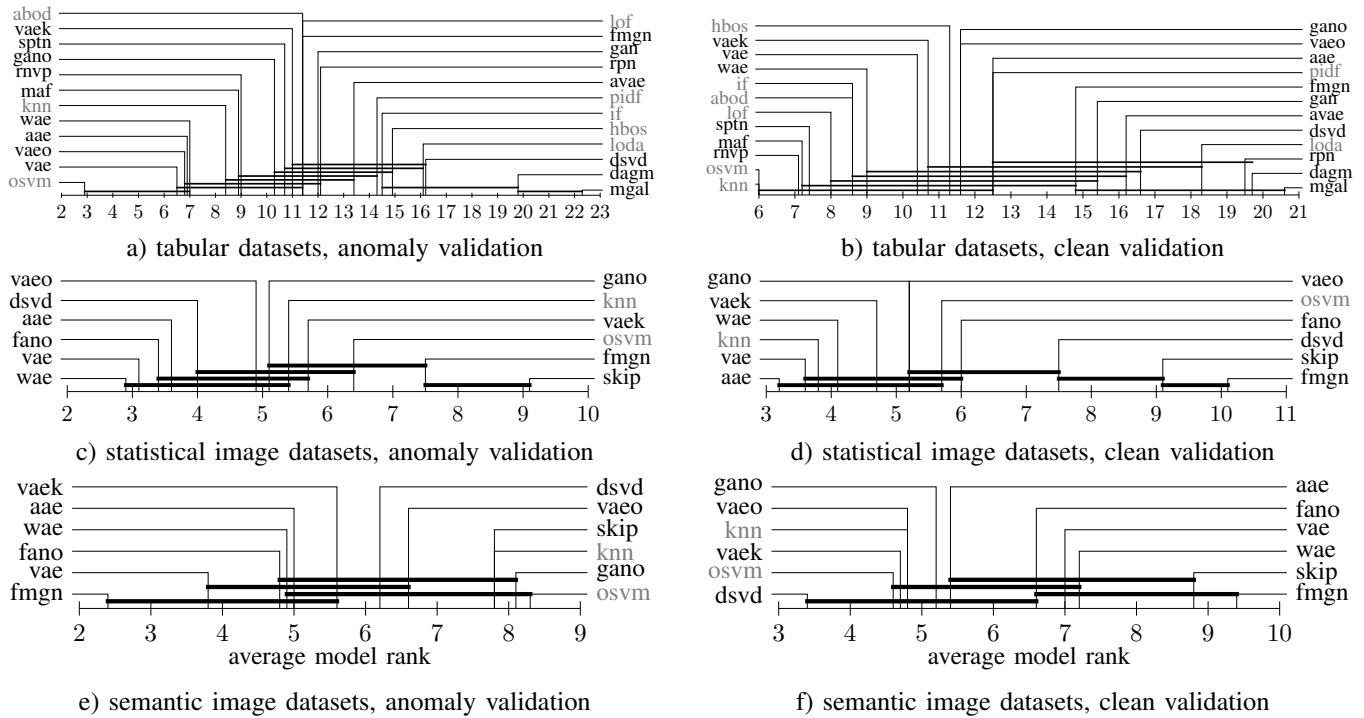



\subsection{Dataset context}
\label{sec:dataset_context}
The results of experimental comparison on all dataset types are presented in the form of critical difference diagrams (CDD) as recommended by Dem\v{s}ar~\citep{demvsar2006statistical}, are in Fig.~\ref{fig:critical_diag}. Diagrams show the average rank of detectors across the datasets together with a confidence band that indicates that a statistical test cannot reject the hypothesis that two detectors perform the same. The underlying  AUC values on the testing set for all individual datasets are given in Tab.~\ref{tab:tabular_anomalies}, \ref{tab:tabular_clean}, \ref{tab:images_stat_auc_auc_combined}, and~\ref{tab:images_semantic_auc_auc_combined}. We now comment on the influence of the datatype with respect to two types of hyper-parameter selection strategies differing in the number of anomalies in the validation set as defined in Section~\ref{sec:contexts}: i) anomaly validation context, and ii) clean validation context.

\emph{Tabular data:} OC-SVM works the best and it is \emph{statistically better} than almost all detectors except autoencoder-based generative models and VAE combined with OC-SVM in the case of anomaly validation context. The first 11 places (roughly one half) belong to models that can be divided into three groups: (i) OC-SVM and its variants, which estimate a density level of a distribution; (ii) flow models and kNN which estimate the pdf (un-normalized in case of kNN); (iii) and variants of auto-encoders, where reconstruction error is related to pdf as explained in~\citep{vsmidl2019anomaly}. The same types of methods occupy the top positions in the clean validation context, Fig.~\ref{fig:critical_diag}b), however in a different order. The best is the kNN and all other pdf-modelling methods (flows) have improved relative to the anomaly validation context. The autoencoder-based methods moved beyond classical methods (LOF, ABOD, IF). We believe that models that are in the lower half of the scale in both validation contexts are not suitable for detection of statistical anomalies. We cannot explain the poor performance of MOGAAL, DAGMM and adVAE\footnote{We have contacted the author of pyOD from wherein we took the implementation of MOGAAL and we were assured that his implementation is a copy of that provided by the authors. Therefore, we consider the implementation to be correct.} and we attribute it to different experimental environment. DeepSVDD was primarily implemented for image problems, where it performs relatively.

Moreover, differences in mean ranks of many models in Fig.~\ref{fig:critical_diag} are statistically insignificant at level $p=0.1$, which is disappointing. Assuming the ranks remain the same, another 51 datasets would be needed to make the difference between OC-SVM and VAE statistically significant on tabular data with 50\% anomalies. This indicates that the results are still noisy and can be easily changed for a different choice of datasets.

\emph{Statistical image data:} WAE and VAE models have the best average rank when evaluated on statistical image data, although their lead is not statistically significant over most of the other models as is evident from Fig.~\ref{fig:critical_diag}c). The autoencoder-based methods (AAE,VAE,WAE) perform well also in the clean validation context, complemented by kNN, Fig.~\ref{fig:critical_diag}d).

\emph{Semantic image data}
A different story is being told by Fig.~\ref{fig:critical_diag}e) where the ranking of methods on image datasets with semantic anomalies is dominated by fmGAN by a large margin in the anomaly validation context. However, it is also the worst method in the clean validation context. In an opposite manner, 
OC-SVM and kNN perform very poorly in the anomaly validation context, but they are among the best in the clean validation context.  The best performing method in the clean validation context is DeepSVDD~\citep{ruff2018deep}. We conjecture that with increasing number of anomalies in the validation set, the problem is approaching that of supervised classification with hyper-parameters playing the role of parameters.

The typical anomalies detected by various methods on image datasets are provided in the Supplementary \ref{sec:appendix_extending_image_results}.

\subsection{Hyperparameter selection context}
\label{sec:hyperparameter_context}
The influence of hyperparameter selection procedure on the results in the previous section is now studied in detail for few selected methods. We choose only those that scored among the best in the previous Section. First, we analyze the sensitivity of these methods to the number of anomalies in the validation set. Second, we study hyperparameter selection for two individual methods, variational autoencoder family and OC-SVM.
\subsubsection{Impact of the number of anomalies in the validation set}

\begin{figure*}[hbt!]
    \centering
    \begin{tikzpicture}[]
\begin{groupplot}[group style={vertical sep = 0.5cm, horizontal sep = 1.0cm, group size=3 by 1}]

\nextgroupplot [
  ylabel = {avg. AUC},
  width=5cm, height=7cm, scale only axis=true, 
  xtick={1,2,3,4,5,6,7,8}, 
  xticklabels={clean,$PR@\%0.01$,$PR@\%0.1$,$PR@\%1$,$PR@\%5$,$PR@\%10$,$PR@\%20$,$AUC_{val}$},
  width=5cm, height=7cm, scale only axis=true,
  x tick label style={rotate=50,anchor=east},
  title = {(tabular)},
]

\addplot+ coordinates {
  (1.0, 0.72375)
  (2.0, 0.762)
  (3.0, 0.754)
  (4.0, 0.8047500000000001)
  (5.0, 0.83925)
  (6.0, 0.844)
  (7.0, 0.8562500000000002)
  (8.0, 0.8845000000000001)
};

\addplot+ coordinates {
  (1.0, 0.607)
  (2.0, 0.52475)
  (3.0, 0.5587500000000001)
  (4.0, 0.5867500000000001)
  (5.0, 0.6585)
  (6.0, 0.71375)
  (7.0, 0.7232500000000001)
  (8.0, 0.7502500000000001)
};

\addplot+ coordinates {};

\addplot+ coordinates {
  (1.0, 0.6375)
  (2.0, 0.66375)
  (3.0, 0.6797500000000001)
  (4.0, 0.6940000000000001)
  (5.0, 0.75325)
  (6.0, 0.7837500000000001)
  (7.0, 0.8067500000000001)
  (8.0, 0.8380000000000001)
};

\addplot+ coordinates {
  (1.0, 0.8087500000000002)
  (2.0, 0.8290000000000001)
  (3.0, 0.8310000000000001)
  (4.0, 0.8329999999999999)
  (5.0, 0.8362499999999999)
  (6.0, 0.8357499999999998)
  (7.0, 0.8454999999999998)
  (8.0, 0.8517499999999998)
};

\addplot+ coordinates {
  (1.0, 0.8145)
  (2.0, 0.7222500000000001)
  (3.0, 0.7222500000000001)
  (4.0, 0.77225)
  (5.0, 0.8247500000000001)
  (6.0, 0.8727499999999999)
  (7.0, 0.8854999999999998)
  (8.0, 0.9112500000000001)
};

\addplot+ coordinates {
  (1.0, 0.7329999999999999)
  (2.0, 0.7945)
  (3.0, 0.7665)
  (4.0, 0.7812500000000001)
  (5.0, 0.8154999999999999)
  (6.0, 0.8247499999999999)
  (7.0, 0.8310000000000001)
  (8.0, 0.8714999999999999)
};

\addplot+ coordinates {
  (1.0, 0.7372500000000001)
  (2.0, 0.5530000000000002)
  (3.0, 0.5897500000000002)
  (4.0, 0.6622499999999999)
  (5.0, 0.7882499999999999)
  (6.0, 0.79475)
  (7.0, 0.84175)
  (8.0, 0.8825)
};

\addplot+ coordinates {
  (1.0, 0.7842499999999999)
  (2.0, 0.7955000000000002)
  (3.0, 0.7945)
  (4.0, 0.79525)
  (5.0, 0.826)
  (6.0, 0.836)
  (7.0, 0.853)
  (8.0, 0.8647499999999999)
};

\addplot+ coordinates {
  (1.0, 0.7967500000000001)
  (2.0, 0.7994999999999999)
  (3.0, 0.8039999999999999)
  (4.0, 0.7997500000000002)
  (5.0, 0.8217500000000001)
  (6.0, 0.82675)
  (7.0, 0.8344999999999999)
  (8.0, 0.8482500000000002)
};


\nextgroupplot [
  legend columns = -1,
  legend style = {at={(0.5,1.2)}, anchor=center},
  legend entries = {aae, dsvd, fano, fmgn, knn, osvm, vae, vaeo, wae, rnvp},
  width=5cm, height=7cm, scale only axis=true, 
  xtick={1,2,3,4,5,6,7,8}, 
  xticklabels={clean,$PR@\%0.01$,$PR@\%0.1$,$PR@\%1$,$PR@\%5$,$PR@\%10$,$PR@\%20$,$AUC_{val}$},
  width=5cm, height=7cm, scale only axis=true,
  x tick label style={rotate=50,anchor=east},
  title = {(statistical)},
]

\addplot+ coordinates {
  (1.0, 0.8664864864864865)
  (2.0, 0.867837837837838)
  (3.0, 0.8791891891891892)
  (4.0, 0.8999999999999999)
  (5.0, 0.8981081081081083)
  (6.0, 0.8978378378378378)
  (7.0, 0.901891891891892)
  (8.0, 0.908918918918919)
};

\addplot+ coordinates {
  (1.0, 0.654054054054054)
  (2.0, 0.7918918918918918)
  (3.0, 0.8108108108108109)
  (4.0, 0.8327027027027026)
  (5.0, 0.841891891891892)
  (6.0, 0.8543243243243243)
  (7.0, 0.8597297297297297)
  (8.0, 0.8802702702702703)
};

\addplot+ coordinates {
  (1.0, 0.841891891891892)
  (2.0, 0.8527027027027027)
  (3.0, 0.8589189189189189)
  (4.0, 0.8727027027027027)
  (5.0, 0.8891891891891893)
  (6.0, 0.8959459459459459)
  (7.0, 0.9045945945945946)
  (8.0, 0.9275675675675675)
};

\addplot+ coordinates {
  (1.0, 0.5199999999999999)
  (2.0, 0.5445945945945946)
  (3.0, 0.5948648648648649)
  (4.0, 0.6775675675675675)
  (5.0, 0.73)
  (6.0, 0.7454054054054053)
  (7.0, 0.8075675675675675)
  (8.0, 0.8856756756756757)
};

\addplot+ coordinates {
  (1.0, 0.8262162162162162)
  (2.0, 0.8670270270270269)
  (3.0, 0.8708108108108108)
  (4.0, 0.8781081081081081)
  (5.0, 0.8781081081081081)
  (6.0, 0.8783783783783784)
  (7.0, 0.8799999999999999)
  (8.0, 0.8856756756756757)
};

\addplot+ coordinates {
  (1.0, 0.7997297297297298)
  (2.0, 0.7021621621621621)
  (3.0, 0.7032432432432432)
  (4.0, 0.7451351351351351)
  (5.0, 0.7983783783783783)
  (6.0, 0.8308108108108109)
  (7.0, 0.8394594594594594)
  (8.0, 0.8824324324324324)
};

\addplot+ coordinates {
  (1.0, 0.8856756756756757)
  (2.0, 0.7954054054054054)
  (3.0, 0.8127027027027027)
  (4.0, 0.828918918918919)
  (5.0, 0.8727027027027027)
  (6.0, 0.897027027027027)
  (7.0, 0.905945945945946)
  (8.0, 0.9205405405405408)
};

\addplot+ coordinates {
  (1.0, 0.8164864864864865)
  (2.0, 0.687027027027027)
  (3.0, 0.6967567567567567)
  (4.0, 0.7478378378378379)
  (5.0, 0.851081081081081)
  (6.0, 0.8794594594594595)
  (7.0, 0.9013513513513514)
  (8.0, 0.929189189189189)
};

\addplot+ coordinates {
  (1.0, 0.8916216216216215)
  (2.0, 0.8862162162162162)
  (3.0, 0.8986486486486487)
  (4.0, 0.8997297297297298)
  (5.0, 0.9183783783783783)
  (6.0, 0.9205405405405406)
  (7.0, 0.9224324324324323)
  (8.0, 0.9289189189189189)
};

\addlegendimage{red,densely dashed,every mark/.append style={solid,fill=red!80!black},mark=diamond*} 


\nextgroupplot [
  width=5cm, height=7cm, scale only axis=true, 
  xtick={1,2,3,4,5,6,7,8}, 
  xticklabels={clean,$PR@\%0.01$,$PR@\%0.1$,$PR@\%1$,$PR@\%5$,$PR@\%10$,$PR@\%20$,$AUC_{val}$},
  width=5cm, height=7cm, scale only axis=true,
  x tick label style={rotate=50,anchor=east},
  title = {(semantic)},
]

\addplot+ coordinates {
  (1.0, 0.5649999999999998)
  (2.0, 0.584)
  (3.0, 0.5745000000000001)
  (4.0, 0.5934999999999999)
  (5.0, 0.6)
  (6.0, 0.6079999999999999)
  (7.0, 0.614)
  (8.0, 0.6224999999999999)
};

\addplot+ coordinates {
  (1.0, 0.5920000000000001)
  (2.0, 0.575)
  (3.0, 0.5685)
  (4.0, 0.5975000000000001)
  (5.0, 0.6250000000000001)
  (6.0, 0.6320000000000001)
  (7.0, 0.6325000000000001)
  (8.0, 0.638)
};

\addplot+ coordinates {
  (1.0, 0.5595000000000001)
  (2.0, 0.541)
  (3.0, 0.5544999999999999)
  (4.0, 0.5899999999999999)
  (5.0, 0.5955)
  (6.0, 0.61)
  (7.0, 0.633)
  (8.0, 0.6424999999999998)
};

\addplot+ coordinates {
  (1.0, 0.5)
  (2.0, 0.4885)
  (3.0, 0.49799999999999994)
  (4.0, 0.6194999999999999)
  (5.0, 0.6765)
  (6.0, 0.7)
  (7.0, 0.7249999999999999)
  (8.0, 0.729)
};

\addplot+ coordinates {
  (1.0, 0.5730000000000001)
  (2.0, 0.5834999999999999)
  (3.0, 0.5824999999999999)
  (4.0, 0.5890000000000001)
  (5.0, 0.5945)
  (6.0, 0.5985)
  (7.0, 0.599)
  (8.0, 0.6015)
};

\addplot+ coordinates {
  (1.0, 0.5760000000000001)
  (2.0, 0.5090000000000001)
  (3.0, 0.536)
  (4.0, 0.5625)
  (5.0, 0.5860000000000001)
  (6.0, 0.5925)
  (7.0, 0.5975000000000001)
  (8.0, 0.6045)
};

\addplot+ coordinates {
  (1.0, 0.552)
  (2.0, 0.5469999999999999)
  (3.0, 0.5625)
  (4.0, 0.5995000000000001)
  (5.0, 0.6210000000000002)
  (6.0, 0.6315000000000001)
  (7.0, 0.6315)
  (8.0, 0.6439999999999999)
};

\addplot+ coordinates {
  (1.0, 0.5735000000000001)
  (2.0, 0.558)
  (3.0, 0.5574999999999999)
  (4.0, 0.574)
  (5.0, 0.588)
  (6.0, 0.5994999999999999)
  (7.0, 0.6035000000000001)
  (8.0, 0.619)
};

\addplot+ coordinates {
  (1.0, 0.549)
  (2.0, 0.5820000000000002)
  (3.0, 0.5705)
  (4.0, 0.6065)
  (5.0, 0.614)
  (6.0, 0.615)
  (7.0, 0.625)
  (8.0, 0.6305)
};


\end{groupplot}

\end{tikzpicture}




    \caption{Sensitivity of methods to the number of anomalies available in the validation set for hyperparameter selection visualized in terms of the achieved AUC aggregated over all datasets in each category (columns). The clean validation context is the left-most point on the x-axis, and the anomaly validation context (50\% of available anomalies) is the right-most point. The points in-between were obtained by selecting models with highest precision on the reported portion (e.g. 5\%) of validation samples with the highest anomaly scores.} 
    \label{fig:combined_knowledge_rank_patn_auc_repre}
\end{figure*}
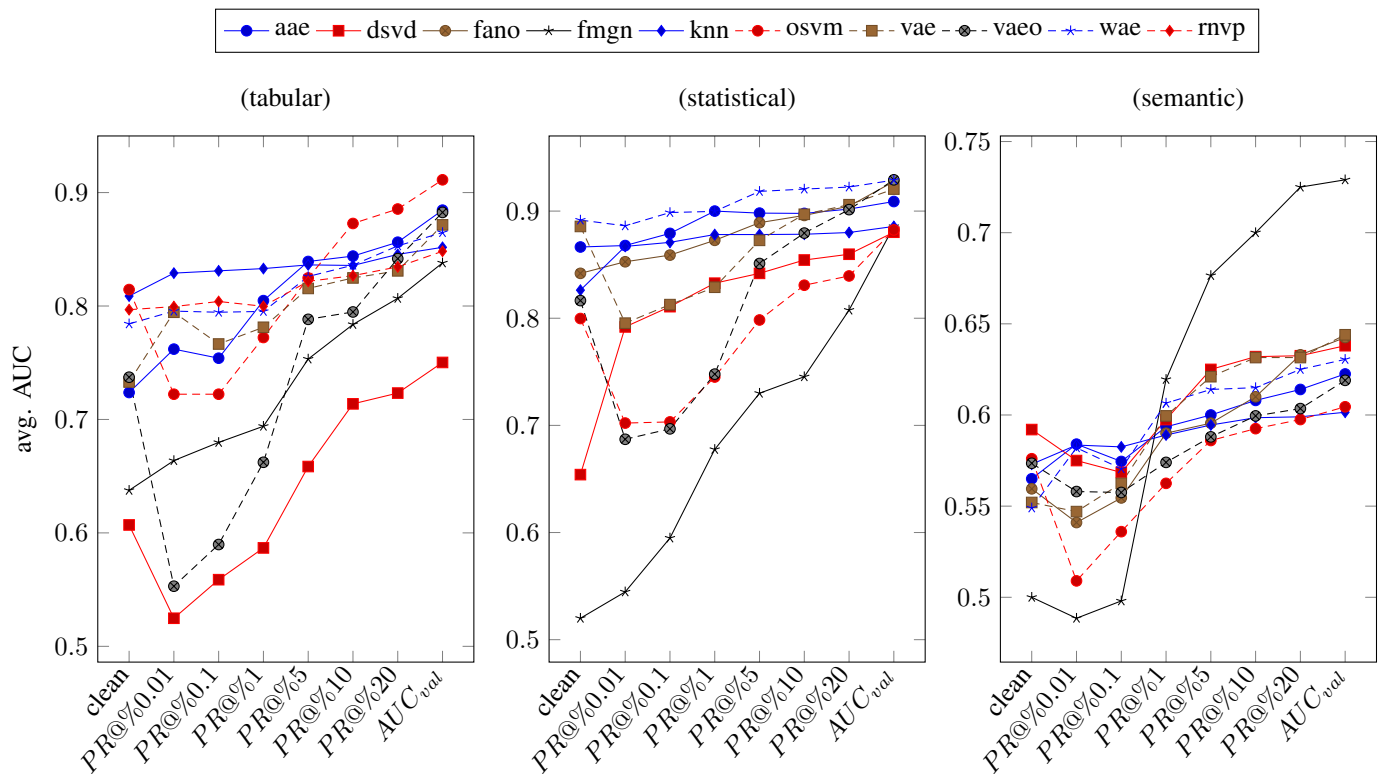

The process of hyperparameter selection described in Sec.~\ref{sub:hyperparameteroptimization} depends on the availability of examples of anomalies in the validation set (recall that it is assumed that the validation dataset does not contain unknown anomalous samples, i.e. is not contaminated). 
Fig.~\ref{fig:combined_knowledge_rank_patn_auc_repre} displays the influence of the number of anomalous samples in the validation set on a finer grid between the two contexts reported before.  Note that for the first point on the x-axis, \textit{clean}, the mechanism of model selection was different from the rest of the graph, as noted in Sec.~\ref{sub:hyperparameteroptimization}. This is the reason for the significant difference between the clean context and the remaining points.

First, we observe that the quality of the models selected using anomalies improves with an increasing number of anomalous samples which is expected. However, for low number of anomalies, many methods perform significantly worse than in the case of clean validation context. This behavior is notable across dataset types, especially for OC-SVM, and to some extent VAE. We conjecture that the hyper-parameter selection procedure of those methods has a tendency to overfit and its hyperparameters are not robust. In contrast to this, the performance of kNN, WAE and RealNVP degrades slowly with declining number of anomalies, which suggests that they are quite robust in difficult operating conditions. We attribute it to the fact that these methods are more exact in their estimation of data likelihood than the rest. 

Second, we notice that the experimental results on the semantic image datasets are generally poor, as the AUC of the best model (fmGAN) on CIFAR10 is 0.72 and similarly on SVHN2, where the best model achieved $0.74$. On the other hand, anomaly detection methods perform well on statistical image datasets. This indicates, contrary to popular belief, that the models fail to learn or identify the important semantic information, or they consider different semantic information anomalous and they should be told \emph{which} semantic aspect of an image should be considered as an anomaly, as for example blurred images might be anomalous as well.

Results of the same study for individual image datasets are presented in the Supplementary, Table~\ref{fig:image_knowledge_rank_pat_auc}. We also provide an illustration of what images were identified as normal and anomalous for the tested methods, Supplementary~\ref{sec:appendix_extending_image_results}.


A practitioner might also desire a method robust with respect to poor choice of hyper-parameters.  In general, deep methods in our experiments have demonstrated higher variance, probably due to the large number of hyperparameters and stochasticity involved in their initialization and training via batched gradient optimization. In this respect, GAN-based models seem to be the least robust, which is in line with~\citep{deecke2018image} stating that GANs are not directly optimized for anomaly detection. This hints at the potential cost of hyperparameter optimization --- with higher performance variance, one is less likely to train a well-performing model in a given number of attempts. 

\subsubsection{Sensitivity Analysis of the VAE family}
\label{sec:vae_results}
Autoencoder-based methods form a whole family with multiple sources of variability, as identified in Sec.~\ref{sec:ae_theory} to be: i) approximation of the likelihood in training (loss function), ii) the richness of latent prior, and iii) the anomaly score. We will analyze the sensitivity of the results to these choices on tabular data in the anomaly validation context. We focus on this family since most of the novel deep generative models for anomaly detection are based on the autoencoder architecture. Additional degrees of freedom include the parametrization of variance of $p_{\vec{\theta}}(\vec{x}|\vec{z}),$ which could be either fixed (called VAE-constant), used in~\citep{yaoUnsupervisedAnomalyDetection2019, wang2020advae, ahnDeepGenerativeModelsBased2020}, scalar (called VAE-scalar), or full diagonal (called VAE-diagonal), used in~\citep{an2015variational,xu2018unsupervised, zenatiEfficientGANBasedAnomaly2018}. In the experiments, all three variations were tested on tabular data, however on image data, the full diagonal was skipped due to computational constraints (and in line with the prior art, where only fixed variance is used).

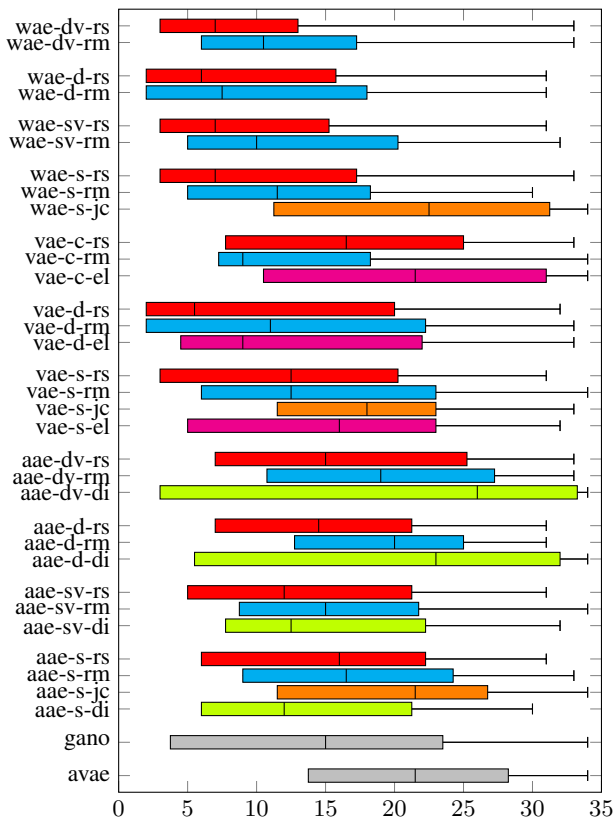
\begin{figure}
    \centering
    \small
    \begin{tikzpicture}
\begin{axis}[
	height=12cm,
	width=8cm,
	ymin=0, ymax=47,
	xmin=0, xmax=35,
	ytick={1,3,5,6,7,8,10,11,12,14,15,16,18,19,20,22,23,24,25,27,28,29,31,32,33,35,36,37,39,40,42,43,45,46},
	yticklabels={avae,gano,aae-s-di,aae-s-jc,aae-s-rm,aae-s-rs,aae-sv-di,aae-sv-rm,aae-sv-rs,aae-d-di,aae-d-rm,aae-d-rs,aae-dv-di,aae-dv-rm,aae-dv-rs,vae-s-el,vae-s-jc,vae-s-rm,vae-s-rs,vae-d-el,vae-d-rm,vae-d-rs,vae-c-el,vae-c-rm,vae-c-rs,wae-s-jc,wae-s-rm,wae-s-rs,wae-sv-rm,wae-sv-rs,wae-d-rm,wae-d-rs,wae-dv-rm,wae-dv-rs}
	]
\addplot [fill=lightgray, boxplot prepared={
draw position=1,
lower whisker=34, lower quartile=13.75,
median=21.5, upper quartile=28.25,
upper whisker=34},
] coordinates {};
\addplot [fill=lightgray, boxplot prepared={
draw position=3,
lower whisker=34, lower quartile=3.75,
median=15.0, upper quartile=23.5,
upper whisker=34},
] coordinates {};
\addplot [fill=lime, boxplot prepared={
draw position=5,
lower whisker=30, lower quartile=6.0,
median=12.0, upper quartile=21.25,
upper whisker=30},
] coordinates {};
\addplot [fill=orange, boxplot prepared={
draw position=6,
lower whisker=34, lower quartile=11.5,
median=21.5, upper quartile=26.75,
upper whisker=34},
] coordinates {};
\addplot [fill=cyan, boxplot prepared={
draw position=7,
lower whisker=33, lower quartile=9.0,
median=16.5, upper quartile=24.25,
upper whisker=33},
] coordinates {};
\addplot [fill=red, boxplot prepared={
draw position=8,
lower whisker=31, lower quartile=6.0,
median=16.0, upper quartile=22.25,
upper whisker=31},
] coordinates {};
\addplot [fill=lime, boxplot prepared={
draw position=10,
lower whisker=32, lower quartile=7.75,
median=12.5, upper quartile=22.25,
upper whisker=32},
] coordinates {};
\addplot [fill=cyan, boxplot prepared={
draw position=11,
lower whisker=34, lower quartile=8.75,
median=15.0, upper quartile=21.75,
upper whisker=34},
] coordinates {};
\addplot [fill=red, boxplot prepared={
draw position=12,
lower whisker=31, lower quartile=5.0,
median=12.0, upper quartile=21.25,
upper whisker=31},
] coordinates {};
\addplot [fill=lime, boxplot prepared={
draw position=14,
lower whisker=34, lower quartile=5.5,
median=23.0, upper quartile=32.0,
upper whisker=34},
] coordinates {};
\addplot [fill=cyan, boxplot prepared={
draw position=15,
lower whisker=31, lower quartile=12.75,
median=20.0, upper quartile=25.0,
upper whisker=31},
] coordinates {};
\addplot [fill=red, boxplot prepared={
draw position=16,
lower whisker=31, lower quartile=7.0,
median=14.5, upper quartile=21.25,
upper whisker=31},
] coordinates {};
\addplot [fill=lime, boxplot prepared={
draw position=18,
lower whisker=34, lower quartile=3.0,
median=26.0, upper quartile=33.25,
upper whisker=34},
] coordinates {};
\addplot [fill=cyan, boxplot prepared={
draw position=19,
lower whisker=33, lower quartile=10.75,
median=19.0, upper quartile=27.25,
upper whisker=33},
] coordinates {};
\addplot [fill=red, boxplot prepared={
draw position=20,
lower whisker=33, lower quartile=7.0,
median=15.0, upper quartile=25.25,
upper whisker=33},
] coordinates {};
\addplot [fill=magenta, boxplot prepared={
draw position=22,
lower whisker=32, lower quartile=5.0,
median=16.0, upper quartile=23.0,
upper whisker=32},
] coordinates {};
\addplot [fill=orange, boxplot prepared={
draw position=23,
lower whisker=33, lower quartile=11.5,
median=18.0, upper quartile=23.0,
upper whisker=33},
] coordinates {};
\addplot [fill=cyan, boxplot prepared={
draw position=24,
lower whisker=34, lower quartile=6.0,
median=12.5, upper quartile=23.0,
upper whisker=34},
] coordinates {};
\addplot [fill=red, boxplot prepared={
draw position=25,
lower whisker=31, lower quartile=3.0,
median=12.5, upper quartile=20.25,
upper whisker=31},
] coordinates {};
\addplot [fill=magenta, boxplot prepared={
draw position=27,
lower whisker=33, lower quartile=4.5,
median=9.0, upper quartile=22.0,
upper whisker=33},
] coordinates {};
\addplot [fill=cyan, boxplot prepared={
draw position=28,
lower whisker=33, lower quartile=2.0,
median=11.0, upper quartile=22.25,
upper whisker=33},
] coordinates {};
\addplot [fill=red, boxplot prepared={
draw position=29,
lower whisker=32, lower quartile=2.0,
median=5.5, upper quartile=20.0,
upper whisker=32},
] coordinates {};
\addplot [fill=magenta, boxplot prepared={
draw position=31,
lower whisker=34, lower quartile=10.5,
median=21.5, upper quartile=31.0,
upper whisker=34},
] coordinates {};
\addplot [fill=cyan, boxplot prepared={
draw position=32,
lower whisker=34, lower quartile=7.25,
median=9.0, upper quartile=18.25,
upper whisker=34},
] coordinates {};
\addplot [fill=red, boxplot prepared={
draw position=33,
lower whisker=33, lower quartile=7.75,
median=16.5, upper quartile=25.0,
upper whisker=33},
] coordinates {};
\addplot [fill=orange, boxplot prepared={
draw position=35,
lower whisker=34, lower quartile=11.25,
median=22.5, upper quartile=31.25,
upper whisker=34},
] coordinates {};
\addplot [fill=cyan, boxplot prepared={
draw position=36,
lower whisker=30, lower quartile=5.0,
median=11.5, upper quartile=18.25,
upper whisker=30},
] coordinates {};
\addplot [fill=red, boxplot prepared={
draw position=37,
lower whisker=33, lower quartile=3.0,
median=7.0, upper quartile=17.25,
upper whisker=33},
] coordinates {};
\addplot [fill=cyan, boxplot prepared={
draw position=39,
lower whisker=32, lower quartile=5.0,
median=10.0, upper quartile=20.25,
upper whisker=32},
] coordinates {};
\addplot [fill=red, boxplot prepared={
draw position=40,
lower whisker=31, lower quartile=3.0,
median=7.0, upper quartile=15.25,
upper whisker=31},
] coordinates {};
\addplot [fill=cyan, boxplot prepared={
draw position=42,
lower whisker=31, lower quartile=2.0,
median=7.5, upper quartile=18.0,
upper whisker=31},
] coordinates {};
\addplot [fill=red, boxplot prepared={
draw position=43,
lower whisker=31, lower quartile=2.0,
median=6.0, upper quartile=15.75,
upper whisker=31},
] coordinates {};
\addplot [fill=cyan, boxplot prepared={
draw position=45,
lower whisker=33, lower quartile=6.0,
median=10.5, upper quartile=17.25,
upper whisker=33},
] coordinates {};
\addplot [fill=red, boxplot prepared={
draw position=46,
lower whisker=33, lower quartile=3.0,
median=7.0, upper quartile=13.0,
upper whisker=33},
] coordinates {};
\end{axis}
\end{tikzpicture}
    \caption{Sensitivity study of various variants of  autoencoder-based methods displayed in the form of boxplots of their ranks in the AUC metric achieved on the tabular datasets. The first three letters of the method's name denote the training loss. Models with the -d- middle part estimate full diagonal of the decoder variance, -s- estimate only a scalar, and -c-  use a fixed scalar variance as a hyperparameter. All variants are using the standard Gaussian latent model. Models using the VampPrior are denoted by extending the decoder variance symbol by the letter v-, i.e. -dv-, -sv-, -cv-. The last part of the name denotes score, -rs stands for the sampled rec. probability~\eqref{eq:score_sample} with $L=100$, -rm for~\eqref{eq:score_mean}, -el for the ELBO~\eqref{eq:vae_loss} composed of -rs and KLD, -jc for~\eqref{eq:jacodeco}, -di for~\eqref{eq:disc_score}.}
    \label{fig:tabular_ae_only_box_auc_meanmax}
\end{figure}

The overall comparison in Fig.~\ref{fig:critical_diag} revealed that WAE and vanilla VAE variants perform best. The other degrees of freedom, namely richness of prior, used anomaly score and parametrization of variance were treated as hyperparameters. Fig.~\ref{fig:tabular_ae_only_box_auc_meanmax} extends the study by showing the distribution of ranks over tabular datasets for different variants of VAE including GANomaly and adVAE.

First, notice that the spread of the method's ranks over various datasets is significant, as even ranks of the best methods vary from 3 to 15. This means that the conclusions below need to be taken with a grain of salt, as the experimental results are extremely noisy.

The ELBO-based score,  -el, together with the orthogonal decomposition of the likelihood~\citep{pidhorskyi2018generative}, -jc, does not perform well. The sampled reconstruction error (an MC estimate of~\eqref{eq:score_sample}, \mbox{-rs}, almost always performs better than the usual reconstruction error, -rm, calculated according to~\eqref{eq:score_mean}. This demonstrates the common approach of replacing the mean of the decoder with that of the encoder is inferior but computationally cheaper (see Tab.~\ref{tab:predict_times} with prediction times). The discriminator score~\eqref{eq:disc_score}, -di, of AAE (an autoencoder combined with GAN) seems to be also on par with the MC estimate~\eqref{eq:score_sample}. 

From the same figure, we also conclude that the models modelling full diagonal in $p_{\vec{\theta}}(\vec{x}|\vec{z})$, -d-, seem to be better than the scalar, -s-, or constant, -c-, variants. This result is important, as many comparisons in the prior art use the VAE-constant, despite the version with full diagonal being discussed in the original publication~\citep{kingma2013auto}.

The rich prior distribution on the latent space proposed in \citep{tomczak2018vae}, VAMP, -v-, does not seem to give an advantage in the anomaly detection except in the AAE. Similarly, recent variants adVAE and GANomaly do not seem to work well on the tabular data, but they were not evaluated on them in the original publications.

\begin{table}
    \footnotesize
    \centering
    \tabcolsep=0.05cm
    \begin{tabular}{c|c c c c c}
         & vae-s-rs & vae-d-rs & vae-s-rm & vae-d-rm & vae-d-jc \\
         \midrule
        $\bar{t}_{pred}$ [s] & 12.10 & 18.51 & 0.11 & 0.15 & 57.31
    \end{tabular}
    \vspace*{0.15cm}
    \caption{Average prediction times on the  tabular datasets for different combinations of VAE scores and decoder variance estimations. The -d- part stands for model with an estimate of the full diagonal of the decoder variance, -s- is a scalar estimate. Sampled reconstruction error~\eqref{eq:score_sample} is denoted as -rs, -rm is the anomaly score~\eqref{eq:score_mean} and -jc is~\eqref{eq:jacodeco}.}
    \label{tab:predict_times}
\end{table}

\subsubsection{Sensitivity study of OC-SVM}
\label{sub:OC-SVM}
Domination of OC-SVM on tabular data in anomaly validation context contrasts to many prior experimental comparisons~\citep{goldstein2016comparative, chalapathyGroupAnomalyDetection2018, deecke2018image, gopalanPIDForestAnomalyDetection2019, iwataSupervisedAnomalyDetection2019, wang2020advae}. The search for the culprit found it to be the hyperparameter selection. This study has varied the $\nu$ parameter, kernels, and their parameters, which is much more than the most prior art does, which is fixing the kernel to RBF and tests few values of its width $\gamma$ and $\nu.$ Inclusion of other kernels into the search for hyperparameters seems to be the major source of improvement in this case. Replacing the OC-SVM with one restricted to use only the RBF kernel and $\nu=0.5$ yields an increase in average rank from $2.9$ to $8.1$ with an average decrease in performance by $0.06$, measured in AUC in the anomaly validation context. This version of OC-SVM is then easily surpassed by variational autoencoders and kNN, as demonstrated in Supplementary, Table \ref{tab:tabular_auc_auc_meanmax_orbf_ranks_only}.  The importance of the choice of kernel is furthermore illustrated by the fact that the sigmoid kernel was the optimal choice for 23 datasets, while the RBF kernel only for 13. Ref.~\citep{goldstein2016comparative} mentions that setting $\nu=0.5$ provides universally good results, which may be the reason why many authors do not tune it. In theory, it should be set to much lower values ($\nu = 0.05$) corresponding to the presumed low ratio of anomalies in data, but with Bayesian optimization, we found that the best estimate of $\nu$ was in some cases even higher, such as $\sim0.75$ on the statlog-vehicle dataset.

\subsection{Economic context}
\label{sec:economic_context}
\begin{figure*}
    \centering
    \begin{subfigure}{\columnwidth}
        \centering
        \small
        \resizebox {\linewidth}{!}{
            \begin{tikzpicture}[]
\begin{groupplot}[group style={vertical sep = 0.0cm, horizontal sep = 0.2cm, group size=2 by 1}]

\nextgroupplot [
  xlabel = {avg. rank},
  ylabel = {avg. training time rank},
  ylabel near ticks, yticklabel pos=left,
  ymin = 0.5, ymax=25,
  width=4cm, height=6cm, scale only axis=true,
  grid=major,
]

\addplot+[
  draw = none
] coordinates {
  (6.9, 11.3)
  (11.4, 8.8)
  (13.4, 12.7)
  (19.8, 16.2)
  (16.2, 10.9)
  (11.4, 18.8)
  (12.0, 16.2)
  (10.3, 9.7)
  (14.9, 4.4)
  (14.5, 5.0)
  (8.4, 2.4)
  (16.1, 3.4)
  (11.4, 2.6)
  (8.9, 21.8)
  (2.9, 3.4)
  (14.3, 15.1)
  (9.0, 23.3)
  (12.1, 7.6)
  (10.7, 23.0)
  (6.5, 12.1)
  (11.0, 16.3)
  (6.8, 14.9)
  (7.0, 20.1)
};

\node at (axis cs:8.2, 12.0) {aae};

\node at (axis cs:13.1, 9.7) {abod};

\node at (axis cs:13.9, 13.3) {avae};

\node at (axis cs:18.9, 16.9) {dagm};

\node at (axis cs:16.7, 11.8) {dsvd};

\node at (axis cs:11.9, 19.5) {fmgn};

\node at (axis cs:13.0, 17.0) {gan};

\node at (axis cs:10.8, 10.3) {gano};

\node at (axis cs:16.8, 5.1) {hbos};

\node at (axis cs:14.1, 5.8) {if};

\node at (axis cs:8.8, 3.2) {knn};

\node at (axis cs:17.8, 2.9) {loda};

\node at (axis cs:11.6, 3.4) {lof};

\node at (axis cs:11.0, 21.8) {maf};


\node at (axis cs:3.4, 4.1) {osvm};

\node at (axis cs:14.8, 15.7) {pidf};

\node at (axis cs:9.1, 24.0) {rnvp};

\node at (axis cs:13.6, 8.0) {rpn};

\node at (axis cs:12.8, 23.7) {sptn};

\node at (axis cs:7.0, 12.7) {vae};

\node at (axis cs:9.3, 17.0) {vaek};

\node at (axis cs:7.3, 15.6) {vaeo};

\node at (axis cs:7.5, 20.8) {wae};

\nextgroupplot [
  xlabel = {avg. rank},
  ylabel = {avg. prediction time rank},
  ylabel near ticks, yticklabel pos=right,
  ymin = 0.5, ymax=25,
  width=4cm, height=6cm, scale only axis=true,
  grid=major,
]

\addplot+[
  draw = none
] coordinates {
  (6.9, 13.3)    
  (11.4, 23.8)    
  (13.4, 21.8)
  (19.8, 1.3)
  (16.2, 7.0)
  (11.4, 2.3)
  (12.0, 2.2)
  (10.3, 5.6)
  (14.9, 4.4)
  (14.5, 16.0)
  (8.4, 15.1)
  (16.1, 10.4)
  (11.4, 11.5)
  (8.9, 12.2)
  (2.9, 10.1)
  (14.3, 22.6)
  (9.0, 8.7)
  (12.1, 10.5)
  (10.7, 17.0)
  (6.5, 18.4)
  (11.0, 18.0)
  (6.8, 14.5)
  (7.0, 19.1)
};

\node at (axis cs:5.5, 13.8) {aae};

\node at (axis cs:11.9, 24.5) {abod};

\node at (axis cs:11.9, 22.4) {avae};

\node at (axis cs:18.3, 2.0) {dagm};

\node at (axis cs:16.7, 7.8) {dsvd};

\node at (axis cs:10.0, 3.0) {fmgn};

\node at (axis cs:13.5, 2.9) {gan};

\node at (axis cs:10.8, 6.3) {gano};

\node at (axis cs:15.4, 5.2) {hbos};

\node at (axis cs:15.1, 16.8) {if};

\node at (axis cs:8.9, 16.0) {knn};

\node at (axis cs:16.6, 11.2) {loda};

\node at (axis cs:12.3, 12.3) {lof};

\node at (axis cs:9.4, 12.8) {maf};


\node at (axis cs:3.4, 10.7) {osvm};

\node at (axis cs:14.8, 23.3) {pidf};

\node at (axis cs:9.5, 9.3) {rnvp};

\node at (axis cs:13.7, 9.8) {rpn};

\node at (axis cs:8.9, 17.5) {sptn};

\node at (axis cs:5.0, 19.0) {vae};

\node at (axis cs:11.5, 18.9) {vaek};

\node at (axis cs:6.0, 15.2) {vaeo};

\node at (axis cs:5.5, 19.8) {wae};

\end{groupplot}
\end{tikzpicture}
        }
        \caption{tabular datasets}
        \label{fig:tabular_total_eval_t_fit_t_combined}
    \end{subfigure}
    \begin{subfigure}{\columnwidth}
        \centering
        \small
        \resizebox {\linewidth}{!}{
            \begin{tikzpicture}[]
\begin{groupplot}[group style={vertical sep = 0.0cm, horizontal sep = 0.2cm, group size=2 by 1}]

\nextgroupplot [
  xlabel = {avg. rank},
  ylabel = {avg. training time rank},
  ylabel near ticks, yticklabel pos=left,
  ymin = 0.5, ymax=13,
  width=4cm, height=6cm, scale only axis=true,
  grid=major,
]

\addplot+[
  draw = none
] coordinates {
  (4.1, 5.0)
  (4.8, 6.5)
  (3.8, 11.8)
  (5.7, 11.2)
  (6.2, 6.8)
  (6.2, 1.6)
  (7.1, 1.8)
  (8.6, 5.2)
  (3.4, 7.3)
  (5.7, 5.4)
  (5.5, 5.6)
  (3.6, 9.9)
};

\node at (axis cs:4.4, 5.5) {aae};

\node at (axis cs:5.0, 7.0) {dsvd};

\node at (axis cs:4.3, 12.3) {fano};

\node at (axis cs:6.2, 11.7) {fmgn};

\node at (axis cs:6.6, 7.3) {gano};

\node at (axis cs:6.0, 2.2) {knn};

\node at (axis cs:7.6, 2.2) {osvm};

\node at (axis cs:8.2, 5.8) {skip};

\node at (axis cs:3.7, 7.8) {vae};

\node at (axis cs:6.5, 5.5) {vaek};

\node at (axis cs:6.1, 5.8) {vaeo};

\node at (axis cs:3.9, 10.4) {wae};

\nextgroupplot [
  xlabel = {avg. rank},
  ylabel = {avg. prediction time rank},
  ylabel near ticks, yticklabel pos=right,
  ymin = 0.5, ymax=13,
  width=4cm, height=6cm, scale only axis=true,
  grid=major,
]

\addplot+[
  draw = none
] coordinates {
  (4.1, 7.6)
  (4.8, 2.4)
  (3.8, 8.8)
  (5.7, 1.0)
  (6.2, 4.8)
  (6.2, 11.1)
  (7.1, 11.3)
  (8.6, 3.8)
  (3.4, 4.6)
  (5.7, 7.7)
  (5.5, 9.6)
  (3.6, 5.5)
};

\node at (axis cs:4.4, 8.1) {aae};

\node at (axis cs:5.3, 3.1) {dsvd};

\node at (axis cs:4.3, 9.5) {fano};

\node at (axis cs:6.2, 1.7) {fmgn};

\node at (axis cs:6.7, 5.3) {gano};

\node at (axis cs:6.0, 11.6) {knn};

\node at (axis cs:7.4, 11.8) {osvm};

\node at (axis cs:8.5, 4.5) {skip};

\node at (axis cs:3.9, 5.0) {vae};

\node at (axis cs:6.2, 8.4) {vaek};

\node at (axis cs:6.0, 10.0) {vaeo};

\node at (axis cs:4.1, 6.0) {wae};

\end{groupplot}
\end{tikzpicture}
        }
        \caption{image datasets}
        \label{fig:images_total_eval_t_fit_t_combined}
    \end{subfigure}
    \caption{Scatter-plots of the average rank in the AUC metric on the tabular  (a) and image (b) data versus average rank of the computational complexity of the displayed methods measured via the the training time (left) and prediction time (right). MO-GAAL has been omitted from the tabular figures, as its performance positioned it too far to the right with the training time rank of $19.4$ and the prediction time rank of $10.0$.}
    \label{fig:images_total_eval_t_fit_t_combined_joined}
\end{figure*}

Practitioners ask for fast and accurate algorithms, but these two features rarely go hand in hand, and a decision on a trade-off has to be made. Interesting methods lie on the Pareto frontier, as in absence of external factor, rationally behaving practitioner does not have a motivation to choose a different model.

Fig.~\ref{fig:tabular_total_eval_t_fit_t_combined}-left shows the trade-off between accuracy and training time for tabular data, where the absolute numbers were replaced by average ranks for robustness. The Pareto frontier contains two methods, which are OC-SVM and kNN. The position of OC-SVM is rather surprising, as its training time is known to scale poorly (quadratically) with respect to the number of samples, but it is caused by most of the tabular datasets being small. Different results may arise for a dataset with many data records. Fig.~\ref{fig:tabular_total_eval_t_fit_t_combined}-right shows a similar trade-off between accuracy and testing (inference) time. OC-SVM is still on the Pareto frontier, but it is expensive, as the complexity grows linearly in the number of samples. fmGAN, GAN and DAGMM methods are there as well -- these methods have fast inference but lower accuracy.

We provide results averaged over the studied contexts on image data. Due to variability of the results in each context, the x-axis will vary. In the averaged ranks, VAE is on the Pareto front in both fit and prediction times, see Fig.~\ref{fig:images_total_eval_t_fit_t_combined}. Its prediction complexity is given mainly by the choice of the number of samples taken in the computation of the sampled reconstruction score~\eqref{eq:score_sample}. The kNN detector has negligible training time, given only by the construction of the tree structure representing data, but seems to be mostly unusable on image data due to slow prediction times on datasets that are large in dimension and number of samples. The fmGAN finds itself in a completely reversed scenario.

\subsection{Other influences}
\label{sec:other_context}
In this section, we report results of three sources of variability of performance of AD methods that were found to have minimal impact.

\emph{Ensembles/Hyperparameter averaging} The benefits of ensembles in prior art seem to be mixed. While~\citep{choiWAICWhyGenerative2019} claims that a combination of VAE or GAN ensembles using WAIC might be useful, Ref.~\citep{nalisnickDeepGenerativeModels2019} claims a negligible effect. In our experiments, we have used ensembles as a way to reduce uncertainty in hyperparameters~\citep{wilson2020case}, meaning that unlike in~\citep{choiWAICWhyGenerative2019}, models in ensembles were of a single type differing only in architecture. The effect of such an ensemble on average AUC was overall zero, sometimes even negative, see Supplementary~\ref{sec:appendix_ensembles}, Table~\ref{tab:ensembles_sensitivity_grouped}. Exceptions are methods based on GANs featuring improvements by $0.02$ in average AUC. These findings are on par with those in~\citep{nalisnickDeepGenerativeModels2019}.

\emph{Bayesian optimization} Bayesian optimization was introduced as an alternative to random search of hyperparameters. It has the potential to find better hyperparameters with a low number of trials. Comparison of the random search and Bayesian optimization under the same conditions is reported in the Supplementary, Table~\ref{tab:tabular_bayes_comparison}. We can conclude that Bayesian optimization was able to find hyperparameters with better performance for the vast majority of the methods. However, this improvement is often quite small and insufficient to change the ranks of the methods. Notable exceptions are the GANomaly and GAN methods which improved by two ranks. 

\emph{Performance metric AUC/TPR} The hypothesis that the methods may perform differently when chosen for different optimization criteria than the usual AUC has not been proved. The results are summarized in the Supplementary, Tab.~\ref{tab:metric_comparison_grouped}. While small changes in performance have been observed, the ranks of the methods remain the same in both criterions, AUC and TPR@5\%. 

\section{Conclusion}
The presented extensive comparison of anomaly detection methods based on deep generative methods, namely variants of flows, variational autoencoders and generative adversarial networks with methods based on alternative paradigms (Support vector machines, random forests, histogram and distance-based methods) revealed that the performance of anomaly detection methods strongly depends on experimental conditions. We have identified the main sources of variation to be the choice of data and hyperparameter tuning. We presented detailed results under a combination of experimental conditions, called contexts, specifically the data context, hyperparameter context and economic context. 

In the dataset context, a clear distinction in performance was found between tabular data, image data containing statistical and semantic anomalies. The main distinction in hyperparameter selection was how many anomalies were present in the validation set. Different order of performance of the tested method was observed for a different combination of dataset type and hyperparameter selection context, explaining various outcomes of comparison in the prior art. We strongly recommend to authors to provide details on the context of their experimental studies in the future.

There are many aspects that have not been covered and remain a topic of future works, such as designing ensembles of methods of various types, or identification of the relevant kind of anomalies in the semantic datasets.

The comparison is not only aimed at researchers, but also at practitioners desiring accurate methods with low computational complexity. Therefore, we visualize these trade-offs between performance and computational cost. Also, we present a list of some of the most important practical observations and recommendations.
\begin{itemize}
    \item Methods with more exact likelihood modeling, such as kNN, flow and autoencoder-based models, perform better in scenarios with a limited number of anomalies available for hyperparameter tuning.
    \item Majority of the methods fail to detect semantic anomalies. The exception is the fmGAN, but only if given enough computational resources and many anomalous samples for cross-validation. 
    \item OC-SVM, when properly tuned, can defeat most of the state-of-the-art on tabular data, although it suffers from overfitting when hyperparameters are selected using too few anomalies in the validation set.
    \item The method of the first choice appears to be the VAE/WAE due to its relatively cheap, precise and consistent performance in most of the experiments. However, it possesses so many degrees of freedom that it forms a full family of methods. It was found that the best performance is obtained when estimating the full variance of samples on the output of the decoder and evaluating anomalies using sampled reconstruction score. 
\end{itemize}

\appendices

\section*{Acknowledgments}
Support of grants GACR GA18-21409S, MSMT CZ.02.1.01/0.0/0.0/16\_019/0000765 and SGS18/188/OHK4/3T/14 is gratefully acknowledged.

\bibliographystyle{plain}
\bibliography{references}

\begin{thebibliography}{10}

\bibitem{ahmed2020detecting}
Faruk Ahmed and Aaron Courville.
\newblock Detecting semantic anomalies.
\newblock In {\em Proceedings of the AAAI Conference on Artificial
  Intelligence}, volume~34, pages 3154--3162, 2020.

\bibitem{ahnDeepGenerativeModelsBased2020}
Hyojung Ahn, Dawoon Jung, and Han-Lim Choi.
\newblock Deep generative models-based anomaly detection for spacecraft control
  systems.
\newblock {\em Sensors}, 20:1991, 04 2020.

\bibitem{akcay2018ganomaly}
Samet Akcay, Amir Atapour-Abarghouei, and Toby~P. Breckon.
\newblock Ganomaly: Semi-supervised anomaly detection via adversarial training.
\newblock In {\em Asian conference on computer vision}, pages 622--637.
  Springer, 2018.

\bibitem{akcay2019skip}
Samet Ak{\c{c}}ay, Amir Atapour-Abarghouei, and Toby~P Breckon.
\newblock Skip-ganomaly: Skip connected and adversarially trained
  encoder-decoder anomaly detection.
\newblock In {\em 2019 International Joint Conference on Neural Networks
  (IJCNN)}, pages 1--8. IEEE, 2019.

\bibitem{an2015variational}
Jinwon An and Sungzoon Cho.
\newblock Variational autoencoder based anomaly detection using reconstruction
  probability.
\newblock {\em Special Lecture on IE}, 2(1):1--18, 2015.

\bibitem{arjovsky2017wasserstein}
Martin Arjovsky, Soumith Chintala, and L\'eon Bottou.
\newblock Wasserstein gan.
\newblock {\em arXiv:1701.07875 [stat]}, 2017.

\bibitem{barnett1974outliers}
Vic Barnett and Toby Lewis.
\newblock {\em Outliers in statistical data}.
\newblock Wiley, 1974.

\bibitem{bergmannMVTecADComprehensive2019}
Paul Bergmann, Michael Fauser, David Sattlegger, and Carsten Steger.
\newblock {{MVTec AD}} \textemdash{} {{A Comprehensive Real}}-{{World Dataset}}
  for {{Unsupervised Anomaly Detection}}.
\newblock page~9, 2019.

\bibitem{bergstra2012random}
James Bergstra and Yoshua Bengio.
\newblock Random search for hyper-parameter optimization.
\newblock {\em The Journal of Machine Learning Research}, 13(1):281--305, 2012.

\bibitem{Julia-2017}
Jeff Bezanson, Alan Edelman, Stefan Karpinski, and Viral~B Shah.
\newblock Julia: A fresh approach to numerical computing.
\newblock {\em SIAM {R}eview}, 59(1):65--98, 2017.

\bibitem{breunig2000lof}
Markus~M Breunig, Hans-Peter Kriegel, Raymond~T Ng, and J{\"o}rg Sander.
\newblock Lof: identifying density-based local outliers.
\newblock In {\em Proceedings of the 2000 ACM SIGMOD international conference
  on Management of data}, pages 93--104, 2000.

\bibitem{campos2016evaluation}
Guilherme~O Campos, Arthur Zimek, J{\"o}rg Sander, Ricardo~JGB Campello,
  Barbora Micenkov{\'a}, Erich Schubert, Ira Assent, and Michael~E Houle.
\newblock On the evaluation of unsupervised outlier detection: measures,
  datasets, and an empirical study.
\newblock {\em Data Mining and Knowledge Discovery}, 30(4):891--927, 2016.

\bibitem{chalapathy2019deep}
Raghavendra Chalapathy and Sanjay Chawla.
\newblock Deep learning for anomaly detection: A survey.
\newblock {\em arXiv:1901.03407 [cs]}, 2019.

\bibitem{chalapathy2018anomaly}
Raghavendra Chalapathy, Aditya~Krishna Menon, and Sanjay Chawla.
\newblock Anomaly detection using one-class neural networks.
\newblock {\em arXiv:1802.06360 [cs]}, 2019.

\bibitem{chalapathyGroupAnomalyDetection2018}
Raghavendra Chalapathy, Edward Toth, and Sanjay Chawla.
\newblock Group {{Anomaly Detection}} using {{Deep Generative Models}}.
\newblock {\em arXiv:1804.04876 [cs]}, April 2018.

\bibitem{chen2018unsupervised}
Xiaoran Chen and Ender Konukoglu.
\newblock Unsupervised detection of lesions in brain {MRI} using constrained
  adversarial auto-encoders.
\newblock {\em arXiv:1806.04972 [cs]}, 2018.

\bibitem{choiWAICWhyGenerative2019}
Hyunsun Choi, Eric Jang, and Alexander~A. Alemi.
\newblock {{WAIC}}, but {{Why}}? {{Generative Ensembles}} for {{Robust Anomaly
  Detection}}.
\newblock {\em arXiv:1810.01392 [cs, stat]}, May 2019.

\bibitem{dai2019diagnosing}
Bin Dai and David Wipf.
\newblock Diagnosing and enhancing vae models.
\newblock {\em arXiv:1903.05789 [cs]}, 2019.

\bibitem{deecke2018image}
Lucas Deecke, Robert Vandermeulen, Lukas Ruff, Stephan Mandt, and Marius Kloft.
\newblock Image anomaly detection with generative adversarial networks.
\newblock In {\em Joint european conference on machine learning and knowledge
  discovery in databases}, pages 3--17. Springer, 2018.

\bibitem{demvsar2006statistical}
Janez Dem{\v{s}}ar.
\newblock Statistical comparisons of classifiers over multiple data sets.
\newblock {\em Journal of Machine learning research}, 7(Jan):1--30, 2006.

\bibitem{diasAnomalyDetectionTrajectory2020a}
Madson L.~D. Dias, C{\'e}sar Lincoln~C. Mattos, Ticiana L.~C. {da Silva},
  Jos{\'e} Ant{\^o}nio~F. {de Macedo}, and Wellington C.~P. Silva.
\newblock Anomaly {{Detection}} in {{Trajectory Data}} with {{Normalizing
  Flows}}.
\newblock {\em arXiv:2004.05958 [cs, stat]}, April 2020.

\bibitem{dinh2016density}
Laurent Dinh, Jascha Sohl-Dickstein, and Samy Bengio.
\newblock Density estimation using {Real NVP}.
\newblock {\em arXiv:1605.08803 [cs]}, 2017.

\bibitem{dodd2003partial}
Lori~E Dodd and Margaret~S Pepe.
\newblock Partial auc estimation and regression.
\newblock {\em Biometrics}, 59(3):614--623, 2003.

\bibitem{donahue2016adversarial}
Jeff Donahue, Philipp Kr{\"a}henb{\"u}hl, and Trevor Darrell.
\newblock Adversarial feature learning.
\newblock {\em arXiv:1605.09782 [cs]}, 2017.

\bibitem{Dua:2019}
Dheeru Dua and Casey Graff.
\newblock {UCI} machine learning repository, 2017.

\bibitem{emmott2013systematic}
Andrew~F. Emmott, Shubhomoy Das, Thomas Dietterich, Alan Fern, and Weng-Keen
  Wong.
\newblock Systematic construction of anomaly detection benchmarks from real
  data.
\newblock In {\em Proceedings of the ACM SIGKDD workshop on outlier detection
  and description}, pages 16--21, 2013.

\bibitem{ergen2017unsupervised}
Tolga Ergen and Suleyman~Serdar Kozat.
\newblock Unsupervised anomaly detection with {LSTM} neural networks.
\newblock {\em IEEE Transactions on Neural Networks and Learning Systems},
  31(8):3127–3141, Aug 2020.

\bibitem{fernandes2019comprehensive}
Gilberto Fernandes, Joel~JPC Rodrigues, Luiz~Fernando Carvalho, Jalal~F
  Al-Muhtadi, and Mario~Lemes Proen{\c{c}}a.
\newblock A comprehensive survey on network anomaly detection.
\newblock {\em Telecommunication Systems}, 70(3):447--489, 2019.

\bibitem{fernandez2014we}
Manuel Fern{\'a}ndez-Delgado, Eva Cernadas, Sen{\'e}n Barro, and Dinani Amorim.
\newblock Do we need hundreds of classifiers to solve real world classification
  problems?
\newblock {\em The journal of machine learning research}, 15(1):3133--3181,
  2014.

\bibitem{germainMADEMaskedAutoencoder2015}
Mathieu Germain, Karol Gregor, Iain Murray, and Hugo Larochelle.
\newblock {{MADE}}: {{Masked Autoencoder}} for {{Distribution Estimation}}.
\newblock {\em arXiv:1502.03509 [cs, stat]}, June 2015.

\bibitem{goldstein2012histogram}
Markus Goldstein and Andreas Dengel.
\newblock Histogram-based outlier score (hbos): A fast unsupervised anomaly
  detection algorithm.
\newblock {\em KI-2012: Poster and Demo Track}, pages 59--63, 2012.

\bibitem{goldstein2016comparative}
Markus Goldstein and Seiichi Uchida.
\newblock A comparative evaluation of unsupervised anomaly detection algorithms
  for multivariate data.
\newblock {\em PloS one}, 11(4), 2016.

\bibitem{goodfellow2014generative}
Ian Goodfellow, Jean Pouget-Abadie, Mehdi Mirza, Bing Xu, David Warde-Farley,
  Sherjil Ozair, Aaron Courville, and Yoshua Bengio.
\newblock Generative adversarial nets.
\newblock In {\em Advances in neural information processing systems}, pages
  2672--2680, 2014.

\bibitem{gopalanPIDForestAnomalyDetection2019}
Parikshit Gopalan, Vatsal Sharan, and Udi Wieder.
\newblock {PIDForest: Anomaly Detection via Partial Identification}.
\newblock {\em arXiv:1912.03582 [cs]}, 2019.

\bibitem{grathwohlFFJORDFreeformContinuous2018}
Will Grathwohl, Ricky T.~Q. Chen, Jesse Bettencourt, Ilya Sutskever, and David
  Duvenaud.
\newblock {{FFJORD}}: {{Free}}-form {{Continuous Dynamics}} for {{Scalable
  Reversible Generative Models}}.
\newblock {\em arXiv:1810.01367 [cs, stat]}, October 2018.

\bibitem{gulrajani2017improved}
Ishaan Gulrajani, Faruk Ahmed, Martin Arjovsky, Vincent Dumoulin, and Aaron~C.
  Courville.
\newblock Improved training of wasserstein gans.
\newblock In {\em Advances in neural information processing systems}, pages
  5767--5777, 2017.

\bibitem{harmeling2006outliers}
Stefan Harmeling, Guido Dornhege, David Tax, Frank Meinecke, and Klaus-Robert
  M{\"u}ller.
\newblock From outliers to prototypes: ordering data.
\newblock {\em Neurocomputing}, 69(13-15):1608--1618, 2006.

\bibitem{he2016deep}
Kaiming He, Xiangyu Zhang, Shaoqing Ren, and Jian Sun.
\newblock Deep residual learning for image recognition.
\newblock In {\em Proceedings of the IEEE conference on computer vision and
  pattern recognition}, pages 770--778, 2016.

\bibitem{skopt}
Tim Head, Manoj Kumar, Holger Nahrstaedt, Gilles Louppe, and Iaroslav
  Shcherbatyi.
\newblock scikit-optimize, 2018.
\newblock 10.5281/zenodo.1157319.

\bibitem{hong2019generative}
Yongjun Hong, Uiwon Hwang, Jaeyoon Yoo, and Sungroh Yoon.
\newblock How generative adversarial networks and their variants work: An
  overview.
\newblock {\em ACM Computing Surveys (CSUR)}, 52(1):1--43, 2019.

\bibitem{innes:2018}
Mike Innes.
\newblock Flux: Elegant machine learning with julia.
\newblock {\em Journal of Open Source Software}, 2018.

\bibitem{iwataSupervisedAnomalyDetection2019}
Tomoharu Iwata and Yuki Yamanaka.
\newblock Supervised {{Anomaly Detection}} based on {{Deep Autoregressive
  Density Estimators}}.
\newblock {\em arXiv:1904.06034 [cs, stat]}, April 2019.

\bibitem{kingma2014adam}
Diederik~P. Kingma and Jimmy Ba.
\newblock Adam: A method for stochastic optimization.
\newblock {\em arXiv:1412.6980 [cs]}, 2017.

\bibitem{kingmaGlowGenerativeFlow2018}
Diederik~P. Kingma and Prafulla Dhariwal.
\newblock Glow: {{Generative Flow}} with {{Invertible}} 1x1 {{Convolutions}}.
\newblock {\em arXiv:1807.03039 [cs, stat]}, July 2018.

\bibitem{kingma2013auto}
Diederik~P Kingma and Max Welling.
\newblock Auto-encoding variational bayes.
\newblock {\em arXiv:1312.6114 [stat]}, 2014.

\bibitem{kiran2018overview}
B~Ravi Kiran, Dilip~Mathew Thomas, and Ranjith Parakkal.
\newblock An overview of deep learning based methods for unsupervised and
  semi-supervised anomaly detection in videos.
\newblock {\em Journal of Imaging}, 4(2):36, 2018.

\bibitem{kirichenkoWhyNormalizingFlows2020}
Polina Kirichenko, Pavel Izmailov, and Andrew~Gordon Wilson.
\newblock Why normalizing flows fail to detect out-of-distribution data.
\newblock {\em arXiv:2006.08545 [stat]}, 2020.

\bibitem{kliger2018novelty}
Mark Kliger and Shachar Fleishman.
\newblock Novelty detection with gan.
\newblock {\em arXiv:1802.10560 [cs]}, 2018.

\bibitem{kobyzevNormalizingFlowsIntroduction2020}
Ivan Kobyzev, Simon J.~D. Prince, and Marcus~A. Brubaker.
\newblock Normalizing {{Flows}}: {{An Introduction}} and {{Review}} of
  {{Current Methods}}.
\newblock {\em IEEE Transactions on Pattern Analysis and Machine Intelligence},
  pages 1--1, 2020.

\bibitem{kriegel2008angle}
Hans-Peter Kriegel, Matthias Schubert, and Arthur Zimek.
\newblock Angle-based outlier detection in high-dimensional data.
\newblock In {\em Proceedings of the 14th ACM SIGKDD international conference
  on Knowledge discovery and data mining}, pages 444--452, 2008.

\bibitem{krizhevsky2009learning}
Alex Krizhevsky and Geoffrey Hinton.
\newblock Learning multiple layers of features from tiny images.
\newblock {\em University of Toronto}, 05 2012.

\bibitem{kwon2019survey}
Donghwoon Kwon, Hyunjoo Kim, Jinoh Kim, Sang~C Suh, Ikkyun Kim, and Kuinam~J
  Kim.
\newblock A survey of deep learning-based network anomaly detection.
\newblock {\em Cluster Computing}, pages 1--13, 2019.

\bibitem{lecun-mnisthandwrittendigit-2010}
Yann LeCun and Corinna Cortes.
\newblock {MNIST} handwritten digit database.
\newblock 2010.

\bibitem{liu2008isolation}
Fei~Tony Liu, Kai~Ming Ting, and Zhi-Hua Zhou.
\newblock Isolation forest.
\newblock In {\em 2008 Eighth IEEE International Conference on Data Mining},
  pages 413--422. IEEE, 2008.

\bibitem{liu2019generative}
Yezheng Liu, Zhe Li, Chong Zhou, Yuanchun Jiang, Jianshan Sun, Meng Wang, and
  Xiangnan He.
\newblock Generative adversarial active learning for unsupervised outlier
  detection.
\newblock {\em IEEE Transactions on Knowledge and Data Engineering}, 2019.

\bibitem{makhzani2015adversarial}
Alireza Makhzani, Jonathon Shlens, Navdeep Jaitly, Ian Goodfellow, and Brendan
  Frey.
\newblock Adversarial autoencoders.
\newblock {\em arXiv:1511.05644 [cs]}, 2016.

\bibitem{moustafa2019holistic}
Nour Moustafa, Jiankun Hu, and Jill Slay.
\newblock A holistic review of network anomaly detection systems: A
  comprehensive survey.
\newblock {\em Journal of Network and Computer Applications}, 128:33--55, 2019.

\bibitem{muMNISTCRobustnessBenchmark2019}
Norman Mu and Justin Gilmer.
\newblock {{MNIST}}-{{C}}: {{A Robustness Benchmark}} for {{Computer Vision}}.
\newblock {\em arXiv:1906.02337 [cs]}, June 2019.

\bibitem{nalisnickDeepGenerativeModels2019}
Eric Nalisnick, Akihiro Matsukawa, Yee~Whye Teh, Dilan Gorur, and Balaji
  Lakshminarayanan.
\newblock Do {{Deep Generative Models Know What They Don}}'t {{Know}}?
\newblock {\em arXiv:1810.09136 [cs, stat]}, February 2019.

\bibitem{netzer2011reading}
Yuval Netzer, Tao Wang, Adam Coates, Alessandro Bissacco, Bo~Wu, and Andrew~Y.
  Ng.
\newblock Reading digits in natural images with unsupervised feature learning.
\newblock In {\em NIPS Workshop on Deep Learning and Unsupervised Feature
  Learning 2011}, 2011.

\bibitem{pangLearningRepresentationsUltrahighdimensional2018}
Guansong Pang, Longbing Cao, Ling Chen, and Huan Liu.
\newblock Learning {{Representations}} of {{Ultrahigh}}-dimensional {{Data}}
  for {{Random Distance}}-based {{Outlier Detection}}.
\newblock {\em arXiv:1806.04808 [cs, stat]}, June 2018.

\bibitem{pang2020deep}
Guansong Pang, Chunhua Shen, Longbing Cao, and Anton van~den Hengel.
\newblock Deep learning for anomaly detection: A review.
\newblock {\em arXiv:2007.02500 [cs]}, 2020.

\bibitem{papamakariosNormalizingFlowsProbabilistic2019}
George Papamakarios, Eric Nalisnick, Danilo~Jimenez Rezende, Shakir Mohamed,
  and Balaji Lakshminarayanan.
\newblock Normalizing {{Flows}} for {{Probabilistic Modeling}} and
  {{Inference}}.
\newblock {\em arXiv:1912.02762 [cs, stat]}, December 2019.

\bibitem{papamakariosMaskedAutoregressiveFlow2018}
George Papamakarios, Theo Pavlakou, and Iain Murray.
\newblock Masked {{Autoregressive Flow}} for {{Density Estimation}}.
\newblock {\em arXiv:1705.07057 [cs, stat]}, June 2018.

\bibitem{scikit-learn}
F.~Pedregosa, G.~Varoquaux, A.~Gramfort, V.~Michel, B.~Thirion, O.~Grisel,
  M.~Blondel, P.~Prettenhofer, R.~Weiss, V.~Dubourg, J.~Vanderplas, A.~Passos,
  D.~Cournapeau, M.~Brucher, M.~Perrot, and E.~Duchesnay.
\newblock Scikit-learn: Machine learning in {P}ython.
\newblock {\em Journal of Machine Learning Research}, 12:2825--2830, 2011.

\bibitem{pereira2018unsupervised}
Joao Pereira and Margarida Silveira.
\newblock Unsupervised anomaly detection in energy time series data using
  variational recurrent autoencoders with attention.
\newblock In {\em 2018 17th IEEE International Conference on Machine Learning
  and Applications (ICMLA)}, pages 1275--1282. IEEE, 2018.

\bibitem{perera2019ocgan}
Pramuditha Perera, Ramesh Nallapati, and Bing Xiang.
\newblock Ocgan: One-class novelty detection using {GAN}s with constrained
  latent representations.
\newblock In {\em Proceedings of the IEEE Conference on Computer Vision and
  Pattern Recognition}, pages 2898--2906, 2019.

\bibitem{pevny2016loda}
Tom{\'a}{\v{s}} Pevn{\'y}.
\newblock Loda: Lightweight on-line detector of anomalies.
\newblock {\em Machine Learning}, 102(2):275--304, 2016.

\bibitem{pevny2020sum}
Tomas Pevny, Vasek Smidl, Martin Trapp, Ondrej Polacek, and Tomas Oberhuber.
\newblock Sum-product-transform networks: Exploiting symmetries using
  invertible transformations.
\newblock {\em arXiv:2005.01297 [stat]}, 2020.

\bibitem{pidhorskyi2018generative}
Stanislav Pidhorskyi, Ranya Almohsen, and Gianfranco Doretto.
\newblock Generative probabilistic novelty detection with adversarial
  autoencoders.
\newblock In {\em Advances in neural information processing systems}, pages
  6822--6833, 2018.

\bibitem{pimentel2014review}
Marco~AF Pimentel, David~A Clifton, Lei Clifton, and Lionel Tarassenko.
\newblock A review of novelty detection.
\newblock {\em Signal Processing}, 99:215--249, 2014.

\bibitem{principi2017acoustic}
Emanuele Principi, Fabio Vesperini, Stefano Squartini, and Francesco Piazza.
\newblock Acoustic novelty detection with adversarial autoencoders.
\newblock In {\em 2017 International Joint Conference on Neural Networks
  (IJCNN)}, pages 3324--3330. IEEE, 2017.

\bibitem{ramaswamy2000efficient}
Sridhar Ramaswamy, Rajeev Rastogi, and Kyuseok Shim.
\newblock Efficient algorithms for mining outliers from large data sets.
\newblock In {\em Proceedings of the 2000 ACM SIGMOD international conference
  on Management of data}, pages 427--438, 2000.

\bibitem{renLikelihoodRatiosOutofDistribution2019}
Jie Ren, Peter~J. Liu, Emily Fertig, Jasper Snoek, Ryan Poplin, Mark~A.
  DePristo, Joshua~V. Dillon, and Balaji Lakshminarayanan.
\newblock Likelihood {{Ratios}} for {{Out}}-of-{{Distribution Detection}}.
\newblock {\em arXiv:1906.02845 [cs, stat]}, December 2019.

\bibitem{ruff2020unifying}
Lukas Ruff, Jacob~R Kauffmann, Robert~A Vandermeulen, Gr{\'e}goire Montavon,
  Wojciech Samek, Marius Kloft, Thomas~G Dietterich, and Klaus-Robert
  M{\"u}ller.
\newblock A unifying review of deep and shallow anomaly detection.
\newblock {\em arXiv:2009.11732 [cs]}, 2020.

\bibitem{ruff2018deep}
Lukas Ruff, Robert Vandermeulen, Nico Goernitz, Lucas Deecke, Shoaib~Ahmed
  Siddiqui, Alexander Binder, Emmanuel M{\"u}ller, and Marius Kloft.
\newblock Deep one-class classification.
\newblock In {\em International conference on machine learning}, pages
  4393--4402, 2018.

\bibitem{salimans2016improved}
Tim Salimans, Ian Goodfellow, Wojciech Zaremba, Vicki Cheung, Alec Radford, and
  Xi~Chen.
\newblock Improved techniques for training {GAN}s.
\newblock In {\em Advances in neural information processing systems}, pages
  2234--2242, 2016.

\bibitem{schleglFAnoGANFastUnsupervised2019}
Thomas Schlegl, Philipp Seeb{\"o}ck, Sebastian~M. Waldstein, Georg Langs, and
  Ursula {Schmidt-Erfurth}.
\newblock F-{{AnoGAN}}: {{Fast}} unsupervised anomaly detection with generative
  adversarial networks.
\newblock {\em Medical Image Analysis}, 54:30--44, May 2019.

\bibitem{schlegl2017unsupervised}
Thomas Schlegl, Philipp Seeb{\"o}ck, Sebastian~M Waldstein, Ursula
  Schmidt-Erfurth, and Georg Langs.
\newblock Unsupervised anomaly detection with generative adversarial networks
  to guide marker discovery.
\newblock In {\em International Conference on Information Processing in Medical
  Imaging}, pages 146--157. Springer, 2017.

\bibitem{schmidtNormalizingFlowsNovelty2019}
Maximilian Schmidt and Marko Simic.
\newblock Normalizing flows for novelty detection in industrial time series
  data.
\newblock {\em arXiv:1906.06904 [cs, stat]}, June 2019.

\bibitem{scholkopf2001estimating}
Bernhard Sch{\"o}lkopf, John~C Platt, John Shawe-Taylor, Alex~J Smola, and
  Robert~C Williamson.
\newblock Estimating the support of a high-dimensional distribution.
\newblock {\em Neural computation}, 13(7):1443--1471, 2001.

\bibitem{vskvara2018generative}
V{\'\i}t {\v{S}}kv{\'a}ra, Tom{\'a}{\v{s}} Pevn{\'y}, and V{\'a}clav
  {\v{S}}m{\'\i}dl.
\newblock Are generative deep models for novelty detection truly better?
\newblock {\em arXiv:1807.05027 [cs]}, 2018.

\bibitem{vskvara2020detection}
V{\'\i}t {\v{S}}kv{\'a}ra, V{\'a}clav {\v{S}}m{\'\i}dl, Tom{\'a}{\v{s}}
  Pevn{\`y}, Jakub Seidl, Ale{\v{s}} Havr{\'a}nek, and David Tskhakaya.
\newblock Detection of alfv{\'e}n eigenmodes on {COMPASS} with generative
  neural networks.
\newblock {\em Fusion Science and Technology}, 2020.

\bibitem{vsmidl2019anomaly}
V{\'a}clav {\v{S}}m{\'\i}dl, Jan B{\'\i}m, and Tom{\'a}{\v{s}} Pevn{\'y}.
\newblock Anomaly scores for generative models.
\newblock {\em arXiv:1905.11890 [stat]}, 2019.

\bibitem{steinwart2005a}
Ingo Steinwart, Don Hush, and Clint Scovel.
\newblock A classification framework for anomaly detection.
\newblock {\em Journal of Machine Learning Research}, 6(8):211--232, 2005.

\bibitem{tolstikhin2017wasserstein}
Ilya Tolstikhin, Olivier Bousquet, Sylvain Gelly, and Bernhard Schoelkopf.
\newblock Wasserstein auto-encoders.
\newblock {\em arXiv:1711.01558 [stat]}, 2019.

\bibitem{tomczak2018vae}
Jakub Tomczak and Max Welling.
\newblock {VAE} with a {VampPrior}.
\newblock In {\em International Conference on Artificial Intelligence and
  Statistics}, pages 1214--1223, 2018.

\bibitem{wang2019progress}
Hongzhi Wang, Mohamed~Jaward Bah, and Mohamed Hammad.
\newblock Progress in outlier detection techniques: A survey.
\newblock {\em IEEE Access}, 7:107964--108000, 2019.

\bibitem{wang2020advae}
Xuhong Wang, Ying Du, Shijie Lin, Ping Cui, Yuntian Shen, and Yupu Yang.
\newblock advae: A self-adversarial variational autoencoder with gaussian
  anomaly prior knowledge for anomaly detection.
\newblock {\em Knowledge-Based Systems}, 190:105187, 2020.

\bibitem{wilson2020case}
Andrew~Gordon Wilson.
\newblock The case for bayesian deep learning.
\newblock {\em arXiv:2001.10995 [cs]}, 2020.

\bibitem{xiao2017fashion}
Han Xiao, Kashif Rasul, and Roland Vollgraf.
\newblock Fashion-mnist: a novel image dataset for benchmarking machine
  learning algorithms.
\newblock {\em arXiv:1708.07747 [cs]}, 2017.

\bibitem{xu2018unsupervised}
Haowen Xu, Wenxiao Chen, Nengwen Zhao, Zeyan Li, Jiahao Bu, Zhihan Li, Ying
  Liu, Youjian Zhao, Dan Pei, Yang Feng, et~al.
\newblock Unsupervised anomaly detection via variational auto-encoder for
  seasonal kpis in web applications.
\newblock In {\em Proceedings of the 2018 World Wide Web Conference}, pages
  187--196, 2018.

\bibitem{yamaguchi2019adaflow}
Masataka Yamaguchi, Yuma Koizumi, and Noboru Harada.
\newblock Adaflow: Domain-adaptive density estimator with application to
  anomaly detection and unpaired cross-domain translation.
\newblock In {\em ICASSP 2019-2019 IEEE International Conference on Acoustics,
  Speech and Signal Processing (ICASSP)}, pages 3647--3651. IEEE, 2019.

\bibitem{yaoUnsupervisedAnomalyDetection2019}
Rong Yao, Chongdang Liu, Linxuan Zhang, and Peng Peng.
\newblock Unsupervised {{Anomaly Detection Using Variational Auto}}-{{Encoder}}
  based {{Feature Extraction}}.
\newblock In {\em 2019 {{IEEE International Conference}} on {{Prognostics}} and
  {{Health Management}} ({{ICPHM}})}, pages 1--7, {San Francisco, CA, USA},
  June 2019. {IEEE}.

\bibitem{zenatiEfficientGANBasedAnomaly2018}
Houssam Zenati, Chuan~Sheng Foo, Bruno Lecouat, Gaurav Manek, and
  Vijay~Ramaseshan Chandrasekhar.
\newblock Efficient {GAN}-based anomaly detection.
\newblock {\em arXiv:1802.06222 [cs]}, 2019.

\bibitem{zhao2017infovae}
Shengjia Zhao, Jiaming Song, and Stefano Ermon.
\newblock Infovae: Information maximizing variational autoencoders.
\newblock {\em arXiv:1706.02262 [cs]}, 2018.

\bibitem{zhao2019pyod}
Yue Zhao, Zain Nasrullah, and Zheng Li.
\newblock Pyod: A python toolbox for scalable outlier detection.
\newblock {\em Journal of Machine Learning Research}, 20(96):1--7, 2019.

\bibitem{zong2018deep}
Bo~Zong, Qi~Song, Martin~Renqiang Min, Wei Cheng, Cristian Lumezanu, Daeki Cho,
  and Haifeng Chen.
\newblock Deep autoencoding gaussian mixture model for unsupervised anomaly
  detection.
\newblock In {\em International Conference on Learning Representations}, 2018.

\end{thebibliography}

\begin{table*}
\centering \Huge{Supplementary material for:\\ Comparison of Anomaly Detectors: 
Context Matters
}
\end{table*}
\vfill*{}
\pagebreak
\appendices 
\counterwithin{figure}{section}
\counterwithin{table}{section}
\counterwithin{footnote}{section}

\section{Experimental setup}

\subsection{Data splits}
Our decision for not doing repeated experiments on different random crossvalidation data splits in MNIST, SVHN2, FashionMNIST nad CIFAR10 datasets was motivated by a need to reduce the computational requirements of experiments and justified by preliminary results that are summarized in Tab.~\ref{tab:images_seed_consistency}. The consistent performance across different resamplings of the datasets is due to their large size and homogeneity. 

\begin{table*}
    \centering
    \footnotesize
    \begin{subtable}{.45\linewidth}
    \centering
    \tabcolsep=0.10cm
    \begin{tabular}{rrrrrr}
\toprule
 & \textbf{seed = 1} & \textbf{seed = 2} & \textbf{seed = 3} & \textbf{seed = 4} & \textbf{seed = 5} \\\midrule
cifar10 & \color{blue}{\textbf{0.63}} & \color{blue}{\textbf{0.63}} & \color{blue}{\textbf{0.63}} & 0.62 & \color{blue}{\textbf{0.63}} \\
fmnist & 0.77 & 0.77 & 0.77 & \color{blue}{\textbf{0.79}} & 0.77 \\
mnist & \color{blue}{\textbf{0.88}} & \color{blue}{\textbf{0.88}} & 0.87 & 0.87 & \color{blue}{\textbf{0.88}} \\
svhn2 & \color{blue}{\textbf{0.54}} & \color{blue}{\textbf{0.54}} & 0.53 & \color{blue}{\textbf{0.54}} & \color{blue}{\textbf{0.54}} \\\bottomrule
\end{tabular}

    \vspace*{0.05cm}
    \caption{VAE}
    \end{subtable}
    \begin{subtable}{.45\linewidth}
    \centering
    \tabcolsep=0.10cm
    \begin{tabular}{rrrrrr}
\toprule
 & \textbf{seed = 1} & \textbf{seed = 2} & \textbf{seed = 3} & \textbf{seed = 4} & \textbf{seed = 5} \\\midrule
cifar10 & \color{blue}{\textbf{0.65}} & \color{blue}{\textbf{0.65}} & 0.64 & \color{blue}{\textbf{0.65}} & \color{blue}{\textbf{0.65}} \\
fmnist & \color{blue}{\textbf{0.82}} & \color{blue}{\textbf{0.82}} & \color{blue}{\textbf{0.82}} & \color{blue}{\textbf{0.82}} & 0.81 \\
mnist & 0.88 & \color{blue}{\textbf{0.89}} & 0.88 & 0.88 & 0.88 \\
svhn2 & \color{blue}{\textbf{0.52}} & \color{blue}{\textbf{0.52}} & \color{blue}{\textbf{0.52}} & \color{blue}{\textbf{0.52}} & \color{blue}{\textbf{0.52}} \\\bottomrule
\end{tabular}

    \vspace*{0.05cm}
    \caption{GANomaly}
    \end{subtable}
    \caption{Summary tables for the VAE and GANomaly models on different seeds of the image datasets. Reported values are the AUC metrics on the test data, averaged over 10 anomaly classes.
    }
    \label{tab:images_seed_consistency}
\end{table*}

On the other hand in the Tab.~\ref{tab:dataset_per_seed_ranks} we can see that in the case of tabular data and the other image datasets the performance on each split varies to a greater extent. The biggest changes can be observed on MVTec-AD datasets, due to it's small size, whereas MNIST-C shows only small changes. On tabular data the situation is even more chaotic due to the dataset diversity, thus using only one seed for the random split is in this case unreliable.

\begin{table*}
    \centering
    \begin{subtable}{\linewidth}
        \footnotesize
        \centering
        \tabcolsep=0.05cm
        \begin{tabular}{lrrrrrrrrrrrrrrrrrrrrrrrrr}
\toprule
& & \textbf{aae} & \textbf{avae} & \textbf{gano} & \textbf{vae} & \textbf{wae} & \textbf{abod} & \textbf{hbos} & \textbf{if} & \textbf{knn} & \textbf{loda} & \textbf{lof} & \textbf{osvm} & \textbf{pidf} & \textbf{maf} & \textbf{rnvp} & \textbf{sptn} & \textbf{fmgn} & \textbf{gan} & \textbf{mgal} & \textbf{dagm} & \textbf{dsvd} & \textbf{rpn} & \textbf{vaek} & \textbf{vaeo} \\\midrule
 & & 6.9 & 13.4 & 10.3 & 6.5 & 7.0 & 11.4 & 14.9 & 14.5 & 8.4 & 16.1 & 11.4 & \color{red}{\textbf{2.9}} & 14.3 & 8.9 & 9.0 & 10.7 & 11.4 & 12.0 & 22.3 & 19.8 & 16.2 & 12.1 & 11.0 & 6.8 \\ \cmidrule{3-26}
\multirow{5}{*}{\rot{SEED}} & 1 & 9.1 & 13.6 & 10.8 & 6.6 & 8.0 & 12.7 & 15.0 & 12.6 & 9.4 & 14.4 & 11.5 & \color{red}{\textbf{3.8}} & 13.6 & 8.6 & 9.5 & 11.7 & 10.0 & 11.2 & 18.2 & 20.4 & 17.3 & 11.6 & 11.8 & 6.4 \\
& 2 & 7.4 & 14.2 & 10.0 & 6.0 & 7.5 & 12.7 & 16.4 & 14.2 & 10.4 & 15.9 & 11.8 & \color{red}{\textbf{5.0}} & 14.5 & 8.3 & 7.7 & 10.8 & 11.2 & 10.8 & 18.4 & 20.2 & 17.0 & 11.3 & 10.9 & 6.0 \\
& 3 & 6.9 & 14.5 & 10.5 & 6.3 & 7.0 & 12.8 & 15.8 & 14.4 & 9.4 & 14.6 & 12.3 & \color{red}{\textbf{3.9}} & 14.1 & 8.5 & 8.6 & 11.9 & 11.0 & 10.8 & 16.6 & 19.7 & 17.1 & 12.1 & 11.4 & 5.6 \\
& 4 & 8.5 & 12.0 & 8.4 & 7.3 & 6.6 & 13.2 & 16.2 & 13.9 & 9.9 & 15.9 & 12.4 & \color{red}{\textbf{4.7}} & 13.7 & 8.0 & 7.8 & 12.6 & 11.2 & 10.4 & 19.0 & 19.4 & 16.5 & 10.4 & 11.0 & 6.6 \\
& 5 & 6.8 & 14.0 & 9.4 & 6.8 & 7.8 & 12.9 & 15.4 & 13.8 & 8.8 & 15.2 & 12.4 & \color{red}{\textbf{3.9}} & 14.3 & 9.1 & 9.4 & 12.6 & 11.6 & 10.8 & 17.6 & 20.1 & 16.8 & 11.5 & 10.6 & 6.0 \\\bottomrule
\end{tabular}

        \vspace*{0.15cm}
        \caption{tabular}
    \end{subtable}
    \begin{subtable}{.48\linewidth}
        \footnotesize
        \tabcolsep=0.05cm
        \vspace*{0.15cm}
        \begin{tabular}{lrrrrrrrrrrrrr}
\toprule
& & \textbf{aae} & \textbf{gano} & \textbf{skip} & \textbf{vae} & \textbf{wae} & \textbf{knn} & \textbf{osvm} & \textbf{fano} & \textbf{fmgn} & \textbf{dsvd} & \textbf{vaek} & \textbf{vaeo} \\\midrule
& & 4.1 & 9.9 & 10.6 & 2.9 & \color{red}{\textbf{2.4}} & 5.4 & 6.0 & 3.2 & 3.1 & 6.6 & 4.5 & \color{red}{\textbf{2.4}} \\ \cmidrule{3-14}
\multirow{5}{*}{\rot{SEED}} & 1 & 3.9 & 9.9 & 10.6 & 2.9 & \color{red}{\textbf{2.1}} & 5.7 & 5.9 & 3.3 & 2.8 & 6.6 & 4.7 & 2.3 \\ 
 & 2 & 3.9 & 9.9 & 10.6 & 3.0 & \color{red}{\textbf{2.2}} & 5.4 & 6.0 & 3.0 & 2.7 & 6.6 & 4.6 & 2.8 \\
 & 3 & 3.7 & 9.9 & 10.7 & 3.1 & \color{red}{\textbf{2.0}} & 5.7 & 6.0 & 3.1 & 2.4 & 6.8 & 4.6 & 2.6 \\
 & 4 & 3.7 & 9.9 & 10.6 & 2.9 & 2.3 & 5.4 & 6.1 & 3.0 & \color{red}{\textbf{2.1}} & 6.2 & 4.5 & 2.3 \\
 & 5 & 3.9 & 9.9 & 10.6 & 2.9 & \color{red}{\textbf{2.1}} & 5.5 & 6.0 & 3.0 & 3.0 & 6.0 & 4.6 & 2.6 \\\bottomrule
\end{tabular}

        \vspace*{0.15cm}
        \caption{MNIST-C}
    \end{subtable}
    \begin{subtable}{.48\linewidth}
        \footnotesize
        \tabcolsep=0.05cm
        \vspace*{0.15cm}
        \begin{tabular}{lrrrrrrrrrrrrr}
\toprule
& & \textbf{aae} & \textbf{gano} & \textbf{skip} & \textbf{vae} & \textbf{wae} & \textbf{knn} & \textbf{osvm} & \textbf{fano} & \textbf{fmgn} & \textbf{dsvd} & \textbf{vaek} & \textbf{vaeo} \\\midrule
& & 4.7 & 7.0 & 9.0 & \color{red}{\textbf{3.0}} & 3.7 & 6.0 & 6.7 & 4.0 & 4.3 & 9.7 & 9.0 & 8.0 \\ \cmidrule{3-14}
\multirow{5}{*}{\rot{SEED}} & 1 & 4.0 & 9.0 & 8.7 & \color{red}{\textbf{2.3}} & 3.7 & 4.7 & 7.3 & 5.0 & 6.3 & 9.0 & 8.7 & 7.0 \\
& 2 & 2.7 & 8.3 & 8.3 & 2.7 & 6.3 & 7.0 & 7.0 & \color{red}{\textbf{1.3}} & 3.0 & 8.0 & 9.7 & 10.3 \\
& 3 & 4.0 & 9.3 & 6.3 & \color{red}{\textbf{2.0}} & 4.0 & 6.0 & 7.7 & 3.3 & 5.3 & 6.7 & 9.7 & 9.7 \\
& 4 & 3.7 & 9.7 & 8.7 & 4.3 & 3.0 & 6.3 & 5.3 & 6.7 & \color{red}{\textbf{2.7}} & 7.0 & 9.0 & 9.3 \\
& 5 & 3.7 & 9.0 & 8.3 & 2.3 & \color{red}{\textbf{1.3}} & 5.7 & 7.7 & 4.3 & 4.7 & 7.7 & 9.3 & 11.0 \\\bottomrule
\end{tabular}

        \vspace*{0.15cm}
        \caption{MVTec-AD}
    \end{subtable}
    \caption{Comparison of average ranks in AUC on different train/val/test splits of datasets that were subject to repeated experiments. The first row corresponds to the average rank in AUC, which has been averaged over all 5  repetitions.}
    \label{tab:dataset_per_seed_ranks}
\end{table*}

\subsection{Hyperparameter settings}
An overview of sampled hyperparameters for all models used in our experiments is in Tab.~\ref{tab:classical_hyperparameters}--\ref{tab:flow_hyperparameters}. The default hyperparameters of classical methods used in the context of validation set without anomalies are present in Tab.~\ref{tab:default_hyperparameters}

\begin{table}[ht]
    \centering
    \tabcolsep=0.05cm
    \begin{tabular}{lcc}
\toprule
\textbf{model} & \textbf{parameter} &  \textbf{value set} \\
\midrule
\multirow{2}{*}{ABOD}     
            & $n$ & $\lbrace1, 2, \ldots, 100 \rbrace$ \\
            & method & fast  \\
\midrule
\multirow{3}{*}{HBOS}     
            & no. bins & $\lbrace 2, 4, \ldots, 100 \rbrace$ \\
            & $\alpha$ & $\lbrace 0.05, 0.1, \ldots, 1 \rbrace$ \\
            & tolerance & $\lbrace 0, 0.05, \ldots, 1 \rbrace$ \\
\midrule
\multirow{3}{*}{IF}
            & no. estimators & $\lbrace 50, 100, \ldots, 500 \rbrace$ \\
            & max samples & $\lbrace 0.5, 0.6, \ldots, 1 \rbrace$ \\
            & max features & $\lbrace 0.5, 0.6, \ldots, 1 \rbrace$ \\
\midrule
\multirow{1}{*}{kNN}
            & $n$ & $\lbrace1, 3, \ldots, 101 \rbrace$ \\
\midrule
\multirow{2}{*}{LODA}
            & no. bins & $\lbrace 2, 4, \ldots, 100 \rbrace$ \\
            & no. cuts & $\lbrace 40, 60, \ldots, 500 \rbrace$ \\
\midrule
\multirow{1}{*}{LOF}     
            & $n$ & $\lbrace1, 2, \ldots, 100 \rbrace$ \\
\midrule
\multirow{3}{*}{OC-SVM}
            & $\gamma$ & $ 10^x, x \in \lbrace -4, -3.9, \ldots, 2  \rbrace $ \\
            & $\nu$ & $ \lbrace 0.01, 0.5, 0.99  \rbrace $ \\
            & kernel & $ \lbrace \text{rbf}, \text{sigmoid}, \text{polynomial}  \rbrace $ \\
\midrule
\multirow{5}{*}{PIDForest}
            & max depth & $  \lbrace 6, 8, \ldots, 10  \rbrace $ \\
            & no. trees & $  \lbrace 50, 75, \ldots, 200  \rbrace $ \\
            & max samples & $  \lbrace 50, 100, 250, 500, 1000, 5000  \rbrace $ \\
            &max buckets & $  \lbrace 3, 4, 5, 6  \rbrace $ \\
            & $\epsilon$ & $ \lbrace 0.05, 0.1, 0.2  \rbrace $ \\
            \bottomrule
\end{tabular}
    \vspace*{0.15cm}
    \caption{Hyperparameters of the classical models.}
    \label{tab:classical_hyperparameters}
\end{table}

\begin{table}[ht]
    \centering
    \tabcolsep=0.05cm
    \begin{tabular}{lcc}
\toprule
\textbf{model} & \textbf{parameter} &  \textbf{value set} \\
\midrule
\multirow{10}{*}{common} 
            & dim($ \vec{z} $) & $\lbrace 8, 16, \ldots, 256 \rbrace$ \\
            & h & $\lbrace 16, 32, \ldots, 512\rbrace$ \\
            & $\eta$ & $\lbrace 10^{-4}, 10^{-3}\rbrace$ \\
            & batch size & $\lbrace 32, 64, 128\rbrace$ \\
            & activation & $\lbrace \text{relu}, \text{swish}, \text{tanh} \rbrace$ \\
            & no. dense layers & $\lbrace 3, 4 \rbrace $ \\
            & no. conv layers & $\lbrace 2, 3, 4 \rbrace $ \\
            & channels & (16, 32, 64, 128) \\
            & kernelsizes & (3, 5, 7, 9) \\
            & scalings & (1, 2, 2, 2) \\
            \midrule
\multirow{4}{*}{WAE}
            & $\lambda$ & $\lbrace 0.1,1 \rbrace$ \\
            & prior & $\lbrace \mathcal{N}(\vec{0}, \vec{I}), \text{VampPrior} \rbrace $ \\
            & K & $\lbrace 2, 4, \ldots, 64 \rbrace$ \\
            & kernel & $\lbrace \text{rbf}, \text{imq}, \text{rq} \rbrace$ \\
            & $\sigma_k$ & $\lbrace 10^{-3}, 10^{-2}, 10^{-1}, 10^{0} \rbrace$ \\
            \midrule
\multirow{4}{*}{AAE}
            & $\lambda$ & $\lbrace 0.1,1 \rbrace$ \\
            & prior & $\lbrace \mathcal{N}(\vec{0}, \vec{I}), \text{VampPrior} \rbrace $ \\
            & K & $\lbrace 2, 4, \ldots, 64 \rbrace$ \\
            & no. discriminator layers & $\lbrace 1, 2, 3 \rbrace$ \\
            & $\alpha$ & $\lbrace 0, 0.1, \ldots, 1 \rbrace$ \\
            \midrule
\multirow{7}{*}{adVAE}
            & dim($ \vec{z} $) & $\lbrace 2, 4, \ldots, 256 \rbrace$ \\
            & batch size & $\lbrace 32, 64 \rbrace$ \\
            & $\gamma$ & $\lbrace 5.10^{-4}, 10^{-3}, 5.10^{-3} \rbrace $ \\
            & $\lambda$ & $\lbrace 5.10^{-3}, 10^{-2}, 5.10^{-2} \rbrace $ \\
            & $m_x$ & $\lbrace 1, 1.5 \rbrace $ \\
            & $m_z$ & $\lbrace 40, 50, 60 \rbrace $\\
            & decay & $\lbrace 0, 0.1, \ldots, 0.5 \rbrace $ \\
            \midrule
\multirow{5}{*}{(skip)GANomaly}
            & decay & $\lbrace 0, 0.1, \ldots, 0.5 \rbrace $ \\
            & $w_{adv}, w_{con}, w_{enc}$ & $\lbrace 1, 10, 20, \ldots, 100 \rbrace $ \\
            & $\lambda$ & $\lbrace 0.1, 0.2, \ldots, 0.9 \rbrace $ \\
            & $R(x), L(X)$ & $\lbrace \text{MAE}, \text{MSE} \rbrace $ \\
            & no. conv layers & $\lbrace 1, 2, 3, 4 \rbrace $  \\
            & no. channels & $\lbrace 8, 16, \ldots, 128 \rbrace $ \\
            \midrule
\multirow{3}{*}{(fm)GAN}
            & dim($ \vec{z} $) & $\lbrace 2, 4, \ldots, 256 \rbrace$ \\
            & no. dense layers & $\lbrace 2, 3, 4 \rbrace $ \\
            & $\alpha$ & $\lbrace 10^{-3}, 10^{-2}, \ldots,  10^{3} \rbrace $ \\
            \midrule
\multirow{6}{*}{DeepSVDD}
            & $\eta_{\text{AE}}$ & $ \lbrace 10^{-4}, 10^{-3}  \rbrace $ \\
            & batch norm. & $\lbrace \text{true}, \text{false} \rbrace $ \\
            & batch size & $\lbrace 64, 128 \rbrace $ \\
            & objective & $\lbrace \text{soft boundary}, \text{one class} \rbrace$ \\
            & $\nu$ & $ \lbrace 0.01, 0.1, 0.5, 0.99  \rbrace $ \\
            & decay & $ 10^{-6} $ \\
            \midrule
\multirow{4}{*}{fAnoGAN}
            & $\eta_{\text{E}}$ & $ \lbrace 10^{-4}, 10^{-3}  \rbrace $ \\
            & weight clip & $ \lbrace 0.001, 0.005, 0.01, 0.05, 0.1  \rbrace $ \\
            & $n_{critic}$ & 5\\
            & no. generator iterations & 10000\\
            \midrule
\multirow{6}{*}{DAGMM}
            & $\eta$ & $ \lbrace 10^{-5}, 10^{-4}, 10^{-3}  \rbrace $ \\
            & dim($ \vec{z} $) & $\lbrace 1, 2 \rbrace $ \\
            & batch size & $\lbrace 32, 64, \ldots 256 \rbrace $ \\
            & no. dense layers & $\lbrace 2, 3 \rbrace$ \\
            & no. components & $\lbrace 2, 3, \ldots, 8 \rbrace$ \\
            & $\lambda_1$ & $ \lbrace 0.1, 0.5, 1.0  \rbrace $ \\
            & $\lambda_2$ & $ \lbrace 0.005, 0.05, 0.1, 0.5, 1.0  \rbrace $ \\
            & activation & $\lbrace \text{tanh} \rbrace$ \\
            & dropout & $\lbrace 0.0, 0.1, \ldots, 0.5 \rbrace$ \\
            \midrule
\multirow{6}{*}{REPEN}
            & dim($ \vec{z} $) & $\lbrace 2, 4, \ldots, 64 \rbrace $ \\
            & h & $\lbrace 2, 4, \ldots, 256 \rbrace $ \\
            & confidence margin & $\lbrace 512, 1024, 2048 \rbrace $ \\
            & batch size & $\lbrace 32, 64, \ldots 256 \rbrace $ \\
            & no. dense layers & $\lbrace 1, 2 \rbrace$ \\
            & ensemble size & $ \lbrace 25, 50  \rbrace $ \\
            & subsample size & $ \lbrace 32, 64, \ldots, 258  \rbrace $ \\
            & activation & $\lbrace \text{tanh}, \text{relu} \rbrace$ \\
            \bottomrule
\end{tabular}

    \vspace*{0.15cm}
    \caption{Hyperparameters for the neural network based models.}
    \label{tab:nn_hyperparameters}
\end{table}

\begin{table}[ht]
    \centering
    \tabcolsep=0.05cm
    \begin{tabular}{lcc}
\toprule
\textbf{model} & \textbf{parameter} &  \textbf{value set} \\
\midrule
\multirow{5}{*}{MAF + RealNVP}
    & h & $\lbrace 16, 32, \ldots, 1024\rbrace$ \\
    & no. flows & $\lbrace 2, 4, 8 \rbrace$ \\
    & $\eta$ & $ 10^{-4}$ \\
    & batch size & $ \lbrace 32, 64, 128 \rbrace $ \\
    & no. layers & $\lbrace 2, 3 \rbrace $ \\
    & activation & $\lbrace \text{relu, tanh} \rbrace $ \\
    & batch norm. & $\lbrace \text{true, false} \rbrace $ \\
    & init. identity. & $\lbrace \text{true, false} \rbrace $ \\
    & $L_2$ reg. & $\lbrace 0.0, 10^{-5}, 10^{-6} \rbrace$ \\
    \midrule
MAF
    & ordering & $\lbrace \text{natural, random} \rbrace $ \\
    \midrule
RealNVP
    & tanh scaling & $\lbrace \text{true, false} \rbrace $ \\
    \midrule
\multirow{6}{*}{SPTN}
    & no. components & $\lbrace 2, 4, 8, 16 \rbrace$ \\
    & batch size & $ \lbrace 32, 64, 128 \rbrace $ \\
    & no. flows & $\lbrace 1, 2, 3 \rbrace $ \\
    & sharing & $\lbrace \text{dense, all, none} \rbrace $  \\
    & first dense & $\lbrace \text{true, false} \rbrace $ \\
    & activation & $\text{identity}$ \\
    \bottomrule
\end{tabular}

    \vspace*{0.15cm}
    \caption{Hyperparameters for the flow-based models.}
    \label{tab:flow_hyperparameters}
\end{table}

\begin{table}[ht]
    \centering
    \tabcolsep=0.05cm
    \begin{tabular}{lcc}
\toprule
\textbf{model} & \textbf{parameter} &  \textbf{value} \\
\midrule
\multirow{2}{*}{ABOD}     
            & $n$ & $0.1 N$ \\
            & method & fast  \\
\midrule
\multirow{3}{*}{HBOS}     
            & no. bins & $10$ \\
            & $\alpha$ & $0.1$ \\
            & tolerance & $0.1$ \\
\midrule
\multirow{3}{*}{IF}
            & no. estimators & $100$ \\
            & max samples & auto \\
            & max features & $1.0$ \\
\midrule
kNN         & $n$ & $\max(10, 0.03 N)$ \\
\midrule
LODA        & no. bins, no. cuts & automated procedure~\cite{pevny2016loda} \\
\midrule
LOF         & $n$ & $\max(10, 0.03 N)$ \\
\midrule
\multirow{3}{*}{OC-SVM}
            & $\gamma$ & 1/(median $L2$ distance of training samples) \\
            & $\nu$ & $ 0.5 $ \\
            & kernel & rbf \\
\midrule
\multirow{5}{*}{PIDForest}
            & max depth & $  10 $ \\
            & no. trees & $  50 $ \\
            & max samples &  $\min \left( 100, \max(25, N-\text{mod}(N, 50))\right)$ \\
            &max buckets & $  3 $ \\
            & $\epsilon$ & $ 0.1 $ \\
            \bottomrule
\end{tabular}
    \vspace*{0.15cm}
    \caption{Default hyperparameters of classical models for experiments without validation anomalies. $N$ is the number of samples in the training dataset.}
    \label{tab:default_hyperparameters}
\end{table}

\section{Dataset Context Details}
\label{app:Datasets}
The underlying data for critical diagrams in Section~\ref{sec:dataset_context} are displayed in Tables ~\ref{tab:tabular_anomalies}, \ref{tab:tabular_clean}, \ref{tab:images_stat_auc_auc_combined}, and~\ref{tab:images_semantic_auc_auc_combined} in the form of testing AUC for all individual datasets and all tested methods.

\begin{table*}
    \footnotesize
    \centering
    \tabcolsep=0.05cm
    \begin{tabular}{rrrrrrrrrrrrrrrrrrrrrrrrr}
\toprule
\textbf{dataset} & \textbf{aae} & \textbf{avae} & \textbf{gano} & \textbf{vae} & \textbf{wae} & \textbf{abod} & \textbf{hbos} & \textbf{if} & \textbf{knn} & \textbf{loda} & \textbf{lof} & \textbf{osvm} & \textbf{pidf} & \textbf{maf} & \textbf{rnvp} & \textbf{sptn} & \textbf{fmgn} & \textbf{gan} & \textbf{mgal} & \textbf{dagm} & \textbf{dsvd} & \textbf{rpn} & \textbf{vaek} & \textbf{vaeo} \\\midrule
aba & 0.92 & 0.87 & 0.89 & 0.92 & 0.91 & \color{blue}{\textbf{0.93}} & 0.75 & 0.87 & \color{blue}{\textbf{0.93}} & 0.84 & 0.90 & \color{blue}{\textbf{0.93}} & 0.89 & 0.91 & 0.90 & 0.91 & 0.78 & 0.80 & 0.62 & 0.67 & 0.82 & 0.84 & 0.91 & 0.90 \\
ann & 0.82 & 0.83 & 0.80 & 0.84 & 0.88 & 0.78 & 0.89 & 0.78 & 0.78 & 0.69 & 0.80 & \color{blue}{\textbf{0.99}} & 0.93 & 0.85 & 0.86 & 0.87 & 0.81 & 0.74 & 0.65 & 0.60 & 0.65 & 0.77 & 0.81 & 0.81 \\
arr & 0.71 & 0.76 & 0.77 & 0.74 & 0.76 & 0.74 & 0.77 & 0.78 & 0.74 & 0.77 & 0.73 & \color{blue}{\textbf{0.81}} & 0.75 & 0.76 & 0.77 & 0.74 & 0.74 & 0.73 & 0.55 & 0.51 & 0.72 & 0.78 & 0.71 & 0.79 \\
bcw & 0.99 & 0.95 & 0.94 & 0.99 & 0.99 & 0.94 & 0.97 & 0.97 & 0.93 & 0.94 & 0.94 & 0.99 & 0.91 & 0.99 & 0.98 & 0.95 & \color{blue}{\textbf{1.00}} & 0.99 & 0.64 & 0.71 & 0.83 & 0.96 & 0.93 & 0.99 \\
blt & \color{blue}{\textbf{0.99}} & 0.84 & 0.90 & 0.95 & 0.93 & 0.87 & 0.88 & 0.90 & 0.89 & 0.82 & 0.91 & 0.89 & 0.87 & 0.96 & 0.93 & 0.94 & 0.71 & 0.58 & 0.82 & 0.85 & 0.94 & 0.95 & 0.93 & 0.94 \\
bts & \color{blue}{\textbf{1.00}} & 0.85 & 0.99 & 0.98 & 0.96 & 0.99 & 0.98 & 0.98 & 0.98 & 0.95 & 0.96 & \color{blue}{\textbf{1.00}} & 0.77 & 0.98 & 0.99 & 0.99 & 0.96 & 0.95 & 0.68 & 0.78 & 0.96 & 0.96 & 0.97 & 0.97 \\
crd & 0.88 & 0.61 & 0.69 & 0.72 & 0.53 & 0.56 & 0.50 & 0.69 & 0.61 & 0.74 & 0.67 & \color{blue}{\textbf{0.90}} & 0.64 & 0.60 & 0.51 & 0.50 & 0.67 & 0.66 & 0.72 & 0.78 & 0.86 & 0.75 & 0.63 & 0.83 \\
eco & \color{blue}{\textbf{0.90}} & 0.86 & 0.84 & 0.85 & 0.87 & 0.87 & 0.81 & 0.83 & 0.88 & 0.77 & 0.80 & 0.89 & 0.84 & \color{blue}{\textbf{0.90}} & 0.85 & 0.88 & 0.85 & 0.87 & 0.58 & 0.70 & 0.76 & 0.88 & 0.86 & 0.85 \\
gls & \color{blue}{\textbf{0.87}} & 0.77 & 0.77 & 0.68 & 0.76 & 0.79 & 0.62 & 0.52 & 0.71 & 0.51 & 0.81 & 0.78 & 0.40 & 0.73 & 0.74 & 0.78 & 0.86 & 0.84 & 0.65 & 0.70 & 0.67 & 0.73 & 0.73 & 0.72 \\
hab & \color{blue}{\textbf{0.97}} & \color{blue}{\textbf{0.97}} & 0.87 & 0.95 & \color{blue}{\textbf{0.97}} & 0.95 & 0.92 & 0.93 & 0.95 & 0.95 & 0.96 & 0.95 & \color{blue}{\textbf{0.97}} & 0.96 & 0.95 & 0.96 & 0.76 & 0.81 & 0.68 & 0.84 & 0.96 & 0.95 & 0.95 & 0.93 \\
har & 0.97 & 0.69 & 0.99 & \color{blue}{\textbf{1.00}} & \color{blue}{\textbf{1.00}} & 0.76 & 0.84 & 0.71 & 0.76 & 0.81 & 0.97 & \color{blue}{\textbf{1.00}} & 0.47 & \color{blue}{\textbf{1.00}} & \color{blue}{\textbf{1.00}} & 0.96 & \color{blue}{\textbf{1.00}} & 0.99 & 0.58 & 0.60 & 0.84 & 0.99 & 0.48 & \color{blue}{\textbf{1.00}} \\
htr & 0.96 & 0.93 & 0.94 & 0.96 & 0.96 & 0.95 & 0.96 & 0.95 & 0.95 & 0.95 & 0.95 & \color{blue}{\textbf{0.97}} & 0.94 & 0.95 & 0.95 & 0.95 & 0.96 & 0.96 & 0.61 & 0.87 & 0.92 & 0.93 & 0.95 & 0.96 \\
ion & 0.98 & 0.98 & 0.98 & 0.98 & 0.97 & 0.98 & 0.78 & 0.92 & 0.98 & 0.87 & 0.96 & 0.98 & 0.90 & 0.98 & \color{blue}{\textbf{0.99}} & 0.97 & 0.90 & 0.80 & 0.68 & 0.47 & 0.97 & 0.92 & 0.96 & 0.97 \\
irs & 0.93 & 0.83 & 0.97 & 0.96 & 0.96 & 0.97 & 0.99 & 0.89 & 0.94 & \color{blue}{\textbf{1.00}} & 0.88 & 0.93 & 0.99 & 0.79 & 0.80 & 0.93 & \color{blue}{\textbf{1.00}} & 0.99 & 0.73 & 0.84 & 0.29 & 0.78 & 0.88 & 0.92 \\
iso & 0.74 & 0.70 & 0.79 & 0.75 & 0.75 & 0.64 & 0.55 & 0.60 & 0.77 & 0.55 & 0.82 & \color{blue}{\textbf{0.84}} & 0.60 & 0.71 & 0.70 & 0.60 & 0.78 & 0.78 & 0.50 & 0.54 & 0.62 & 0.69 & 0.81 & 0.81 \\
kdd & 0.96 & 0.95 & 0.99 & 0.99 & \color{blue}{\textbf{1.00}} & 0.99 & 0.99 & \color{blue}{\textbf{1.00}} & \color{blue}{\textbf{1.00}} & 0.90 & 0.98 & \color{blue}{\textbf{1.00}} & 0.99 & 0.99 & \color{blue}{\textbf{1.00}} & 0.99 & \color{blue}{\textbf{1.00}} & \color{blue}{\textbf{1.00}} & 0.02 & 0.73 & 0.24 & \color{blue}{\textbf{1.00}} & 0.99 & 0.99 \\
lbr & 0.71 & 0.64 & 0.74 & 0.64 & 0.75 & 0.65 & 0.58 & 0.55 & \color{blue}{\textbf{0.78}} & 0.56 & 0.70 & \color{blue}{\textbf{0.78}} & 0.55 & 0.73 & 0.77 & 0.55 & 0.77 & \color{blue}{\textbf{0.78}} & 0.51 & 0.59 & 0.58 & 0.68 & \color{blue}{\textbf{0.78}} & \color{blue}{\textbf{0.78}} \\
ltr & 0.79 & 0.78 & 0.77 & 0.77 & 0.79 & 0.68 & 0.56 & 0.62 & 0.80 & 0.59 & \color{blue}{\textbf{0.83}} & 0.81 & 0.60 & 0.76 & 0.75 & 0.67 & 0.76 & 0.74 & 0.48 & 0.54 & 0.65 & 0.75 & 0.82 & 0.82 \\
mam & 0.89 & 0.89 & 0.89 & 0.89 & 0.89 & 0.85 & 0.84 & 0.88 & 0.88 & 0.89 & 0.85 & \color{blue}{\textbf{0.91}} & 0.86 & 0.87 & 0.89 & 0.88 & 0.78 & 0.82 & 0.75 & 0.89 & \color{blue}{\textbf{0.91}} & 0.88 & 0.90 & 0.90 \\
mgc & 0.94 & 0.91 & 0.89 & \color{blue}{\textbf{0.97}} & 0.96 & 0.94 & 0.83 & 0.90 & 0.94 & 0.82 & 0.93 & 0.94 & 0.91 & 0.96 & 0.96 & 0.96 & 0.85 & 0.84 & 0.55 & 0.52 & 0.81 & 0.85 & 0.89 & 0.90 \\
mlt & \color{blue}{\textbf{0.99}} & 0.98 & 0.98 & 0.98 & 0.98 & 0.91 & 0.73 & 0.87 & 0.98 & 0.74 & 0.98 & \color{blue}{\textbf{0.99}} & 0.83 & 0.98 & \color{blue}{\textbf{0.99}} & 0.94 & \color{blue}{\textbf{0.99}} & \color{blue}{\textbf{0.99}} & 0.47 & 0.63 & 0.74 & 0.97 & \color{blue}{\textbf{0.99}} & \color{blue}{\textbf{0.99}} \\
mnb & 0.93 & 0.90 & 0.88 & 0.92 & 0.91 & 0.81 & 0.91 & 0.81 & 0.86 & 0.92 & 0.70 & \color{blue}{\textbf{0.94}} & 0.82 & 0.90 & 0.90 & 0.86 & 0.73 & 0.83 & 0.67 & 0.65 & 0.85 & \color{blue}{\textbf{0.94}} & 0.85 & 0.88 \\
pen & 0.97 & 0.95 & 0.98 & 0.99 & 0.99 & 0.99 & 0.77 & 0.96 & 0.99 & 0.90 & \color{blue}{\textbf{1.00}} & 0.99 & 0.95 & 0.98 & 0.98 & 0.99 & 0.96 & 0.92 & 0.59 & 0.65 & 0.86 & 0.85 & 0.98 & 0.99 \\
pgb & 0.98 & 0.98 & 0.98 & \color{blue}{\textbf{0.99}} & 0.98 & 0.97 & 0.88 & 0.97 & 0.98 & 0.96 & 0.98 & 0.98 & 0.96 & \color{blue}{\textbf{0.99}} & \color{blue}{\textbf{0.99}} & 0.98 & 0.75 & 0.73 & 0.59 & 0.71 & 0.97 & 0.95 & \color{blue}{\textbf{0.99}} & \color{blue}{\textbf{0.99}} \\
pim & 0.83 & 0.78 & 0.81 & 0.86 & 0.86 & 0.83 & 0.81 & 0.83 & 0.84 & 0.81 & 0.82 & \color{blue}{\textbf{0.89}} & 0.78 & 0.86 & 0.85 & 0.84 & 0.78 & 0.81 & 0.61 & 0.69 & 0.81 & 0.84 & 0.78 & 0.78 \\
prk & \color{blue}{\textbf{0.89}} & 0.60 & 0.73 & 0.68 & 0.72 & 0.75 & 0.55 & 0.66 & 0.80 & 0.55 & 0.70 & 0.88 & 0.45 & 0.72 & 0.71 & 0.74 & 0.78 & 0.79 & 0.64 & 0.73 & 0.72 & 0.67 & 0.79 & 0.80 \\
sat & 0.95 & 0.87 & 0.96 & 0.93 & 0.94 & 0.96 & 0.95 & 0.94 & 0.97 & 0.90 & 0.98 & \color{blue}{\textbf{0.99}} & 0.95 & 0.91 & 0.93 & 0.84 & 0.97 & 0.97 & 0.74 & 0.77 & 0.82 & 0.97 & 0.95 & 0.97 \\
scc & 0.95 & 0.96 & 0.89 & 0.98 & 0.98 & 0.90 & 0.82 & 0.92 & 0.97 & 0.86 & \color{blue}{\textbf{0.99}} & 0.98 & \color{blue}{\textbf{0.99}} & \color{blue}{\textbf{0.99}} & 0.96 & 0.90 & 0.96 & 0.97 & 0.59 & 0.65 & 0.87 & 0.87 & 0.96 & 0.97 \\
seg & 0.92 & 0.91 & 0.94 & 0.90 & 0.92 & 0.95 & 0.86 & 0.90 & \color{blue}{\textbf{0.96}} & 0.93 & 0.94 & 0.95 & 0.93 & 0.92 & 0.92 & 0.93 & 0.89 & 0.89 & 0.60 & 0.62 & 0.72 & 0.87 & 0.94 & 0.95 \\
sei & 0.76 & 0.73 & 0.74 & 0.74 & 0.73 & 0.74 & 0.73 & 0.70 & 0.74 & 0.70 & 0.65 & \color{blue}{\textbf{0.77}} & 0.74 & 0.73 & 0.73 & 0.74 & 0.68 & 0.71 & 0.56 & 0.54 & 0.74 & 0.72 & 0.73 & 0.73 \\
sht & 0.99 & 0.99 & 0.99 & \color{blue}{\textbf{1.00}} & 0.98 & \color{blue}{\textbf{1.00}} & 0.93 & 0.98 & \color{blue}{\textbf{1.00}} & 0.90 & \color{blue}{\textbf{1.00}} & \color{blue}{\textbf{1.00}} & 0.99 & \color{blue}{\textbf{1.00}} & 0.99 & \color{blue}{\textbf{1.00}} & 0.85 & 0.87 & 0.65 & 0.76 & 0.93 & 0.99 & \color{blue}{\textbf{1.00}} & \color{blue}{\textbf{1.00}} \\
snr & 0.73 & 0.65 & 0.76 & 0.65 & 0.75 & 0.64 & 0.49 & 0.55 & 0.64 & 0.52 & 0.85 & 0.84 & 0.50 & 0.65 & 0.66 & 0.58 & 0.81 & 0.81 & 0.57 & 0.56 & 0.47 & 0.73 & 0.74 & \color{blue}{\textbf{0.86}} \\
sph & 0.80 & 0.35 & 0.52 & 0.69 & 0.28 & 0.35 & 0.30 & 0.35 & 0.50 & 0.47 & 0.40 & \color{blue}{\textbf{0.82}} & 0.28 & 0.26 & 0.30 & 0.28 & 0.74 & 0.80 & 0.55 & 0.72 & 0.50 & 0.50 & 0.50 & 0.80 \\
spm & 0.77 & 0.80 & 0.78 & 0.86 & 0.87 & 0.77 & 0.82 & 0.82 & 0.78 & 0.63 & 0.81 & \color{blue}{\textbf{0.94}} & 0.84 & 0.86 & 0.85 & 0.83 & 0.91 & 0.91 & 0.54 & 0.64 & 0.60 & 0.83 & 0.54 & 0.81 \\
vhc & \color{blue}{\textbf{0.80}} & 0.65 & 0.73 & 0.77 & 0.77 & 0.74 & 0.77 & 0.69 & 0.73 & 0.70 & 0.60 & 0.72 & 0.70 & 0.75 & 0.77 & 0.74 & 0.65 & 0.70 & 0.57 & 0.64 & 0.70 & 0.68 & 0.63 & 0.71 \\
wf1 & 0.81 & 0.71 & 0.78 & 0.89 & 0.87 & 0.72 & 0.87 & 0.83 & 0.83 & 0.81 & 0.75 & \color{blue}{\textbf{0.95}} & 0.85 & 0.75 & 0.75 & 0.77 & 0.85 & 0.84 & 0.63 & 0.80 & 0.75 & 0.92 & 0.82 & 0.92 \\
wf2 & 0.92 & 0.73 & 0.75 & 0.90 & 0.91 & 0.73 & 0.86 & 0.84 & 0.84 & 0.82 & 0.76 & \color{blue}{\textbf{0.94}} & 0.85 & 0.74 & 0.79 & 0.77 & 0.87 & 0.84 & 0.60 & 0.80 & 0.75 & 0.92 & 0.79 & \color{blue}{\textbf{0.94}} \\
wne & \color{blue}{\textbf{1.00}} & 0.98 & 0.96 & 0.99 & 0.98 & 0.95 & 0.92 & 0.91 & 0.98 & 0.83 & 0.97 & 0.99 & 0.73 & 0.98 & 0.95 & 0.96 & 0.97 & 0.92 & 0.61 & 0.66 & 0.94 & 0.93 & 0.95 & 0.99 \\
wrb & 0.73 & 0.73 & 0.81 & 0.86 & 0.85 & 0.80 & 0.87 & 0.81 & 0.82 & 0.72 & 0.76 & 0.85 & \color{blue}{\textbf{0.91}} & 0.78 & 0.82 & 0.81 & 0.82 & 0.78 & 0.56 & 0.61 & 0.57 & 0.66 & 0.80 & 0.80 \\
yst & 0.74 & 0.70 & 0.66 & \color{blue}{\textbf{0.75}} & 0.73 & 0.66 & 0.53 & 0.63 & 0.66 & 0.67 & 0.68 & \color{blue}{\textbf{0.75}} & 0.60 & 0.72 & 0.72 & 0.67 & 0.62 & 0.65 & 0.65 & 0.69 & 0.70 & 0.68 & 0.54 & 0.64 \\\midrule
avg. AUC & 0.88 &  0.81 &  0.85 &  0.87 &  0.86 &  0.82 &  0.78 &  0.81 &  0.85 &  0.78 &  0.84 &  \color{blue}{\textbf{0.91}} &  0.79 &  0.85 &  0.85 &  0.83 &  0.84 &  0.84 &  0.6  & 0.68 &  0.75 &  0.84 &  0.83 &  0.88 \\
avg. rank & 6.9 & 13.4 & 10.3 & 6.5 & 7.0 & 11.4 & 14.9 & 14.5 & 8.4 & 16.1 & 11.4 & \color{red}{\textbf{2.9}} & 14.3 & 8.9 & 9.0 & 10.7 & 11.4 & 12.0 & 22.3 & 19.8 & 16.2 & 12.1 & 11.0 & 6.8 \\\bottomrule
\end{tabular}

    \vspace*{0.15cm}
    \caption{Performance of models on \textit{tabular} datasets with hyperparameter selection using \textit{50\% of all available anomalies} in the validation set, reported in the AUC metric on the test data, averaged over 5 random cross-validation repetitions. The final row contains the average model rank.}
    \label{tab:tabular_anomalies}
\end{table*}

\begin{table*}
    \footnotesize
    \centering
    \tabcolsep=0.05cm
    \begin{tabular}{rrrrrrrrrrrrrrrrrrrrrrrrr}
\toprule
\textbf{dataset} & \textbf{aae} & \textbf{avae} & \textbf{gano} & \textbf{vae} & \textbf{wae} & \textbf{abod} & \textbf{hbos} & \textbf{if} & \textbf{knn} & \textbf{loda} & \textbf{lof} & \textbf{osvm} & \textbf{pidf} & \textbf{maf} & \textbf{rnvp} & \textbf{sptn} & \textbf{fmgn} & \textbf{gan} & \textbf{mgal} & \textbf{dagm} & \textbf{dsvd} & \textbf{rpn} & \textbf{vaek} & \textbf{vaeo} \\\midrule
aba & 0.87 & 0.76 & 0.79 & 0.90 & 0.90 & \color{blue}{\textbf{0.93}} & 0.77 & 0.86 & \color{blue}{\textbf{0.93}} & 0.27 & 0.89 & 0.91 & 0.90 & 0.85 & 0.89 & 0.89 & 0.37 & 0.39 & 0.38 & 0.67 & 0.82 & 0.50 & 0.89 & 0.89 \\
ann & 0.82 & 0.58 & 0.69 & 0.79 & 0.76 & 0.67 & 0.66 & 0.74 & 0.67 & 0.43 & 0.64 & 0.64 & \color{blue}{\textbf{0.92}} & 0.79 & 0.80 & 0.80 & 0.60 & 0.43 & 0.51 & 0.48 & 0.56 & 0.50 & 0.71 & 0.53 \\
arr & 0.35 & 0.72 & 0.69 & 0.67 & 0.75 & 0.75 & 0.77 & \color{blue}{\textbf{0.78}} & 0.75 & 0.73 & 0.75 & 0.75 & 0.69 & 0.74 & 0.75 & 0.74 & 0.50 & 0.43 & 0.43 & 0.50 & 0.47 & 0.63 & 0.71 & 0.71 \\
bcw & 0.95 & 0.70 & 0.90 & 0.95 & 0.96 & 0.94 & 0.97 & 0.97 & 0.94 & 0.93 & 0.94 & 0.94 & 0.76 & 0.97 & \color{blue}{\textbf{0.98}} & 0.94 & 0.97 & 0.86 & 0.33 & 0.57 & 0.76 & 0.53 & 0.92 & 0.91 \\
blt & 0.67 & 0.88 & 0.87 & 0.89 & 0.88 & 0.91 & 0.90 & 0.94 & 0.93 & 0.68 & \color{blue}{\textbf{0.95}} & 0.93 & 0.84 & 0.92 & 0.91 & 0.88 & 0.46 & 0.43 & 0.32 & 0.54 & 0.92 & 0.50 & 0.93 & 0.92 \\
bts & \color{blue}{\textbf{0.99}} & 0.69 & 0.90 & \color{blue}{\textbf{0.99}} & 0.96 & 0.96 & 0.98 & 0.98 & 0.98 & 0.92 & 0.94 & 0.98 & 0.30 & \color{blue}{\textbf{0.99}} & \color{blue}{\textbf{0.99}} & \color{blue}{\textbf{0.99}} & 0.83 & 0.83 & 0.54 & 0.78 & 0.92 & 0.50 & 0.87 & 0.88 \\
crd & 0.42 & 0.61 & 0.46 & 0.40 & 0.45 & 0.47 & 0.48 & 0.66 & 0.46 & 0.49 & 0.59 & 0.53 & 0.57 & 0.43 & 0.45 & 0.49 & \color{blue}{\textbf{0.67}} & 0.51 & 0.50 & 0.45 & 0.62 & 0.50 & 0.52 & 0.56 \\
eco & 0.74 & 0.58 & 0.79 & 0.83 & 0.80 & \color{blue}{\textbf{0.86}} & 0.72 & 0.81 & 0.85 & 0.64 & 0.80 & 0.85 & 0.76 & 0.83 & 0.85 & 0.79 & 0.75 & 0.67 & 0.64 & 0.54 & 0.52 & 0.50 & 0.79 & 0.79 \\
gls & \color{blue}{\textbf{0.86}} & 0.44 & 0.68 & 0.64 & 0.60 & 0.79 & 0.60 & 0.56 & 0.72 & 0.56 & 0.78 & 0.76 & 0.40 & 0.66 & 0.72 & 0.76 & 0.60 & 0.72 & 0.44 & 0.45 & 0.47 & 0.50 & 0.57 & 0.49 \\
hab & 0.91 & 0.78 & 0.77 & 0.95 & 0.93 & 0.95 & 0.91 & 0.94 & 0.95 & 0.37 & \color{blue}{\textbf{0.96}} & 0.95 & 0.95 & 0.95 & 0.95 & 0.93 & 0.48 & 0.49 & 0.58 & 0.74 & 0.81 & 0.50 & 0.79 & 0.73 \\
har & 0.65 & 0.60 & 0.78 & 0.38 & 0.55 & 0.66 & 0.39 & 0.60 & 0.69 & 0.47 & \color{blue}{\textbf{0.93}} & 0.64 & 0.32 & 0.48 & 0.56 & \color{blue}{\textbf{0.93}} & 0.85 & 0.92 & 0.53 & 0.56 & 0.46 & 0.50 & 0.33 & 0.29 \\
htr & 0.93 & 0.86 & 0.94 & 0.94 & \color{blue}{\textbf{0.95}} & - & \color{blue}{\textbf{0.95}} & \color{blue}{\textbf{0.95}} & 0.94 & 0.35 & 0.94 & 0.93 & 0.92 & \color{blue}{\textbf{0.95}} & 0.94 & \color{blue}{\textbf{0.95}} & 0.79 & 0.93 & 0.26 & 0.84 & 0.91 & 0.50 & 0.94 & 0.94 \\
ion & 0.94 & 0.82 & \color{blue}{\textbf{0.98}} & 0.96 & 0.95 & \color{blue}{\textbf{0.98}} & 0.73 & 0.92 & \color{blue}{\textbf{0.98}} & 0.92 & 0.96 & \color{blue}{\textbf{0.98}} & 0.87 & 0.97 & 0.96 & 0.97 & 0.48 & 0.67 & 0.34 & 0.42 & 0.91 & 0.50 & 0.91 & 0.87 \\
irs & 0.60 & 0.72 & 0.90 & 0.75 & 0.78 & 0.93 & 0.97 & 0.85 & 0.89 & 0.12 & 0.83 & 0.84 & \color{blue}{\textbf{0.99}} & 0.75 & 0.80 & 0.93 & 0.95 & 0.97 & 0.59 & 0.68 & 0.04 & 0.50 & 0.88 & 0.64 \\
iso & 0.44 & 0.48 & 0.55 & 0.42 & 0.65 & 0.61 & 0.54 & 0.54 & 0.70 & 0.53 & 0.74 & \color{blue}{\textbf{0.75}} & 0.54 & 0.63 & 0.64 & 0.60 & 0.53 & 0.55 & 0.44 & 0.49 & 0.37 & 0.50 & 0.71 & 0.73 \\
kdd & 0.93 & 0.22 & 0.95 & 0.97 & 0.99 & - & 0.96 & 0.99 & 0.95 & 0.15 & 0.97 & 0.98 & 0.91 & 0.99 & \color{blue}{\textbf{1.00}} & 0.97 & 0.98 & 0.98 & 0.01 & 0.57 & 0.02 & 0.50 & - & 0.88 \\
lbr & 0.42 & 0.59 & 0.66 & 0.65 & 0.68 & 0.66 & 0.56 & 0.55 & 0.62 & 0.57 & 0.56 & 0.76 & 0.54 & 0.71 & 0.64 & 0.53 & 0.48 & 0.52 & 0.49 & 0.49 & 0.53 & 0.50 & 0.65 & \color{blue}{\textbf{0.78}} \\
ltr & 0.49 & 0.53 & 0.66 & 0.46 & 0.70 & 0.68 & 0.54 & 0.58 & 0.75 & 0.51 & 0.78 & \color{blue}{\textbf{0.79}} & 0.56 & 0.68 & 0.68 & 0.67 & 0.50 & 0.50 & 0.48 & 0.46 & 0.42 & 0.50 & 0.78 & \color{blue}{\textbf{0.79}} \\
mam & 0.74 & 0.78 & 0.87 & 0.84 & 0.81 & 0.85 & 0.85 & 0.87 & 0.87 & 0.18 & 0.83 & 0.86 & 0.83 & 0.84 & 0.87 & 0.87 & 0.36 & 0.48 & 0.31 & 0.67 & 0.88 & 0.50 & \color{blue}{\textbf{0.90}} & 0.89 \\
mgc & 0.91 & 0.88 & 0.78 & 0.95 & 0.93 & 0.89 & 0.80 & 0.85 & 0.91 & 0.24 & 0.90 & 0.86 & 0.86 & \color{blue}{\textbf{0.96}} & 0.95 & 0.95 & 0.74 & 0.80 & 0.49 & 0.41 & 0.70 & 0.50 & 0.84 & 0.82 \\
mlt & 0.49 & 0.55 & 0.76 & 0.45 & 0.96 & 0.89 & 0.73 & 0.80 & 0.96 & 0.63 & 0.96 & \color{blue}{\textbf{0.98}} & 0.76 & 0.96 & 0.95 & 0.94 & 0.50 & 0.51 & 0.45 & 0.55 & 0.38 & 0.50 & 0.83 & 0.92 \\
mnb & 0.85 & 0.78 & 0.76 & 0.83 & 0.76 & - & 0.81 & 0.80 & 0.85 & 0.12 & 0.81 & \color{blue}{\textbf{0.88}} & 0.79 & 0.70 & 0.79 & 0.75 & 0.49 & 0.50 & 0.23 & 0.41 & 0.62 & 0.50 & 0.83 & 0.84 \\
pen & 0.97 & 0.67 & 0.97 & 0.98 & 0.97 & 0.96 & 0.77 & 0.91 & 0.97 & 0.63 & 0.98 & 0.96 & 0.89 & 0.97 & 0.98 & \color{blue}{\textbf{0.99}} & 0.73 & 0.74 & 0.52 & 0.58 & 0.65 & 0.50 & 0.78 & 0.70 \\
pgb & 0.95 & 0.92 & 0.94 & 0.97 & 0.97 & 0.98 & 0.80 & 0.95 & 0.98 & 0.25 & 0.98 & 0.97 & 0.92 & 0.98 & 0.98 & 0.98 & 0.49 & 0.61 & 0.25 & 0.71 & 0.95 & 0.85 & \color{blue}{\textbf{0.99}} & \color{blue}{\textbf{0.99}} \\
pim & 0.73 & 0.80 & 0.78 & 0.84 & 0.73 & \color{blue}{\textbf{0.85}} & 0.82 & \color{blue}{\textbf{0.85}} & 0.83 & 0.62 & 0.75 & 0.83 & 0.79 & 0.83 & 0.81 & 0.79 & 0.62 & 0.53 & 0.53 & 0.51 & 0.71 & 0.68 & 0.69 & 0.71 \\
prk & 0.77 & 0.59 & 0.66 & 0.62 & 0.70 & 0.72 & 0.48 & 0.64 & 0.69 & \color{blue}{\textbf{0.81}} & 0.67 & 0.74 & 0.37 & 0.73 & 0.68 & 0.69 & 0.46 & 0.51 & 0.48 & 0.27 & 0.47 & 0.50 & 0.74 & 0.77 \\
sat & 0.77 & 0.59 & 0.90 & 0.82 & 0.85 & 0.92 & 0.94 & 0.91 & 0.94 & 0.48 & 0.95 & 0.93 & 0.93 & 0.85 & 0.80 & 0.76 & 0.90 & \color{blue}{\textbf{0.96}} & 0.74 & 0.55 & 0.66 & 0.50 & 0.92 & 0.89 \\
scc & 0.83 & 0.59 & 0.89 & 0.54 & 0.86 & 0.82 & 0.78 & 0.89 & 0.97 & 0.80 & \color{blue}{\textbf{0.99}} & 0.98 & 0.90 & 0.95 & 0.91 & 0.89 & 0.59 & 0.47 & 0.49 & 0.56 & 0.87 & 0.50 & 0.97 & 0.95 \\
seg & 0.83 & 0.39 & 0.87 & 0.88 & 0.88 & \color{blue}{\textbf{0.92}} & 0.80 & 0.82 & \color{blue}{\textbf{0.92}} & 0.28 & 0.89 & 0.91 & 0.87 & 0.86 & 0.86 & 0.88 & 0.80 & 0.79 & 0.44 & 0.43 & 0.19 & 0.50 & 0.91 & 0.90 \\
sei & 0.70 & \color{blue}{\textbf{0.78}} & 0.71 & 0.72 & 0.71 & 0.74 & 0.73 & 0.70 & 0.72 & 0.30 & 0.65 & 0.72 & 0.73 & 0.72 & 0.74 & 0.71 & 0.55 & 0.49 & 0.40 & 0.53 & 0.47 & 0.50 & 0.68 & 0.68 \\
sht & 0.99 & 0.82 & 0.97 & 0.98 & 0.97 & - & 0.89 & 0.96 & 0.98 & 0.03 & 0.95 & 0.97 & 0.95 & \color{blue}{\textbf{1.00}} & 0.99 & \color{blue}{\textbf{1.00}} & 0.48 & 0.15 & 0.46 & 0.40 & 0.89 & 0.50 & 0.96 & 0.97 \\
snr & 0.58 & 0.49 & 0.59 & 0.57 & 0.59 & 0.62 & 0.53 & 0.60 & 0.61 & 0.57 & 0.55 & 0.63 & 0.52 & 0.56 & 0.64 & 0.57 & 0.61 & \color{blue}{\textbf{0.65}} & 0.49 & 0.51 & 0.42 & 0.50 & 0.61 & 0.60 \\
sph & \color{blue}{\textbf{0.64}} & 0.34 & 0.42 & 0.49 & 0.30 & 0.25 & 0.25 & 0.29 & 0.23 & 0.55 & 0.22 & 0.24 & 0.21 & 0.24 & 0.23 & 0.25 & 0.53 & 0.51 & 0.46 & 0.48 & 0.32 & 0.50 & 0.25 & 0.28 \\
spm & 0.69 & 0.54 & 0.57 & 0.83 & \color{blue}{\textbf{0.84}} & 0.73 & 0.80 & 0.83 & 0.68 & 0.40 & 0.66 & 0.70 & 0.81 & 0.83 & 0.83 & 0.83 & 0.82 & 0.63 & 0.46 & 0.49 & 0.58 & 0.50 & 0.38 & 0.28 \\
vhc & 0.67 & 0.61 & 0.64 & 0.75 & 0.72 & 0.72 & 0.71 & 0.67 & 0.68 & 0.56 & 0.58 & 0.60 & 0.66 & \color{blue}{\textbf{0.76}} & 0.73 & 0.73 & 0.62 & 0.49 & 0.48 & 0.50 & 0.47 & 0.50 & 0.56 & 0.53 \\
wf1 & 0.72 & 0.51 & 0.50 & 0.59 & 0.68 & 0.69 & \color{blue}{\textbf{0.85}} & 0.80 & 0.76 & 0.30 & 0.74 & 0.77 & 0.74 & 0.75 & 0.63 & 0.75 & 0.69 & 0.69 & 0.63 & 0.37 & 0.72 & 0.50 & 0.71 & 0.66 \\
wf2 & 0.65 & 0.53 & 0.64 & 0.44 & 0.65 & 0.69 & \color{blue}{\textbf{0.85}} & 0.81 & 0.77 & 0.35 & 0.74 & 0.78 & 0.75 & 0.74 & 0.67 & 0.77 & 0.71 & 0.70 & 0.48 & 0.47 & 0.72 & 0.50 & 0.69 & 0.65 \\
wne & 0.34 & 0.45 & 0.92 & 0.25 & 0.92 & 0.97 & 0.89 & 0.87 & 0.98 & 0.91 & 0.98 & \color{blue}{\textbf{0.99}} & 0.56 & 0.91 & 0.88 & 0.93 & 0.93 & 0.85 & 0.61 & 0.66 & 0.94 & 0.50 & 0.94 & 0.96 \\
wrb & 0.73 & 0.48 & 0.59 & 0.81 & 0.75 & 0.70 & 0.72 & 0.69 & 0.67 & 0.44 & 0.66 & 0.70 & 0.74 & 0.78 & \color{blue}{\textbf{0.82}} & 0.77 & 0.74 & 0.76 & 0.56 & 0.50 & 0.52 & 0.50 & 0.66 & 0.67 \\
yst & 0.42 & 0.55 & 0.60 & 0.63 & 0.58 & 0.67 & 0.55 & 0.63 & 0.68 & 0.51 & \color{blue}{\textbf{0.70}} & 0.67 & 0.61 & 0.68 & 0.62 & 0.61 & 0.35 & 0.40 & 0.47 & 0.58 & 0.62 & 0.50 & 0.54 & 0.50 \\\midrule
avg. AUC & 0.72 &  0.63 &  0.76 &  0.73 &  0.78 &  0.79 &  0.74 &  0.78 &  \color{blue}{\textbf{0.81}} &  0.49 &  0.80  & \color{blue}{\textbf{0.81}} &  0.72 &  0.80  & 0.80  & 0.80  & 0.64 &  0.63 &  0.46 &  0.53 &  0.61 &  0.52 &  0.75 &  0.74 \\
avg. rank & 12.5 & 16.2 & 11.6 & 10.4 & 9.0 & 8.6 & 11.3 & 8.6 & \color{red}{\textbf{6.0}} & 18.3 & 8.0 & \color{red}{\textbf{6.0}} & 12.5 & 7.2 & 7.1 & 7.4 & 14.8 & 15.4 & 20.6 & 19.7 & 16.6 & 19.5 & 10.7 & 11.6 \\\bottomrule
\end{tabular}
 
    \vspace*{0.15cm}
    \caption{Performance of models on \textit{tabular} datasets with hyperparameter selection \textit{without anomalies}. Where available, metrics akin to likelihood on validation data without anomalies were used instead. Reported values correspond to the AUC metric on the test data, averaged over 5 random cross-validation repetitions. The final row contains the average model rank. Missing values indicate a failure of a particular method to train with default parameters within the given computational budget.}
    \label{tab:tabular_clean}
\end{table*}

\begin{table*}
    \centering
    \begin{subtable}{.49\linewidth}
        \footnotesize
        \tabcolsep=0.05cm
        \begin{tabular}{rrrrrrrrrrrrr}
\toprule
\textbf{dataset} & \textbf{aae} & \textbf{gano} & \textbf{skip} & \textbf{vae} & \textbf{wae} & \textbf{knn} & \textbf{osvm} & \textbf{fano} & \textbf{fmgn} & \textbf{dsvd} & \textbf{vaek} & \textbf{vaeo} \\\midrule
mnist:0 & \color{blue}{\textbf{1.00}} & \color{blue}{\textbf{1.00}} & 0.96 & \color{blue}{\textbf{1.00}} & \color{blue}{\textbf{1.00}} & \color{blue}{\textbf{1.00}} & \color{blue}{\textbf{1.00}} & \color{blue}{\textbf{1.00}} & 0.96 & \color{blue}{\textbf{1.00}} & \color{blue}{\textbf{1.00}} & 0.99 \\
mnist:1 & \color{blue}{\textbf{1.00}} & \color{blue}{\textbf{1.00}} & \color{blue}{\textbf{1.00}} & \color{blue}{\textbf{1.00}} & \color{blue}{\textbf{1.00}} & \color{blue}{\textbf{1.00}} & \color{blue}{\textbf{1.00}} & \color{blue}{\textbf{1.00}} & 0.99 & \color{blue}{\textbf{1.00}} & \color{blue}{\textbf{1.00}} & \color{blue}{\textbf{1.00}} \\
mnist:2 & 0.96 & 0.97 & 0.86 & 0.96 & 0.96 & 0.93 & 0.93 & 0.94 & 0.89 & \color{blue}{\textbf{0.98}} & 0.94 & 0.94 \\
mnist:3 & \color{blue}{\textbf{0.97}} & \color{blue}{\textbf{0.97}} & 0.87 & \color{blue}{\textbf{0.97}} & \color{blue}{\textbf{0.97}} & 0.96 & 0.94 & 0.95 & 0.74 & \color{blue}{\textbf{0.97}} & \color{blue}{\textbf{0.97}} & 0.96 \\
mnist:4 & \color{blue}{\textbf{0.98}} & 0.97 & 0.92 & \color{blue}{\textbf{0.98}} & 0.97 & 0.96 & 0.96 & 0.96 & 0.85 & \color{blue}{\textbf{0.98}} & 0.97 & 0.96 \\
mnist:5 & \color{blue}{\textbf{0.98}} & \color{blue}{\textbf{0.98}} & 0.85 & \color{blue}{\textbf{0.98}} & \color{blue}{\textbf{0.98}} & 0.97 & 0.96 & 0.96 & 0.78 & 0.97 & 0.97 & 0.97 \\
mnist:6 & 0.99 & \color{blue}{\textbf{1.00}} & 0.96 & \color{blue}{\textbf{1.00}} & 0.99 & 0.99 & 0.98 & 0.99 & 0.93 & \color{blue}{\textbf{1.00}} & 0.99 & 0.99 \\
mnist:7 & 0.98 & \color{blue}{\textbf{0.99}} & 0.95 & 0.98 & 0.98 & 0.97 & 0.96 & 0.97 & 0.96 & 0.98 & 0.97 & 0.96 \\
mnist:8 & 0.92 & 0.96 & 0.81 & 0.92 & 0.92 & 0.91 & 0.90 & 0.94 & 0.80 & \color{blue}{\textbf{0.97}} & 0.91 & 0.92 \\
mnist:9 & 0.98 & 0.98 & 0.93 & 0.98 & 0.98 & 0.97 & 0.96 & 0.98 & 0.91 & \color{blue}{\textbf{0.99}} & 0.98 & 0.98 \\ \midrule
boot & \color{blue}{\textbf{0.99}} & \color{blue}{\textbf{0.99}} & \color{blue}{\textbf{0.99}} & 0.98 & 0.98 & 0.98 & 0.98 & \color{blue}{\textbf{0.99}} & 0.98 & \color{blue}{\textbf{0.99}} & 0.93 & 0.98 \\
bag & 0.92 & 0.95 & 0.87 & 0.91 & 0.93 & 0.90 & 0.90 & 0.92 & 0.70 & \color{blue}{\textbf{0.96}} & 0.88 & 0.91 \\
coat & 0.91 & \color{blue}{\textbf{0.93}} & 0.91 & 0.91 & 0.91 & 0.92 & 0.91 & \color{blue}{\textbf{0.93}} & 0.85 & \color{blue}{\textbf{0.93}} & 0.91 & 0.92 \\
dress & 0.95 & \color{blue}{\textbf{0.96}} & 0.95 & 0.94 & 0.95 & 0.94 & 0.95 & \color{blue}{\textbf{0.96}} & 0.90 & \color{blue}{\textbf{0.96}} & 0.93 & 0.92 \\
pullover & 0.90 & 0.91 & 0.88 & 0.90 & 0.90 & 0.90 & 0.89 & 0.90 & 0.81 & \color{blue}{\textbf{0.92}} & 0.89 & 0.87 \\
sandal & 0.92 & 0.94 & 0.94 & 0.91 & 0.91 & 0.92 & 0.91 & 0.94 & 0.92 & \color{blue}{\textbf{0.96}} & 0.89 & 0.90 \\
shirt & 0.84 & \color{blue}{\textbf{0.87}} & 0.81 & 0.84 & 0.84 & 0.85 & 0.84 & 0.86 & 0.59 & \color{blue}{\textbf{0.87}} & 0.85 & 0.83 \\
sneaker & 0.98 & \color{blue}{\textbf{0.99}} & \color{blue}{\textbf{0.99}} & \color{blue}{\textbf{0.99}} & \color{blue}{\textbf{0.99}} & 0.98 & 0.98 & \color{blue}{\textbf{0.99}} & 0.96 & \color{blue}{\textbf{0.99}} & 0.96 & 0.98 \\
t-shirt & 0.93 & 0.94 & 0.91 & 0.93 & 0.93 & 0.93 & 0.93 & 0.94 & 0.72 & \color{blue}{\textbf{0.95}} & 0.92 & 0.93 \\
trouser & \color{blue}{\textbf{0.99}} & \color{blue}{\textbf{0.99}} & \color{blue}{\textbf{0.99}} & \color{blue}{\textbf{0.99}} & \color{blue}{\textbf{0.99}} & \color{blue}{\textbf{0.99}} & \color{blue}{\textbf{0.99}} & \color{blue}{\textbf{0.99}} & 0.97 & \color{blue}{\textbf{0.99}} & \color{blue}{\textbf{0.99}} & \color{blue}{\textbf{0.99}} \\ \midrule
bright & \color{blue}{\textbf{1.00}} & 0.93 & 0.00 & \color{blue}{\textbf{1.00}} & \color{blue}{\textbf{1.00}} & \color{blue}{\textbf{1.00}} & \color{blue}{\textbf{1.00}} & \color{blue}{\textbf{1.00}} & \color{blue}{\textbf{1.00}} & \color{blue}{\textbf{1.00}} & \color{blue}{\textbf{1.00}} & \color{blue}{\textbf{1.00}} \\
cannye & \color{blue}{\textbf{1.00}} & 0.84 & 0.63 & \color{blue}{\textbf{1.00}} & \color{blue}{\textbf{1.00}} & \color{blue}{\textbf{1.00}} & \color{blue}{\textbf{1.00}} & \color{blue}{\textbf{1.00}} & 0.98 & 0.99 & \color{blue}{\textbf{1.00}} & \color{blue}{\textbf{1.00}} \\
dottedl & 0.99 & 0.64 & 0.43 & \color{blue}{\textbf{1.00}} & \color{blue}{\textbf{1.00}} & 0.92 & 0.88 & \color{blue}{\textbf{1.00}} & 0.98 & 0.89 & 0.96 & 0.92 \\
fog & \color{blue}{\textbf{1.00}} & 0.15 & 0.00 & \color{blue}{\textbf{1.00}} & \color{blue}{\textbf{1.00}} & 0.99 & 0.97 & \color{blue}{\textbf{1.00}} & \color{blue}{\textbf{1.00}} & \color{blue}{\textbf{1.00}} & 0.99 & \color{blue}{\textbf{1.00}} \\
glassb & 0.97 & 0.45 & 0.14 & 0.98 & \color{blue}{\textbf{1.00}} & 0.76 & 0.71 & 0.97 & \color{blue}{\textbf{1.00}} & 0.58 & 0.71 & \color{blue}{\textbf{1.00}} \\
impulsn & \color{blue}{\textbf{1.00}} & 0.78 & 0.11 & \color{blue}{\textbf{1.00}} & \color{blue}{\textbf{1.00}} & \color{blue}{\textbf{1.00}} & \color{blue}{\textbf{1.00}} & \color{blue}{\textbf{1.00}} & \color{blue}{\textbf{1.00}} & \color{blue}{\textbf{1.00}} & \color{blue}{\textbf{1.00}} & \color{blue}{\textbf{1.00}} \\
motionb & 0.80 & 0.44 & 0.20 & 0.80 & 0.98 & 0.72 & 0.71 & 0.99 & \color{blue}{\textbf{1.00}} & 0.63 & 0.90 & 0.99 \\
rotate & 0.61 & 0.51 & 0.37 & 0.65 & 0.64 & 0.65 & 0.64 & 0.68 & \color{blue}{\textbf{0.95}} & 0.59 & 0.68 & 0.87 \\
scale & 0.53 & 0.88 & 0.91 & 0.58 & 0.59 & 0.57 & 0.95 & 0.76 & 0.94 & 0.79 & 0.91 & \color{blue}{\textbf{0.99}} \\
shear & 0.64 & 0.55 & 0.28 & 0.83 & 0.93 & 0.65 & 0.64 & 0.77 & \color{blue}{\textbf{0.97}} & 0.72 & 0.62 & 0.84 \\
shotn & 0.99 & 0.58 & 0.56 & \color{blue}{\textbf{1.00}} & \color{blue}{\textbf{1.00}} & 0.84 & 0.73 & 0.99 & 0.87 & 0.58 & 0.91 & 0.96 \\
spatter & 0.96 & 0.46 & 0.22 & 0.97 & \color{blue}{\textbf{1.00}} & 0.70 & 0.55 & 0.91 & \color{blue}{\textbf{1.00}} & 0.68 & 0.73 & 0.93 \\
stripe & \color{blue}{\textbf{1.00}} & \color{blue}{\textbf{1.00}} & \color{blue}{\textbf{1.00}} & \color{blue}{\textbf{1.00}} & \color{blue}{\textbf{1.00}} & \color{blue}{\textbf{1.00}} & \color{blue}{\textbf{1.00}} & \color{blue}{\textbf{1.00}} & \color{blue}{\textbf{1.00}} & \color{blue}{\textbf{1.00}} & \color{blue}{\textbf{1.00}} & \color{blue}{\textbf{1.00}} \\
translt & 0.99 & 0.70 & 0.62 & \color{blue}{\textbf{1.00}} & 0.99 & 0.99 & 0.97 & 0.97 & 0.72 & 0.90 & 0.99 & 0.99 \\ \midrule
grid & 0.50 & 0.62 & 0.62 & 0.60 & 0.61 & 0.48 & 0.53 & \color{blue}{\textbf{0.70}} & 0.67 & 0.60 & 0.38 & 0.61 \\
transtr & 0.79 & 0.74 & 0.69 & \color{blue}{\textbf{0.81}} & 0.80 & 0.76 & 0.75 & 0.78 & 0.77 & 0.70 & 0.74 & 0.73 \\
wood & \color{blue}{\textbf{0.77}} & 0.57 & 0.54 & \color{blue}{\textbf{0.77}} & 0.75 & \color{blue}{\textbf{0.77}} & 0.75 & 0.69 & 0.71 & 0.56 & 0.69 & 0.65 \\\midrule
avg. AUC & 0.91  & 0.81  & 0.69  & 0.92  & \color{blue}{\textbf{0.93}}  & 0.89  & 0.88  & \color{blue}{\textbf{0.93}}  & 0.89  & 0.88  & 0.89  & \color{blue}{\textbf{0.93}} \\
avg. rank & 3.6 & 5.1 & 9.1 & 3.1 & \color{red}{\textbf{2.9}} & 5.4 & 6.4 & 3.4 & 7.5 & 4.0 & 5.7 & 4.9 \\\bottomrule
\end{tabular}

        \vspace*{0.15cm}
        \caption{ 50\% anomalies in validation}
    \end{subtable}
    \begin{subtable}{.49\linewidth}
        \footnotesize
        \tabcolsep=0.05cm
        \begin{tabular}{rrrrrrrrrrrrr}
\toprule
\textbf{dataset} & \textbf{aae} & \textbf{gano} & \textbf{skip} & \textbf{vae} & \textbf{wae} & \textbf{knn} & \textbf{osvm} & \textbf{fano} & \textbf{fmgn} & \textbf{dsvd} & \textbf{vaek} & \textbf{vaeo} \\\midrule
mnist:0 & 0.99 & \color{blue}{\textbf{1.00}} & 0.94 & 0.99 & 0.99 & \color{blue}{\textbf{1.00}} & 0.99 & 0.98 & 0.50 & \color{blue}{\textbf{1.00}} & \color{blue}{\textbf{1.00}} & 0.99 \\
mnist:1 & \color{blue}{\textbf{1.00}} & \color{blue}{\textbf{1.00}} & \color{blue}{\textbf{1.00}} & \color{blue}{\textbf{1.00}} & \color{blue}{\textbf{1.00}} & \color{blue}{\textbf{1.00}} & \color{blue}{\textbf{1.00}} & \color{blue}{\textbf{1.00}} & 0.50 & \color{blue}{\textbf{1.00}} & \color{blue}{\textbf{1.00}} & \color{blue}{\textbf{1.00}} \\
mnist:2 & 0.92 & \color{blue}{\textbf{0.97}} & 0.79 & 0.88 & 0.88 & 0.91 & 0.88 & 0.82 & 0.50 & 0.86 & 0.92 & 0.92 \\
mnist:3 & 0.93 & \color{blue}{\textbf{0.96}} & 0.77 & 0.91 & 0.92 & 0.94 & 0.91 & 0.89 & 0.50 & 0.95 & \color{blue}{\textbf{0.96}} & \color{blue}{\textbf{0.96}} \\
mnist:4 & 0.95 & \color{blue}{\textbf{0.96}} & 0.89 & 0.91 & 0.93 & 0.95 & 0.92 & 0.91 & 0.50 & 0.95 & 0.95 & 0.95 \\
mnist:5 & \color{blue}{\textbf{0.96}} & \color{blue}{\textbf{0.96}} & 0.75 & 0.93 & 0.92 & 0.94 & 0.91 & 0.91 & 0.50 & 0.92 & 0.93 & 0.92 \\
mnist:6 & \color{blue}{\textbf{0.99}} & \color{blue}{\textbf{0.99}} & 0.91 & 0.98 & 0.98 & \color{blue}{\textbf{0.99}} & 0.97 & 0.98 & 0.50 & 0.97 & \color{blue}{\textbf{0.99}} & \color{blue}{\textbf{0.99}} \\
mnist:7 & 0.97 & \color{blue}{\textbf{0.98}} & 0.93 & 0.97 & \color{blue}{\textbf{0.98}} & 0.97 & 0.96 & 0.96 & 0.50 & 0.97 & 0.97 & 0.97 \\
mnist:8 & 0.85 & \color{blue}{\textbf{0.95}} & 0.80 & 0.80 & 0.80 & 0.89 & 0.86 & 0.78 & 0.50 & \color{blue}{\textbf{0.95}} & 0.87 & 0.88 \\
mnist:9 & 0.96 & \color{blue}{\textbf{0.98}} & 0.85 & 0.95 & 0.95 & 0.96 & 0.94 & 0.94 & 0.50 & 0.97 & 0.97 & 0.96 \\ \midrule
boot & 0.98 & \color{blue}{\textbf{0.99}} & 0.97 & 0.98 & 0.98 & 0.98 & 0.97 & 0.97 & 0.50 & 0.95 & 0.89 & 0.87 \\
bag & 0.89 & \color{blue}{\textbf{0.94}} & 0.82 & 0.87 & 0.86 & 0.90 & 0.90 & 0.84 & 0.50 & 0.75 & 0.86 & 0.84 \\
coat & 0.90 & \color{blue}{\textbf{0.93}} & 0.90 & 0.89 & 0.87 & 0.91 & 0.91 & 0.89 & 0.50 & 0.86 & 0.90 & 0.89 \\
dress & 0.92 & \color{blue}{\textbf{0.96}} & 0.91 & 0.92 & 0.90 & 0.94 & 0.94 & 0.93 & 0.50 & 0.90 & 0.93 & 0.92 \\
pullover & 0.89 & 0.91 & 0.86 & 0.89 & 0.87 & 0.90 & 0.89 & 0.88 & 0.50 & \color{blue}{\textbf{0.92}} & 0.88 & 0.87 \\
sandal & 0.90 & \color{blue}{\textbf{0.92}} & 0.89 & 0.89 & 0.90 & 0.90 & 0.90 & 0.91 & 0.50 & 0.74 & 0.83 & 0.84 \\
shirt & 0.82 & \color{blue}{\textbf{0.87}} & 0.78 & 0.81 & 0.78 & 0.85 & 0.84 & 0.80 & 0.50 & 0.78 & 0.84 & 0.82 \\
sneaker & 0.98 & \color{blue}{\textbf{0.99}} & 0.98 & 0.98 & 0.98 & 0.98 & 0.98 & 0.98 & 0.50 & 0.73 & 0.95 & 0.95 \\
t-shirt & 0.91 & \color{blue}{\textbf{0.94}} & 0.89 & 0.90 & 0.88 & 0.93 & 0.93 & 0.87 & 0.50 & 0.92 & 0.90 & 0.89 \\
trouser & \color{blue}{\textbf{0.99}} & \color{blue}{\textbf{0.99}} & \color{blue}{\textbf{0.99}} & \color{blue}{\textbf{0.99}} & \color{blue}{\textbf{0.99}} & \color{blue}{\textbf{0.99}} & \color{blue}{\textbf{0.99}} & \color{blue}{\textbf{0.99}} & 0.50 & \color{blue}{\textbf{0.99}} & 0.98 & \color{blue}{\textbf{0.99}} \\ \midrule
bright & \color{blue}{\textbf{1.00}} & 0.00 & 0.00 & \color{blue}{\textbf{1.00}} & \color{blue}{\textbf{1.00}} & \color{blue}{\textbf{1.00}} & \color{blue}{\textbf{1.00}} & \color{blue}{\textbf{1.00}} & 0.50 & 0.22 & 0.95 & 0.98 \\
cannye & \color{blue}{\textbf{1.00}} & 0.28 & 0.12 & \color{blue}{\textbf{1.00}} & \color{blue}{\textbf{1.00}} & 0.99 & 0.97 & \color{blue}{\textbf{1.00}} & 0.50 & 0.08 & \color{blue}{\textbf{1.00}} & \color{blue}{\textbf{1.00}} \\
dottedl & 0.99 & 0.51 & 0.28 & \color{blue}{\textbf{1.00}} & \color{blue}{\textbf{1.00}} & 0.80 & 0.82 & \color{blue}{\textbf{1.00}} & 0.50 & 0.59 & 0.88 & 0.86 \\
fog & \color{blue}{\textbf{1.00}} & 0.00 & 0.00 & \color{blue}{\textbf{1.00}} & \color{blue}{\textbf{1.00}} & 0.92 & 0.80 & \color{blue}{\textbf{1.00}} & 0.56 & 0.08 & 0.95 & 0.93 \\
glassb & \color{blue}{\textbf{0.97}} & 0.03 & 0.00 & \color{blue}{\textbf{0.97}} & \color{blue}{\textbf{0.97}} & 0.26 & 0.12 & 0.96 & 0.51 & 0.02 & 0.30 & 0.21 \\
impulsn & \color{blue}{\textbf{1.00}} & 0.58 & 0.00 & \color{blue}{\textbf{1.00}} & \color{blue}{\textbf{1.00}} & \color{blue}{\textbf{1.00}} & \color{blue}{\textbf{1.00}} & \color{blue}{\textbf{1.00}} & 0.52 & 0.12 & \color{blue}{\textbf{1.00}} & \color{blue}{\textbf{1.00}} \\
motionb & 0.37 & 0.19 & 0.00 & 0.71 & \color{blue}{\textbf{0.95}} & 0.29 & 0.14 & 0.50 & 0.54 & 0.36 & 0.41 & 0.33 \\
rotate & 0.24 & 0.35 & 0.12 & 0.51 & \color{blue}{\textbf{0.63}} & 0.50 & 0.45 & 0.43 & 0.50 & 0.44 & 0.49 & 0.48 \\
scale & 0.21 & 0.12 & \color{blue}{\textbf{0.65}} & 0.22 & 0.15 & 0.30 & 0.11 & 0.17 & 0.50 & 0.09 & 0.40 & 0.33 \\
shear & 0.57 & 0.40 & 0.08 & 0.83 & \color{blue}{\textbf{0.93}} & 0.58 & 0.54 & 0.54 & 0.50 & 0.47 & 0.55 & 0.54 \\
shotn & 0.99 & 0.34 & 0.21 & \color{blue}{\textbf{1.00}} & \color{blue}{\textbf{1.00}} & 0.64 & 0.62 & 0.94 & 0.59 & 0.07 & 0.78 & 0.77 \\
spatter & 0.96 & 0.23 & 0.01 & 0.96 & \color{blue}{\textbf{1.00}} & 0.57 & 0.56 & 0.91 & 0.97 & 0.40 & 0.62 & 0.59 \\
stripe & \color{blue}{\textbf{1.00}} & 0.33 & 0.95 & \color{blue}{\textbf{1.00}} & \color{blue}{\textbf{1.00}} & \color{blue}{\textbf{1.00}} & \color{blue}{\textbf{1.00}} & \color{blue}{\textbf{1.00}} & 0.53 & 0.74 & \color{blue}{\textbf{1.00}} & \color{blue}{\textbf{1.00}} \\
translt & \color{blue}{\textbf{0.98}} & 0.55 & 0.51 & 0.95 & 0.96 & 0.94 & 0.92 & 0.84 & 0.52 & 0.70 & 0.97 & 0.96 \\ \midrule
grid & 0.53 & 0.50 & 0.57 & \color{blue}{\textbf{0.60}} & 0.50 & 0.45 & 0.49 & 0.40 & 0.50 & 0.51 & 0.37 & 0.40 \\
transtr & 0.78 & 0.56 & 0.68 & \color{blue}{\textbf{0.81}} & 0.78 & 0.75 & 0.72 & 0.64 & 0.50 & 0.69 & 0.74 & 0.75 \\
wood & \color{blue}{\textbf{0.77}} & 0.57 & 0.52 & \color{blue}{\textbf{0.77}} & 0.76 & 0.75 & 0.74 & 0.59 & 0.50 & 0.54 & 0.64 & 0.66 \\\midrule
avg. AUC & 0.87  & 0.67  & 0.6  & \color{blue}{\textbf{0.89}}  & \color{blue}{\textbf{0.89}}  & 0.83  & 0.8  & 0.84  & 0.52  & 0.65  & 0.83  & 0.82 \\
avg. rank & \color{red}{\textbf{3.2}} & 5.2 & 9.1 & 3.6 & 4.1 & 3.8 & 5.7 & 6.0 & 10.1 & 7.5 & 4.7 & 5.2 \\\bottomrule
\end{tabular}

        \vspace*{0.15cm}
        \caption{no anomalies in validation}
    \end{subtable}
    \caption{Performance of models on the image data containing statistical anomalies with hyperparameter selection based on the availability of anomalies in validation data. Reported are the AUC values on the test dataset, averaged 5 random cross-validation repetitions where available.}
    \label{tab:images_stat_auc_auc_combined}
\end{table*}

\begin{table*}
    \centering
    \begin{subtable}{.48\linewidth}
        \footnotesize
        \tabcolsep=0.04cm
        \begin{tabular}{rrrrrrrrrrrrr}
\toprule
\textbf{dataset} & \textbf{aae} & \textbf{gano} & \textbf{skip} & \textbf{vae} & \textbf{wae} & \textbf{knn} & \textbf{osvm} & \textbf{fano} & \textbf{fmgn} & \textbf{dsvd} & \textbf{vaek} & \textbf{vaeo} \\\midrule
airpln & 0.69 & 0.65 & 0.76 & 0.68 & 0.68 & 0.70 & 0.72 & 0.77 & \color{blue}{\textbf{0.81}} & 0.75 & 0.68 & 0.68 \\
automb & 0.43 & 0.54 & 0.67 & 0.62 & 0.52 & 0.46 & 0.52 & 0.60 & \color{blue}{\textbf{0.83}} & 0.67 & 0.51 & 0.60 \\
bird & 0.69 & 0.68 & 0.68 & 0.68 & 0.69 & 0.70 & 0.70 & 0.69 & 0.69 & 0.67 & \color{blue}{\textbf{0.71}} & 0.70 \\
cat & 0.56 & 0.54 & \color{blue}{\textbf{0.66}} & 0.59 & 0.60 & 0.52 & 0.51 & 0.58 & 0.61 & 0.61 & 0.51 & 0.51 \\
deer & 0.76 & \color{blue}{\textbf{0.77}} & 0.75 & 0.75 & 0.73 & 0.76 & 0.76 & 0.75 & 0.60 & 0.74 & 0.76 & 0.76 \\
dog & 0.57 & 0.56 & \color{blue}{\textbf{0.67}} & 0.60 & 0.61 & 0.53 & 0.53 & 0.60 & 0.66 & 0.60 & 0.52 & 0.54 \\
frog & 0.74 & \color{blue}{\textbf{0.76}} & 0.74 & \color{blue}{\textbf{0.76}} & 0.70 & 0.73 & 0.73 & \color{blue}{\textbf{0.76}} & 0.70 & 0.72 & \color{blue}{\textbf{0.76}} & 0.75 \\
horse & 0.54 & 0.58 & 0.64 & 0.58 & 0.54 & 0.54 & 0.54 & 0.60 & \color{blue}{\textbf{0.76}} & 0.61 & 0.56 & 0.56 \\
ship & 0.72 & 0.67 & 0.78 & 0.72 & 0.72 & 0.71 & 0.70 & \color{blue}{\textbf{0.80}} & 0.76 & 0.74 & 0.70 & 0.69 \\
truck & 0.54 & 0.54 & 0.66 & 0.66 & 0.55 & 0.48 & 0.59 & 0.66 & \color{blue}{\textbf{0.76}} & 0.70 & 0.46 & 0.62 \\ \midrule
svhn2:0 & 0.65 & 0.61 & 0.54 & 0.65 & 0.65 & 0.61 & 0.60 & 0.66 & \color{blue}{\textbf{0.80}} & 0.65 & 0.64 & 0.62 \\
svhn2:1 & 0.68 & 0.61 & 0.57 & 0.68 & 0.68 & 0.63 & 0.61 & 0.62 & \color{blue}{\textbf{0.79}} & 0.63 & 0.67 & 0.64 \\
svhn2:2 & 0.62 & 0.58 & 0.53 & 0.62 & 0.64 & 0.59 & 0.58 & 0.58 & \color{blue}{\textbf{0.74}} & 0.61 & 0.62 & 0.59 \\
svhn2:3 & 0.59 & 0.56 & 0.52 & 0.59 & 0.58 & 0.57 & 0.56 & 0.57 & \color{blue}{\textbf{0.68}} & 0.58 & 0.59 & 0.58 \\
svhn2:4 & 0.63 & 0.59 & 0.52 & 0.65 & 0.66 & 0.61 & 0.59 & 0.62 & \color{blue}{\textbf{0.74}} & 0.58 & 0.64 & 0.61 \\
svhn2:5 & 0.60 & 0.57 & 0.52 & 0.60 & 0.61 & 0.57 & 0.57 & 0.59 & \color{blue}{\textbf{0.69}} & 0.57 & 0.60 & 0.59 \\
svhn2:6 & 0.59 & 0.56 & 0.53 & 0.59 & 0.59 & 0.58 & 0.56 & 0.60 & \color{blue}{\textbf{0.72}} & 0.57 & 0.60 & 0.56 \\
svhn2:7 & 0.65 & 0.60 & 0.53 & 0.66 & 0.66 & 0.61 & 0.60 & 0.61 & \color{blue}{\textbf{0.80}} & 0.61 & 0.64 & 0.62 \\
svhn2:8 & 0.60 & 0.58 & 0.52 & 0.60 & 0.60 & 0.56 & 0.55 & 0.60 & \color{blue}{\textbf{0.74}} & 0.57 & 0.59 & 0.58 \\
svhn2:9 & 0.60 & 0.57 & 0.51 & 0.60 & 0.60 & 0.57 & 0.57 & 0.59 & \color{blue}{\textbf{0.70}} & 0.58 & 0.60 & 0.58 \\\midrule
avg. AUC & 0.62 & 0.61 & 0.62 & 0.64 & 0.63 & 0.6 & 0.6 & 0.64 & \color{blue}{\textbf{0.73}} & 0.64 & 0.62 & 0.62 \\
avg. rank & 5.0 & 8.1 & 7.8 & 3.8 & 4.9 & 7.8 & 8.3 & 4.8 & \color{red}{\textbf{2.4}} & 6.2 & 5.6 & 6.6 \\\bottomrule
\end{tabular}

        \vspace*{0.15cm}
        \caption{ 50\% anomalies in validation}
    \end{subtable}
    \begin{subtable}{.48\linewidth}
        \footnotesize
        \tabcolsep=0.04cm
        \begin{tabular}{rrrrrrrrrrrrr}
\toprule
\textbf{dataset} & \textbf{aae} & \textbf{gano} & \textbf{skip} & \textbf{vae} & \textbf{wae} & \textbf{knn} & \textbf{osvm} & \textbf{fano} & \textbf{fmgn} & \textbf{dsvd} & \textbf{vaek} & \textbf{vaeo} \\\midrule
airpln & 0.67 & 0.52 & 0.53 & 0.68 & 0.68 & 0.66 & 0.68 & 0.67 & 0.50 & \color{blue}{\textbf{0.69}} & 0.64 & 0.65 \\
automb & 0.35 & 0.50 & 0.46 & 0.36 & 0.35 & 0.42 & 0.45 & 0.37 & 0.50 & \color{blue}{\textbf{0.67}} & 0.42 & 0.43 \\
bird & 0.69 & 0.62 & 0.65 & 0.68 & 0.68 & 0.69 & 0.69 & 0.68 & 0.50 & 0.58 & 0.69 & \color{blue}{\textbf{0.70}} \\
cat & 0.56 & 0.50 & \color{blue}{\textbf{0.60}} & 0.59 & \color{blue}{\textbf{0.60}} & 0.50 & 0.50 & 0.58 & 0.50 & 0.50 & 0.48 & 0.50 \\
deer & 0.72 & 0.73 & 0.70 & 0.70 & 0.69 & 0.75 & \color{blue}{\textbf{0.76}} & 0.74 & 0.50 & 0.72 & 0.74 & 0.75 \\
dog & 0.57 & 0.48 & 0.57 & 0.60 & \color{blue}{\textbf{0.61}} & 0.52 & 0.52 & 0.54 & 0.50 & 0.53 & 0.49 & 0.50 \\
frog & 0.62 & \color{blue}{\textbf{0.73}} & 0.72 & 0.56 & 0.52 & 0.72 & \color{blue}{\textbf{0.73}} & 0.64 & 0.50 & 0.50 & 0.72 & 0.72 \\
horse & 0.50 & 0.50 & 0.50 & 0.50 & 0.51 & 0.52 & \color{blue}{\textbf{0.53}} & 0.49 & 0.50 & \color{blue}{\textbf{0.53}} & 0.49 & 0.49 \\
ship & 0.69 & 0.57 & 0.61 & 0.71 & \color{blue}{\textbf{0.72}} & 0.69 & 0.70 & \color{blue}{\textbf{0.72}} & 0.50 & 0.68 & 0.66 & 0.67 \\
truck & 0.36 & 0.45 & 0.40 & 0.36 & 0.35 & 0.43 & 0.46 & 0.36 & 0.50 & \color{blue}{\textbf{0.64}} & 0.41 & 0.42 \\ \midrule
svhn2:0 & 0.59 & 0.61 & 0.50 & 0.56 & 0.56 & 0.58 & 0.57 & 0.55 & 0.50 & \color{blue}{\textbf{0.63}} & 0.60 & 0.59 \\
svhn2:1 & 0.62 & 0.60 & 0.54 & 0.59 & 0.58 & 0.58 & 0.57 & 0.62 & 0.50 & \color{blue}{\textbf{0.63}} & 0.62 & 0.61 \\
svhn2:2 & 0.56 & 0.57 & 0.51 & 0.53 & 0.53 & 0.55 & 0.55 & 0.53 & 0.50 & \color{blue}{\textbf{0.58}} & \color{blue}{\textbf{0.58}} & 0.57 \\
svhn2:3 & 0.52 & \color{blue}{\textbf{0.55}} & 0.49 & 0.50 & 0.50 & 0.54 & 0.54 & 0.52 & 0.50 & 0.54 & \color{blue}{\textbf{0.55}} & 0.54 \\
svhn2:4 & 0.57 & 0.57 & 0.52 & 0.54 & 0.54 & 0.57 & 0.56 & 0.54 & 0.50 & 0.58 & \color{blue}{\textbf{0.59}} & 0.58 \\
svhn2:5 & 0.53 & 0.49 & 0.49 & 0.51 & 0.50 & 0.55 & 0.55 & 0.52 & 0.50 & 0.55 & \color{blue}{\textbf{0.56}} & \color{blue}{\textbf{0.56}} \\
svhn2:6 & 0.52 & \color{blue}{\textbf{0.56}} & 0.49 & 0.50 & 0.50 & 0.55 & 0.54 & 0.49 & 0.50 & \color{blue}{\textbf{0.56}} & 0.52 & 0.51 \\
svhn2:7 & \color{blue}{\textbf{0.61}} & 0.59 & 0.51 & 0.57 & 0.56 & 0.57 & 0.56 & 0.56 & 0.50 & 0.60 & 0.60 & 0.59 \\
svhn2:8 & 0.52 & 0.54 & 0.47 & 0.49 & 0.49 & 0.53 & 0.53 & 0.53 & 0.50 & \color{blue}{\textbf{0.56}} & 0.55 & 0.54 \\
svhn2:9 & 0.53 & 0.56 & 0.48 & 0.51 & 0.51 & 0.54 & 0.53 & 0.54 & 0.50 & \color{blue}{\textbf{0.57}} & 0.56 & 0.55 \\\midrule
avg. AUC & 0.56 & 0.56 & 0.54 & 0.55 & 0.55 & 0.57 & 0.58 & 0.56 & 0.5 & \color{blue}{\textbf{0.59}} & 0.57 & 0.57 \\
avg. rank & 5.4 & 5.2 & 8.8 & 7.0 & 7.2 & 4.8 & 4.6 & 6.6 & 9.4 & \color{red}{\textbf{3.4}} & 4.7 & 4.8 \\\bottomrule
\end{tabular}

        \vspace*{0.15cm}
        \caption{no anomalies in validation}
    \end{subtable}
    \caption{Performance of models on the image data containing semantic anomalies with hyperparameter selection based on the availability of anomalies in validation data. Reported are the AUC values on the test dataset, averaged 5 random cross-validation repetitions where available.}
    \label{tab:images_semantic_auc_auc_combined}
\end{table*}

\section{Extending image datasets results}
\label{sec:appendix_extending_image_results}
In Sec.~\ref{sec:dataset_context} we stated that some models perform poorly on datasets with semantic anomalies, in which there are multiple sources of variation and the true anomalous information is pronounced in few of them. It is often the case that the methods focus more on the other sources of variation mainly the statistical aspects like background. As an entertaining example we have devised a simple anomaly detector \textit{blpix} for detecting the airplane class of the CIFAR10 dataset. As the name suggest it sums up all blue pixels in an image, resulting in anomaly score $s_{\text{blpix}}(x) = - \sum_{i,j} 1_{x_{i,j} = \text{blue}}$. Despite its simplicity, it achieves 0.68 AUC on both the validation and the test split, which is comparable with other models, see Fig.~\ref{fig:normal_loi}.

\begin{figure*}[h]
 \center 
 \input{images/airplane.tikz}	
 \caption{CIFAR10 images with the lowest anomaly scores (\emph{most normal}) for each model. Models was trained only on airplane images. Blpix is detector based on number of blue pixels.} 
 \label{fig:normal_loi}
\end{figure*}

\begin{figure*}
\centering
\begin{subfigure}[b]{\textwidth}
    \centering
   \resizebox{0.65\linewidth}{!}{\input{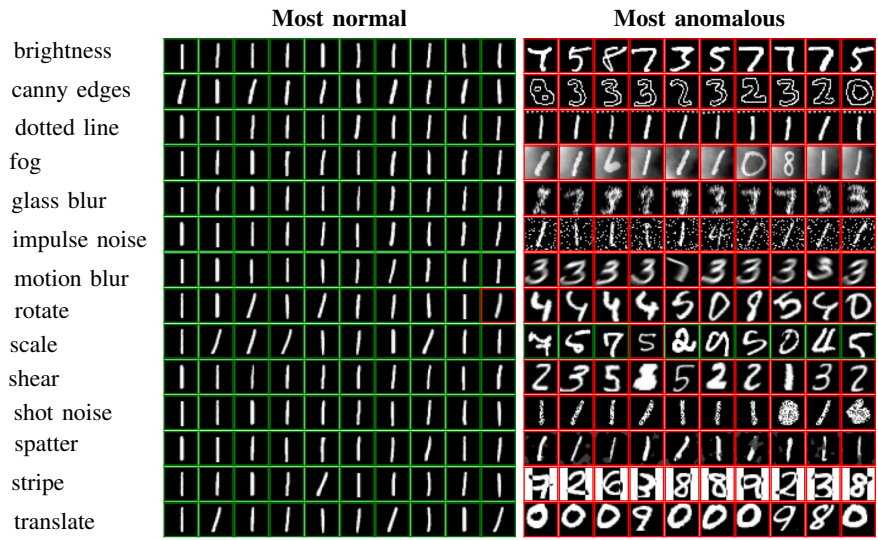}}
   \caption{WAE}
   \label{fig:wae-mnistc} 
\end{subfigure}

\begin{subfigure}[b]{\textwidth}
    \centering
   \resizebox{0.65\linewidth}{!}{\input{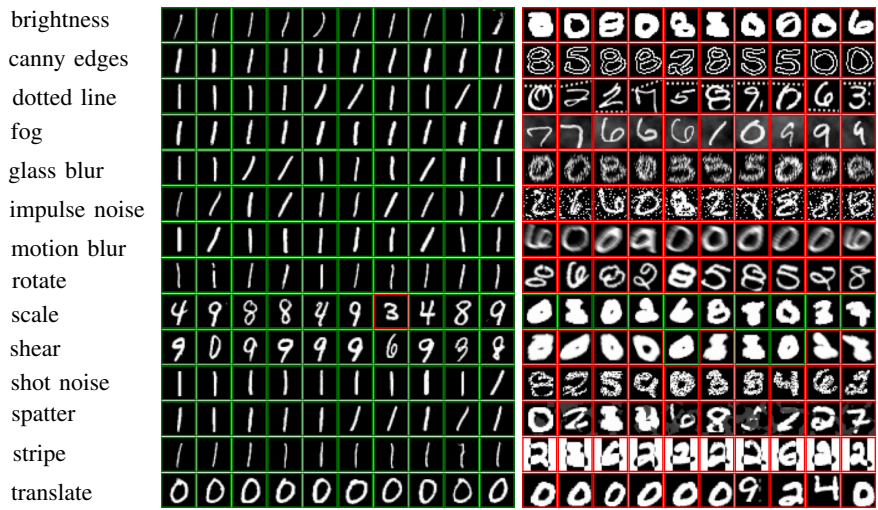}}
   \caption{fAnoGAN}
   \label{fig:fmgan-mnistc} 
\end{subfigure}

\begin{subfigure}[b]{\textwidth}
    \centering
   \resizebox{0.65\linewidth}{!}{\input{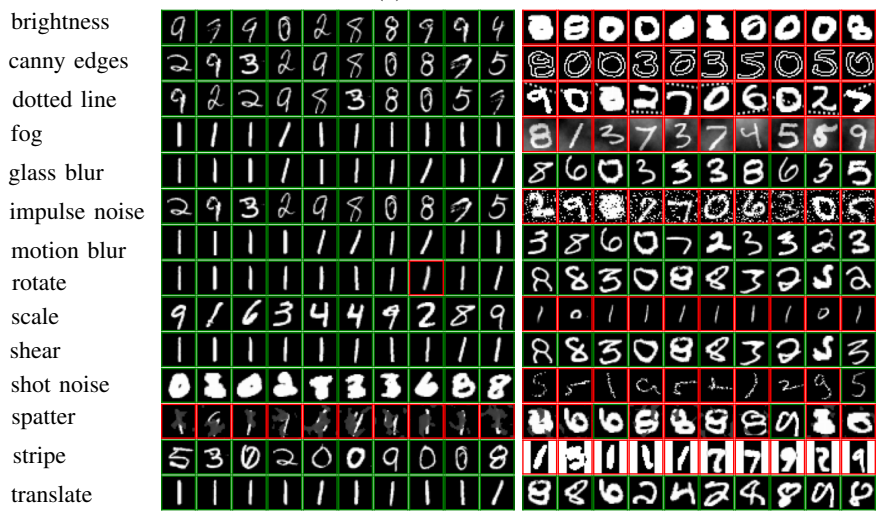}}
   \caption{OCSVM}
   \label{fig:ocsvm-mnistc} 
\end{subfigure}

\caption{Comparison of models on MNIST-C dataset. Left column consists of samples with the lowest anomaly score (most normal), right column with the highest score (most anomalous) and each row is different anomalous corruption. Color of frame around images corresponds to their true class (labels): \textit{green} means normal and \textit{red} means anomalous.}
\label{fig:mnistc_examples}
\end{figure*}

More visual examples of anomalies for the best performing models on MNIST, FashionMNIST, CIFAR10 and SVHN2 datasets are shown in Fig.~\ref{fig:fmgn_loi_examples}--\ref{fig:wae_loi_examples}. Similar examples are also shown for MVTec-AD and MNIST-C datasets in figures~\ref{fig:mvtec_examples} and~\ref{fig:mnistc_examples}

\begin{figure*}
\centering
\input{images/figs/fmgan-loi-colored.tikz}
\caption{Examples of fmGAN (fmgn), NC means normal class on which the models were trained. Color of frame around images corresponds to their true class (labels): \emph{green} means normal and \emph{red} means anomalous.}
\label{fig:fmgn_loi_examples}
\end{figure*}

\begin{figure*}
\centering
\input{images/figs/wae-loi-colored.tikz}
\caption{Examples of WAE with, NC means normal class on which the models were trained. Color of frame around images corresponds to their true class (labels): \emph{green} means normal and \emph{red} means anomalous.}
\label{fig:wae_loi_examples}
\end{figure*}

\begin{figure*}
\centering
\begin{subfigure}[b]{\textwidth}
   \input{images/figs/mvtec-wae.tikz}
   \caption{WAE}
   \label{fig:wae-mvtec} 
\end{subfigure}

\begin{subfigure}[b]{\textwidth}
   \input{images/figs/mvtec-fmgan-notag.tikz}
   \caption{fmGAN}
   \label{fig:fmgan-mvtec} 
\end{subfigure}

\begin{subfigure}[b]{\textwidth}
   \input{images/figs/mvtec-ocsvm-notag.tikz}
   \caption{OCSVM}
   \label{fig:ocsvm-mvtec}
\end{subfigure}

\caption{Comparison of models on MVTec-AD dataset. Left column consists of samples with the lowest anomaly score (most normal) and right column with the highest score (most anomalous). Color of frame around images corresponds to their true class (labels): \emph{green} means normal and \emph{red} means anomalous. }
\label{fig:mvtec_examples}
\end{figure*}

\section{Hyperparameter context}
Following the extension of image results, here we provide insights into the sensitivity of methods to the number of anomalies as seen from the point of view of different image datasets, which further illustrates the dataset context as an important factor in the model ranking. In Fig.~\ref{fig:image_knowledge_rank_pat_auc} we show the analogy of  Fig.~\ref{fig:combined_knowledge_rank_patn_auc_repre} for a more granular grouping of image datasets. The first thing that stands out that the performance on MVTec-AD dataset is constant while increasing the number of labeled samples from $0.01\%$ to $0.1\%$. This is due to the size of the dataset, which at these percentages does not contain enough samples to choose the best hyperparameters of each method and the choice is essentially arbitrary but consistent within this range. We admit that such points should not be even reported in the figures, but we kept it there for consistency with other datasets that have more samples. Although this behaviour is present also in the case of the tabular data, where the sample size are as low as $\sim80$ with Breast Tissue, the presence of large datasets such as the KDD99(10\%)  with $\sim500k$ samples allows us to see the general trend more clearly.

The second thing that we have already touched upon many times is the fact that with each dataset group the best performing models at both ends are different, with the exception of fmGAN, which does not work in the case of clean validation context (left-most points). 

Lastly we would like to shed some light on the use of the PR@\% metric in the context of the plots in question. Though throughout the text we have used AUC metric exclusively, the problem with using AUC on a portion of most anomalous samples is that we have a high chance of encountering samples of one class only, where the metric is not well defined. Therefore we opted for precision, where we assume that the threshold for discrete prediction is always lower than the last sample, as sorted by the anomaly score and thus all of them are labeled as positives.

\begin{figure*}[hbt!]
    \centering
    \resizebox {\linewidth}{!}{
    \begin{tikzpicture}[]
\begin{groupplot}[group style={vertical sep = 0.5cm, horizontal sep = 1.0cm, group size=4 by 1}]

\nextgroupplot [
  title = {(MNIST + FashionMNIST)},
  ylabel = {avg. AUC},
  legend style = {at={(0.3,1.30)}, anchor=west},
  width=5cm, height=7cm, scale only axis=true, 
  xtick={1,2,3,4,5,6,7,8}, 
  xticklabels={clean,$PR@\%0.01$,$PR@\%0.1$,$PR@\%1$,$PR@\%5$,$PR@\%10$,$PR@\%20$,$AUC_{val}$},
  width=5cm, height=7cm, scale only axis=true,
  x tick label style={rotate=50,anchor=east},
]

\addplot+ coordinates {
  (1.0, 0.9350000000000002)
  (2.0, 0.9410000000000001)
  (3.0, 0.9410000000000001)
  (4.0, 0.9410000000000001)
  (5.0, 0.9405000000000001)
  (6.0, 0.9399999999999998)
  (7.0, 0.9430000000000002)
  (8.0, 0.9545000000000001)
};

\addplot+ coordinates {
  (1.0, 0.9039999999999999)
  (2.0, 0.9019999999999999)
  (3.0, 0.8909999999999998)
  (4.0, 0.9034999999999999)
  (5.0, 0.9239999999999998)
  (6.0, 0.9295)
  (7.0, 0.938)
  (8.0, 0.968)
};

\addplot+ coordinates {
  (1.0, 0.9115000000000002)
  (2.0, 0.8880000000000002)
  (3.0, 0.8880000000000002)
  (4.0, 0.8985000000000003)
  (5.0, 0.9060000000000002)
  (6.0, 0.9215)
  (7.0, 0.9189999999999999)
  (8.0, 0.9555)
};

\addplot+ coordinates {
  (1.0, 0.5)
  (2.0, 0.3985)
  (3.0, 0.4534999999999999)
  (4.0, 0.5549999999999999)
  (5.0, 0.6385)
  (6.0, 0.651)
  (7.0, 0.7444999999999999)
  (8.0, 0.8605)
};

\addplot+ coordinates {
  (1.0, 0.9414999999999999)
  (2.0, 0.9414999999999999)
  (3.0, 0.9414999999999999)
  (4.0, 0.9414999999999999)
  (5.0, 0.9400000000000001)
  (6.0, 0.9400000000000001)
  (7.0, 0.9410000000000001)
  (8.0, 0.9484999999999999)
};

\addplot+ coordinates {
  (1.0, 0.9295000000000002)
  (2.0, 0.7180000000000002)
  (3.0, 0.7225000000000001)
  (4.0, 0.7939999999999999)
  (5.0, 0.8559999999999999)
  (6.0, 0.8995)
  (7.0, 0.9084999999999998)
  (8.0, 0.9435)
};

\addplot+ coordinates {
  (1.0, 0.922)
  (2.0, 0.796)
  (3.0, 0.8045000000000002)
  (4.0, 0.8420000000000002)
  (5.0, 0.8845000000000001)
  (6.0, 0.9275)
  (7.0, 0.9400000000000002)
  (8.0, 0.9535000000000003)
};

\addplot+ coordinates {
  (1.0, 0.9209999999999999)
  (2.0, 0.6945)
  (3.0, 0.6615)
  (4.0, 0.716)
  (5.0, 0.8455)
  (6.0, 0.891)
  (7.0, 0.9199999999999999)
  (8.0, 0.9450000000000003)
};

\addplot+ coordinates {
  (1.0, 0.9180000000000001)
  (2.0, 0.9414999999999998)
  (3.0, 0.9414999999999998)
  (4.0, 0.9404999999999999)
  (5.0, 0.9399999999999998)
  (6.0, 0.942)
  (7.0, 0.9434999999999999)
  (8.0, 0.9540000000000003)
};


\nextgroupplot [
  title = {(SVHN2 + CIFAR10)},
  legend style = {at={(1.1,1.20)}, anchor=center},
  legend columns = -1
  width=5cm, height=7cm, scale only axis=true, 
  xtick={1,2,3,4,5,6,7,8}, 
  xticklabels={clean,$PR@\%0.01$,$PR@\%0.1$,$PR@\%1$,$PR@\%5$,$PR@\%10$,$PR@\%20$,$AUC_{val}$},
  width=5cm, height=7cm, scale only axis=true,
  x tick label style={rotate=50,anchor=east},
]

\addplot+ coordinates {
  (1.0, 0.5649999999999998)
  (2.0, 0.584)
  (3.0, 0.5745000000000001)
  (4.0, 0.5934999999999999)
  (5.0, 0.6)
  (6.0, 0.6079999999999999)
  (7.0, 0.614)
  (8.0, 0.6224999999999999)
};

\addplot+ coordinates {
  (1.0, 0.5920000000000001)
  (2.0, 0.575)
  (3.0, 0.5685)
  (4.0, 0.5975000000000001)
  (5.0, 0.6250000000000001)
  (6.0, 0.6320000000000001)
  (7.0, 0.6325000000000001)
  (8.0, 0.638)
};

\addplot+ coordinates {
  (1.0, 0.5595000000000001)
  (2.0, 0.541)
  (3.0, 0.5544999999999999)
  (4.0, 0.5899999999999999)
  (5.0, 0.5955)
  (6.0, 0.61)
  (7.0, 0.633)
  (8.0, 0.6424999999999998)
};

\addplot+ coordinates {
  (1.0, 0.5)
  (2.0, 0.4885)
  (3.0, 0.49799999999999994)
  (4.0, 0.6194999999999999)
  (5.0, 0.6765)
  (6.0, 0.7)
  (7.0, 0.7249999999999999)
  (8.0, 0.729)
};

\addplot+ coordinates {
  (1.0, 0.5730000000000001)
  (2.0, 0.5834999999999999)
  (3.0, 0.5824999999999999)
  (4.0, 0.5890000000000001)
  (5.0, 0.5945)
  (6.0, 0.5985)
  (7.0, 0.599)
  (8.0, 0.6015)
};

\addplot+ coordinates {
  (1.0, 0.5760000000000001)
  (2.0, 0.5090000000000001)
  (3.0, 0.536)
  (4.0, 0.5625)
  (5.0, 0.5860000000000001)
  (6.0, 0.5925)
  (7.0, 0.5975000000000001)
  (8.0, 0.6045)
};

\addplot+ coordinates {
  (1.0, 0.552)
  (2.0, 0.5469999999999999)
  (3.0, 0.5625)
  (4.0, 0.5995000000000001)
  (5.0, 0.6210000000000002)
  (6.0, 0.6315000000000001)
  (7.0, 0.6315)
  (8.0, 0.6439999999999999)
};

\addplot+ coordinates {
  (1.0, 0.5735000000000001)
  (2.0, 0.558)
  (3.0, 0.5574999999999999)
  (4.0, 0.574)
  (5.0, 0.588)
  (6.0, 0.5994999999999999)
  (7.0, 0.6035000000000001)
  (8.0, 0.619)
};

\addplot+ coordinates {
  (1.0, 0.549)
  (2.0, 0.5820000000000002)
  (3.0, 0.5705)
  (4.0, 0.6065)
  (5.0, 0.614)
  (6.0, 0.615)
  (7.0, 0.625)
  (8.0, 0.6305)
};

\legend{{}{aae}, {}{dsvd}, {}{fano}, {}{fmgn}, {}{knn}, {}{osvm}, {}{vae}, {}{vaeo}, {}{wae}}

\nextgroupplot [
  title = {(MNIST-C)},
  legend style = {at={(0.3,1.30)}, anchor=west},
  width=5cm, height=7cm, scale only axis=true, 
  xtick={1,2,3,4,5,6,7,8}, 
  xticklabels={clean,$PR@\%0.01$,$PR@\%0.1$,$PR@\%1$,$PR@\%5$,$PR@\%10$,$PR@\%20$,$AUC_{val}$},
  width=5cm, height=7cm, scale only axis=true,
  x tick label style={rotate=50,anchor=east},
]

\addplot+ coordinates {
  (1.0, 0.8057142857142858)
  (2.0, 0.8049999999999999)
  (3.0, 0.8350000000000001)
  (4.0, 0.89)
  (5.0, 0.8857142857142859)
  (6.0, 0.8857142857142859)
  (7.0, 0.8892857142857142)
  (8.0, 0.8914285714285716)
};

\addplot+ coordinates {
  (1.0, 0.31285714285714283)
  (2.0, 0.6821428571428573)
  (3.0, 0.7478571428571429)
  (4.0, 0.7878571428571429)
  (5.0, 0.7821428571428571)
  (6.0, 0.8049999999999999)
  (7.0, 0.8071428571428572)
  (8.0, 0.8107142857142857)
};

\addplot+ coordinates {
  (1.0, 0.8064285714285714)
  (2.0, 0.8407142857142856)
  (3.0, 0.8571428571428571)
  (4.0, 0.8785714285714287)
  (5.0, 0.9021428571428572)
  (6.0, 0.9078571428571429)
  (7.0, 0.9264285714285715)
  (8.0, 0.9314285714285715)
};

\addplot+ coordinates {
  (1.0, 0.5528571428571428)
  (2.0, 0.7592857142857142)
  (3.0, 0.8135714285714286)
  (4.0, 0.8871428571428572)
  (5.0, 0.8714285714285716)
  (6.0, 0.8857142857142858)
  (7.0, 0.925)
  (8.0, 0.9578571428571429)
};

\addplot+ coordinates {
  (1.0, 0.6992857142857142)
  (2.0, 0.8035714285714287)
  (3.0, 0.8135714285714285)
  (4.0, 0.8328571428571429)
  (5.0, 0.8350000000000001)
  (6.0, 0.8357142857142857)
  (7.0, 0.8378571428571429)
  (8.0, 0.8421428571428571)
};

\addplot+ coordinates {
  (1.0, 0.6464285714285715)
  (2.0, 0.7435714285714287)
  (3.0, 0.7400000000000001)
  (4.0, 0.7485714285714286)
  (5.0, 0.7464285714285716)
  (6.0, 0.7714285714285714)
  (7.0, 0.7814285714285714)
  (8.0, 0.8392857142857144)
};

\addplot+ coordinates {
  (1.0, 0.8678571428571428)
  (2.0, 0.8328571428571429)
  (3.0, 0.8664285714285713)
  (4.0, 0.8557142857142858)
  (5.0, 0.8935714285714287)
  (6.0, 0.8935714285714287)
  (7.0, 0.8950000000000001)
  (8.0, 0.915)
};

\addplot+ coordinates {
  (1.0, 0.7128571428571429)
  (2.0, 0.7014285714285714)
  (3.0, 0.7742857142857142)
  (4.0, 0.8314285714285712)
  (5.0, 0.9285714285714285)
  (6.0, 0.9385714285714285)
  (7.0, 0.9457142857142855)
  (8.0, 0.9635714285714286)
};

\addplot+ coordinates {
  (1.0, 0.8992857142857142)
  (2.0, 0.8564285714285714)
  (3.0, 0.8892857142857143)
  (4.0, 0.8935714285714287)
  (5.0, 0.9321428571428572)
  (6.0, 0.9335714285714286)
  (7.0, 0.9364285714285715)
  (8.0, 0.937857142857143)
};


\nextgroupplot [
  title = {(MVTec-AD)},
  legend style = {at={(0.3,1.30)}, anchor=west},
  width=5cm, height=7cm, scale only axis=true, 
  xtick={1,2,3,4,5,6,7,8}, 
  xticklabels={clean,$PR@\%0.01$,$PR@\%0.1$,$PR@\%1$,$PR@\%5$,$PR@\%10$,$PR@\%20$,$AUC_{val}$},
  width=5cm, height=7cm, scale only axis=true,
  x tick label style={rotate=50,anchor=east},
]

\addplot+ coordinates {
  (1.0, 0.6933333333333334)
  (2.0, 0.6733333333333333)
  (3.0, 0.6733333333333333)
  (4.0, 0.6733333333333333)
  (5.0, 0.6733333333333333)
  (6.0, 0.6733333333333333)
  (7.0, 0.6866666666666666)
  (8.0, 0.6866666666666666)
};

\addplot+ coordinates {
  (1.0, 0.58)
  (2.0, 0.57)
  (3.0, 0.57)
  (4.0, 0.57)
  (5.0, 0.5733333333333334)
  (6.0, 0.5833333333333334)
  (7.0, 0.5833333333333334)
  (8.0, 0.62)
};

\addplot+ coordinates {
  (1.0, 0.5433333333333333)
  (2.0, 0.6733333333333333)
  (3.0, 0.6733333333333333)
  (4.0, 0.6733333333333333)
  (5.0, 0.7166666666666667)
  (6.0, 0.6699999999999999)
  (7.0, 0.7066666666666667)
  (8.0, 0.7233333333333333)
};

\addplot+ coordinates {
  (1.0, 0.5)
  (2.0, 0.5166666666666667)
  (3.0, 0.5166666666666667)
  (4.0, 0.5166666666666667)
  (5.0, 0.68)
  (6.0, 0.7200000000000001)
  (7.0, 0.68)
  (8.0, 0.7166666666666667)
};

\addplot+ coordinates {
  (1.0, 0.65)
  (2.0, 0.6666666666666666)
  (3.0, 0.6666666666666666)
  (4.0, 0.6666666666666666)
  (5.0, 0.6666666666666666)
  (6.0, 0.6666666666666666)
  (7.0, 0.6699999999999999)
  (8.0, 0.6699999999999999)
};

\addplot+ coordinates {
  (1.0, 0.65)
  (2.0, 0.4033333333333333)
  (3.0, 0.4033333333333333)
  (4.0, 0.4033333333333333)
  (5.0, 0.6566666666666666)
  (6.0, 0.65)
  (7.0, 0.65)
  (8.0, 0.6766666666666667)
};

\addplot+ coordinates {
  (1.0, 0.7266666666666667)
  (2.0, 0.6166666666666667)
  (3.0, 0.6166666666666667)
  (4.0, 0.6166666666666667)
  (5.0, 0.6966666666666667)
  (6.0, 0.71)
  (7.0, 0.73)
  (8.0, 0.7266666666666667)
};

\addplot+ coordinates {
  (1.0, 0.6033333333333334)
  (2.0, 0.57)
  (3.0, 0.57)
  (4.0, 0.57)
  (5.0, 0.5266666666666667)
  (6.0, 0.5266666666666667)
  (7.0, 0.5700000000000001)
  (8.0, 0.6633333333333332)
};

\addplot+ coordinates {
  (1.0, 0.68)
  (2.0, 0.6566666666666666)
  (3.0, 0.6566666666666666)
  (4.0, 0.6566666666666666)
  (5.0, 0.71)
  (6.0, 0.7166666666666668)
  (7.0, 0.7166666666666668)
  (8.0, 0.7200000000000001)
};

\end{groupplot}

\end{tikzpicture}
    }
    \caption{Sensitivity of methods to the number of anomalies available in the validation set for hyperparameter selection visualized in terms of the achieved AUC aggregated over all datasets in each category (columns). The clean validation context is the left-most point on the x-axis, and the anomaly validation context (50\% of available anomalies) is the right-most point. The points in-between were obtained by selecting models with highest precision on the reported portion (e.g. 5\%) of validation samples with the highest anomaly scores.}
    \label{fig:image_knowledge_rank_pat_auc}
\end{figure*}

\textit{Sensitivity study of OC-SVM:} In order to better support the thesis about insufficient tuning of the OCSVM method, we provide here the average ranks of all methods trained on tabular data, where the fully optimized method was replaced by it's OCSVM-RBF counterpart (only the width of the rbf kernel was sampled and hyperparameter $\nu$ was set to $0.5$). As average ranks are largely dependent also on the composition of the methods this change has a profound effect on ranks of other methods such as VAE, which takes the lead, see Tab.~\ref{tab:tabular_auc_auc_meanmax_orbf_ranks_only}.

\begin{table*}
    \footnotesize
    \centering
    \tabcolsep=0.05cm
    \begin{tabular}{rrrrrrrrrrrrrrrrrrrrrrrrr}
\toprule
& \textbf{aae} & \textbf{avae} & \textbf{gano} & \textbf{vae} & \textbf{wae} & \textbf{abod} & \textbf{hbos} & \textbf{if} & \textbf{knn} & \textbf{loda} & \textbf{lof} & \textbf{orbf} & \textbf{pidf} & \textbf{maf} & \textbf{rnvp} & \textbf{sptn} & \textbf{fmgn} & \textbf{gan} & \textbf{mgal} & \textbf{dagm} & \textbf{dsvd} & \textbf{rpn} & \textbf{vaek} & \textbf{vaeo} \\\midrule
avg. rank & 6.6 & 13.1 & 10.0 & \color{red}{\textbf{6.2}} & 6.7 & 11.2 & 14.6 & 14.3 & 8.0 & 15.8 & 11.2 & 8.1 & 14.2 & 8.6 & 8.6 & 10.4 & 11.1 & 11.7 & 22.3 & 19.6 & 16.1 & 11.8 & 10.8 & 6.4 \\\bottomrule
\end{tabular}

    \vspace*{0.15cm}
    \caption{Average AUC model ranks on tabular data, where \textbf{osvm} from Tab.~\ref{tab:tabular_anomalies} has been replaced with \textbf{orbf} - optimizing the width of rbf kernel and using $\nu = 0.5$.}
    \label{tab:tabular_auc_auc_meanmax_orbf_ranks_only}
\end{table*}

\section{Other influences}
\subsection{Performance metrics}
\label{sec:appendix_metrics}
Throughout the main text, we have reported only results for the AUC metric. Though this has had quite practical implications on the degrees of freedom, we are aware that the choice of metric may favour each method differently. To this extend we provide Tab.~\ref{tab:metric_comparison_grouped} containing simple comparison of ranks as measured in AUC and TPR@5\% of FPR on each of the dataset types. The changes in ranks are more pronounced in the case of tabular datasets, where the biggest improvement when going from AUC to TPR@5\% is seen with the worst performing model, MOGAAL, thus swapping its place with DAGMM. The biggest fall has been experienced by the WAE, VAE method but not as much to affect the overall ranking at the top positions. On image datasets the changes are really minor.

\begin{table*}
    \begin{subtable}{\linewidth}
        \centering
        \footnotesize
        \tabcolsep=0.05cm
        \begin{tabular}{rrrrrrrrrrrrrrrrrrrrrrrrr}
\toprule
\textbf{metric} & \textbf{aae} & \textbf{avae} & \textbf{gano} & \textbf{vae} & \textbf{wae} & \textbf{abod} & \textbf{hbos} & \textbf{if} & \textbf{knn} & \textbf{loda} & \textbf{lof} & \textbf{osvm} & \textbf{pidf} & \textbf{maf} & \textbf{rnvp} & \textbf{sptn} & \textbf{fmgn} & \textbf{gan} & \textbf{mgal} & \textbf{dagm} & \textbf{dsvd} & \textbf{rpn} & \textbf{vaek} & \textbf{vaeo} \\\midrule
AUC & 6.9 & 13.4 & 10.3 & 6.5 & 7.0 & 11.4 & 14.9 & 14.5 & 8.4 & 16.1 & 11.4 & \color{red}{\textbf{2.9}} & 14.3 & 8.9 & 9.0 & 10.7 & 11.4 & 12.0 & 22.3 & 19.8 & 16.2 & 12.1 & 11.0 & 6.8 \\
TPR@5 & 7.0 & 12.6 & 10.1 & 8.4 & 9.6 & 12.3 & 15.1 & 16.2 & 10.2 & 18.0 & 11.1 & \color{red}{\textbf{3.2}} & 14.1 & 10.1 & 10.2 & 12.2 & 9.7 & 10.1 & 19.2 & 20.6 & 16.1 & 12.0 & 11.5 & 6.6 \\
rank. change & 0.1 & -0.8 & -0.2 & 1.9 & \color{blue}{\textbf{2.6}} & 0.9 & 0.2 & 1.7 & 1.8 & 1.9 & -0.3 & 0.3 & -0.2 & 1.2 & 1.2 & 1.5 & -1.7 & -1.9 & \color{red}{\textbf{-3.1}} & 0.8 & -0.1 & -0.1 & 0.5 & -0.2 \\\bottomrule
\end{tabular}

        \vspace*{0.05cm}
        \caption{tabular}
    \end{subtable}
    \begin{subtable}{\linewidth}
        \centering
        \footnotesize
        \vspace*{0.05cm}
        \tabcolsep=0.05cm
        \begin{tabular}{rrrrrrrrrrrrr}
\toprule
\textbf{metric} & \textbf{aae} & \textbf{gano} & \textbf{skip} & \textbf{vae} & \textbf{wae} & \textbf{knn} & \textbf{osvm} & \textbf{fano} & \textbf{fmgn} & \textbf{dsvd} & \textbf{vaek} & \textbf{vaeo} \\\midrule
AUC & 3.6 & 5.1 & 9.1 & 3.1 & \color{red}{\textbf{2.9}} & 5.4 & 6.4 & 3.4 & 7.5 & 4.0 & 5.7 & 4.9 \\
TPR@5 & 3.9 & 5.2 & 9.3 & \color{red}{\textbf{3.8}} & \color{red}{\textbf{3.8}} & 6.7 & 6.7 & 4.7 & 7.5 & 4.2 & 6.7 & 6.1 \\
rank. change & 0.3 & 0.1 & 0.2 & 0.7 & 0.9 & 1.3 & 0.3 & \color{blue}{\textbf{1.3}} & \color{red}{\textbf{0.0}} & 0.2 & 1.0 & 1.2 \\\bottomrule
\end{tabular}

        \vspace*{0.05cm}
        \caption{statistic}
    \end{subtable}
    \begin{subtable}{\linewidth}
        \centering
        \footnotesize
        \tabcolsep=0.05cm
        \vspace*{0.05cm}
        \begin{tabular}{rrrrrrrrrrrrr}
\toprule
\textbf{metric} & \textbf{aae} & \textbf{gano} & \textbf{skip} & \textbf{vae} & \textbf{wae} & \textbf{knn} & \textbf{osvm} & \textbf{fano} & \textbf{fmgn} & \textbf{dsvd} & \textbf{vaek} & \textbf{vaeo} \\\midrule
AUC & 5.0 & 8.1 & 7.8 & 3.8 & 4.9 & 7.8 & 8.3 & 4.8 & \color{red}{\textbf{2.4}} & 6.2 & 5.6 & 6.6 \\
TPR@5 & 5.8 & 8.2 & 7.0 & 4.8 & 4.1 & 7.8 & 8.8 & 4.2 & \color{red}{\textbf{2.6}} & 5.4 & 6.6 & 6.3 \\
rank. change & 0.8 & 0.1 & -0.8 & \color{blue}{\textbf{1.0}} & \color{red}{\textbf{-0.8}} & 0.0 & 0.5 & -0.6 & 0.2 & -0.8 & \color{blue}{\textbf{1.0}} & -0.3 \\\bottomrule
\end{tabular}

        \vspace*{0.05cm}
        \caption{semantic}
    \end{subtable}
    \caption{Comparison of average ranks on different dataset types in the anomaly validation context for two selection criteria: AUC or TPR@5. The results of hyperparameters chosen according to each metric have been averaged over 5 random cross-validation repetitions, where available and reported ranks are averaged over all datasets in the respective categories. The last row shows the rank difference between both metrics.}
    \label{tab:metric_comparison_grouped}
\end{table*}

\subsection{Bayesian optimization}
\label{sec:appendix_bayes}
Bayesian optimization allowed us to fill the gaps in the discrete hyperparameter sampling ranges described in tables~\ref{tab:classical_hyperparameters}--\ref{tab:flow_hyperparameters}, though we have not always included the range as a whole, but for example in case of the neural network architectures we used powers of 2 and optimized the exponent. As we have been building on top of already trained models, we have not found an easy way how to optimize hyperparameters such as weight decay, where we used to indicate $0.0$ as not using this regularization, while still using log-uniform with base 10 for sampling, therefore we have chosen to keep the regularization while enlarging the parameter space to smaller values. In spite of this the definiton of sampling ranges was generally a simple exercise and the biggest hurdle was on the implementation side of things, which we will discuss briefly in the implementation details section~\ref{sec:appendix_bayes_impl}.

\begin{table}[]
    \centering
    \begin{tabular}{cc}
        \toprule
         \textbf{parameter} & value \\ \midrule
         estimator & GP with Matern kernel \\
         acq. function & automatic choice \\
         acq. function optimizer & auto \\
         no. restarts of optimizer & 5 \\
         acq. function no. points & 10000 \\
         noise & gaussian \\
         $\xi$ & 0.01 \\
         $\kappa$ & 1.96 \\ \bottomrule
    \end{tabular}
    \caption{Default parameters of bayesian optimization from scikit-optimise~\citep{scikit-learn} framework used for all models and datasets.}
    \label{tab:params_bayes}
\end{table}

Regarding the hyperparameters for the optimization itself, we used the default parameters as provided by the scikit-optimise framework, see Tab.~\ref{tab:params_bayes}, which is certainly orthogonal to one of the conclustions of the main text, however including another degree of freedom to this large scale comparison was computationally unrealistic. Due to the same constraints, we have limited the Bayesian optimization to tabular data only, as the biggest time constraint is given by the sequential nature of training, whereas random sampling allows to fit multiple hyperparameters on the same dataset in parallel. kNN, ABOD, LOF methods have been purposely left out of this experiment as their hyperparameter space is one dimensional and thus extensively sampled already. In the case of the MOGAAL methods the decision has been made based on the nature of available hyperparameters, which affected only the training part but not the model's architecture, therefore we did not expect much improvement there.

As has been already mentioned we have trained initially 50 random samples (with already precomputed methods we have just selected the random search runs) and then used the Bayesian optimization to propose the next points, while updating the underlying process. At the end we have been left with 100 samples, which we have evaluated in the same way as in the case of random sampling, i.e. choosing the best performing models based on AUC on validation data averaged over 5 repetitions, which also served as the objective function optimized for the Bayesian optimization. Comparison of ranks between random sampling and Bayesian optimization is provided in Tab.~\ref{tab:tabular_bayes_comparison}. We can see that generally all methods improved in terms of average change in AUC, with the exception of adVAE, whose performance degraded on $21$ out of $40$ datasets. On the other hand the methods that did improve significantly was GANomaly, which jumped by 3 ranks on average, surpassing AAE and flow models just to name a few. Other notable improvement has been made by GAN and RealNVP flow. 

Some of the drops in performance may be explained by our choice of default parameters, as for some model--dataset combination the optimization did not explore as much of the hyperparameter space or even got stuck at one point. As a result, the pool of methods may have shrunk close to the original 50 random samples. One other factor that may have affected the results, has been the lack of hyperparameters restrictions that we have manually imposed after random sample has been drawn, such having latent dimension of autoencoders lower than the input dimension. 

When put together, the individual gains we have seen (namely GANomaly, flows and gans) are definitely worth exploring, when one has only one objective to aim at and enough computational budged to explore even the hyperparameters of Bayesian optimization itself.

\begin{table*}
    \footnotesize
    \centering
    \tabcolsep=0.05cm
    \begin{tabular}{rrrrrrrrrrrrrrrrrrrrrrrrr}
\toprule
 & \textbf{aae} & \textbf{avae} & \textbf{gano} & \textbf{vae} & \textbf{wae} & \color{magenta}{\textbf{abod}} & \textbf{hbos} & \textbf{if} & \color{magenta}{\textbf{knn}} & \textbf{loda} & \color{magenta}{\textbf{lof}} & \textbf{osvm} & \textbf{pidf} & \textbf{maf} & \textbf{rnvp} & \textbf{sptn} & \textbf{fmgn} & \textbf{gan} & \color{magenta}{\textbf{mgal}} & \textbf{dagm} & \textbf{dsvd} & \textbf{rpn} & \textbf{vaek} & \textbf{vaeo} \\\midrule
random sampl. & 6.9 & 13.4 & 10.3 & 6.5 & 7.0 & 11.4 & 14.9 & 14.5 & 8.4 & 16.1 & 11.4 & \color{red}{\textbf{2.9}} & 14.3 & 8.9 & 9.0 & 10.7 & 11.4 & 12.0 & 22.3 & 19.8 & 16.2 & 12.1 & 11.0 & 6.8 \\
Bayesian opt. & 7.7 & 15.6 & 7.0 & 5.6 & 6.2 & 12.3 & 14.9 & 15.3 & 9.4 & 16.5 & 12.3 & \color{red}{\textbf{2.8}} & 14.8 & 7.9 & 6.9 & 11.1 & 11.0 & 9.6 & 22.4 & 19.5 & 16.2 & 11.6 & 11.8 & 7.3 \\
rank change & 0.8 & 2.2 & \color{red}{\textbf{-3.3}} & -0.9 & -0.8 & 0.9 & 0.0 & 0.8 & 1.0 & 0.4 & 0.9 & -0.1 & 0.5 & -1.0 & -2.1 & 0.4 & -0.4 & -2.4 & 0.1 & -0.3 & 0.0 & -0.5 & 0.8 & 0.5 \\
avg. change & -0.00 & -0.02 & 0.03 & 0.01 & 0.00 & 0.00 & 0.01 & -0.00 & 0.00 & 0.01 & 0.00 & 0.01 & 0.00 & 0.01 & 0.02 & 0.00 & 0.01 & 0.02 & 0.00 & \color{blue}{\textbf{0.04}} & 0.03 & 0.01 & 0.00 & -0.00 \\
no. improved & 9 & 14 & \color{blue}{\textbf{30}} & 15 & 14 & 0 & 15 & 3 & 0 & 15 & 0 & 8 & 7 & 16 & 25 & 16 & 23 & 27 & 0 & 18 & 26 & 17 & 4 & 3 \\
no. degraded & 13 & \color{blue}{\textbf{21}} & 3 & 5 & 6 & 0 & 7 & 10 & 0 & 7 & 0 & 4 & 5 & 4 & 2 & 3 & 7 & 2 & 0 & 9 & 6 & 5 & 0 & 2 \\
\bottomrule
\end{tabular}

    \vspace*{0.15cm}
    \caption{Comparison of average ranks in the AUC metric between Bayesian sampling and random search sampling. Methods that were not optimized with Bayesian are marked with magenta color. The second to last and last row shows the number of datasets on which the performance improved and degraded respectively.}
    \label{tab:tabular_bayes_comparison}
\end{table*}


\subsection{Ensembles experiments}
\label{sec:appendix_ensembles}
Creation of ensembles/hyperparameter averaging method consisted of two steps. Firstly the performance metrics were precomputed for all the methods and each repetition or anomaly class, in order to quickly assert which hyperparameters to include. Secondly we created multiple ensembles for each combination of the criterion metric and size, totaling in $4$ different ensembles per each repetition, which allowed us to compare individual effects. Resulting average ranks in the AUC metric and average improvement in the AUC are presented in odd/even rows respectively of tables~\ref{tab:ensembles_sensitivity_grouped}.a--c for each dataset type and combination of criterion and ensemble size. We can see that in the majority the effect on AUC is zero, with notable exception of GANomaly on tabular data and feature matching GAN on images containing semantic anomalies. Notable observation is that smaller ensembles perform usually better, which may suggest that by not weighting individual members of the ensembles or by not employing some heuristic for choosing the optimal size, we are crudely including hyperparameters, that only hinder the performance. 

\begin{table*}
    \begin{subtable}{\linewidth}
        \centering
        \footnotesize
        \tabcolsep=0.025cm
        \begin{tabular}{llrrrrrrrrrrrrrrrrrrrrrrrr}
\toprule
\textbf{criterion} & \textbf{size} & \textbf{aae} & \textbf{avae} & \textbf{gano} & \textbf{vae} & \textbf{wae} & \textbf{abod} & \textbf{hbos} & \textbf{if} & \textbf{knn} & \textbf{loda} & \textbf{lof} & \textbf{osvm} & \textbf{pidf} & \textbf{maf} & \textbf{rnvp} & \textbf{sptn} & \textbf{fmgn} & \textbf{gan} & \textbf{mgal} & \textbf{dagm} & \textbf{dsvd} & \textbf{rpn} & \textbf{vaek} & \textbf{vaeo} \\\midrule

$\text{AUC}$ & 5 & 7.8 & 17.8 & 7.2 & 6.5 & 6.4 & 13.0 & 16.4 & 14.5 & 9.8 & 15.6 & 11.8 & \color{red}{\textbf{4.0}} & 14.0 & 7.6 & 7.8 & 11.3 & 10.2 & 9.7 & 17.2 & 19.8 & 17.8 & 12.6 & 11.0 & 6.4 \\
&  & 0.01 & -0.04 & \color{blue}{\textbf{0.02}} & 0.00 & 0.00 & -0.00 & -0.01 & -0.00 & -0.00 & -0.00 & 0.00 & 0.00 & 0.00 & 0.00 & 0.00 & -0.00 & 0.01 & 0.01 & 0.01 & -0.00 & -0.02 & -0.01 & 0.00 & -0.00 \\
&  10 & 7.3 & 19.8 & 7.5 & 6.3 & 5.9 & 12.6 & 16.1 & 14.2 & 10.2 & 15.3 & 11.8 & \color{red}{\textbf{4.1}} & 13.9 & 7.6 & 7.9 & 11.2 & 9.7 & 9.4 & 17.9 & 19.8 & 18.5 & 12.5 & 10.8 & 6.9 \\
&  & 0.00 & -0.11 & \color{blue}{\textbf{0.01}} & 0.00 & 0.00 & -0.00 & -0.02 & -0.01 & -0.01 & -0.01 & -0.00 & -0.00 & -0.00 & -0.00 & 0.00 & -0.00 & 0.00 & 0.00 & -0.04 & -0.02 & -0.04 & -0.01 & 0.00 & -0.01 \\ \cmidrule{1-26}

$\text{TPR@5}$ & 5 & 8.1 & 18.3 & 7.0 & 6.7 & 6.6 & 11.4 & 15.4 & 12.9 & 9.1 & 15.4 & 11.0 & \color{red}{\textbf{4.8}} & 14.0 & 8.0 & 7.6 & 11.4 & 10.0 & 10.4 & 17.2 & 22.2 & 18.4 & 12.8 & 10.7 & 8.7 \\
&  & -0.01 & -0.07 & \color{blue}{\textbf{-0.00}} & -0.01 & -0.01 & \color{blue}{\textbf{-0.00}} & -0.03 & -0.01 & -0.01 & -0.02 & -0.01 & -0.01 & -0.02 & -0.01 & -0.01 & -0.01 & -0.01 & -0.01 & -0.04 & -0.17 & -0.06 & -0.03 & -0.02 & -0.04 \\
&  10 & 7.8 & 19.2 & 7.5 & 6.7 & 6.3 & 11.0 & 15.3 & 13.1 & 9.2 & 15.4 & 11.0 & \color{red}{\textbf{4.7}} & 13.9 & 7.8 & 8.0 & 11.0 & 9.8 & 10.0 & 18.4 & 22.4 & 18.6 & 12.8 & 10.3 & 8.9 \\
&  & -0.01 & -0.14 & \color{blue}{\textbf{-0.00}} & -0.01 & -0.01 & \color{blue}{\textbf{-0.00}} & -0.03 & -0.01 & -0.02 & -0.03 & -0.01 & -0.02 & -0.02 & -0.01 & -0.01 & -0.01 & -0.02 & -0.02 & -0.12 & -0.24 & -0.08 & -0.03 & -0.02 & -0.04 \\ 
\bottomrule
\end{tabular}
        \vspace*{0.05cm}
        \caption{tabular}
    \end{subtable}
    \begin{subtable}{\linewidth}
        \centering
        \footnotesize
        \tabcolsep=0.05cm
        \vspace*{0.05cm}
        \begin{tabular}{llrrrrrrrrrrrr}
\toprule
\textbf{criterion} & \textbf{size} & \textbf{aae} & \textbf{gano} & \textbf{skip} & \textbf{vae} & \textbf{wae} & \textbf{knn} & \textbf{osvm} & \textbf{fano} & \textbf{fmgn} & \textbf{dsvd} & \textbf{vaek} & \textbf{vaeo} \\\midrule
$\text{AUC}$ & 5 & 3.5 & 5.0 & 9.1 & 3.2 & \color{red}{\textbf{3.1}} & 5.4 & 7.0 & 3.4 & 7.1 & 4.5 & 5.7 & 6.5 \\
& & \color{blue}{\textbf{-0.00}} & \color{blue}{\textbf{-0.00}} & -0.01 & \color{blue}{\textbf{-0.00}} & \color{blue}{\textbf{-0.00}} & \color{blue}{\textbf{-0.00}} & -0.01 & -0.01 & \color{blue}{\textbf{-0.00}} & -0.03 & \color{blue}{\textbf{-0.00}} & -0.01 \\
& 10 & 3.1 & 4.9 & 8.9 & 3.0 & \color{red}{\textbf{2.9}} & 5.2 & 8.1 & 3.9 & 6.9 & 4.9 & 5.6 & 6.2 \\
& & -0.01 & -0.02 & -0.02 & \color{blue}{\textbf{-0.00}} & \color{blue}{\textbf{-0.00}} & \color{blue}{\textbf{-0.00}} & -0.03 & -0.02 & -0.01 & -0.05 & \color{blue}{\textbf{-0.00}} & -0.02 \\ \cmidrule{1-14}
$\text{TPR@5}$ & 5 & 3.7 & 4.9 & 9.3 & 3.1 & \color{red}{\textbf{2.9}} & 5.2 & 7.2 & 3.8 & 7.2 & 4.8 & 5.4 & 6.1 \\
& & -0.01 & -0.01 & -0.03 & -0.01 & \color{blue}{\textbf{-0.00}} & \color{blue}{\textbf{-0.00}} & -0.03 & -0.01 & -0.01 & -0.03 & -0.01 & -0.01 \\
& 10 & 3.4 & 4.9 & 9.2 & 3.1 & \color{red}{\textbf{2.9}} & 5.2 & 7.9 & 4.2 & 7.1 & 4.8 & 5.4 & 6.4 \\
& & -0.02 & -0.03 & -0.03 & \color{blue}{\textbf{-0.01}} & \color{blue}{\textbf{-0.01}} & \color{blue}{\textbf{-0.01}} & -0.04 & -0.02 & -0.02 & -0.05 & \color{blue}{\textbf{-0.01}} & -0.02 \\ 
\bottomrule
\end{tabular}

        \vspace*{0.05cm}
        \caption{statistic}
    \end{subtable}
    \begin{subtable}{\linewidth}
        \centering
        \footnotesize
        \tabcolsep=0.05cm
        \vspace*{0.05cm}
        \begin{tabular}{llrrrrrrrrrrrr}
\toprule
\textbf{criterion} & \textbf{size} & \textbf{aae} & \textbf{gano} & \textbf{skip} & \textbf{vae} & \textbf{wae} & \textbf{knn} & \textbf{osvm} & \textbf{fano} & \textbf{fmgn} & \textbf{dsvd} & \textbf{vaek} & \textbf{vaeo} \\\midrule
$\text{AUC}$ & 5 & 5.1 & 7.6 & 7.6 & 3.4 & 5.0 & 7.6 & 8.3 & 5.2 & \color{red}{\textbf{2.8}} & 6.3 & 5.3 & 7.4 \\
 & & -0.00 & -0.00 & 0.00 & -0.00 & -0.01 & -0.00 & -0.00 & -0.01 & \color{blue}{\textbf{0.02}} & 0.00 & -0.00 & -0.01 \\
 & 10 & 5.0 & 7.4 & 7.2 & 3.4 & 5.2 & 7.2 & 8.8 & 5.7 & \color{red}{\textbf{2.8}} & 5.6 & 5.4 & 8.4 \\
 & & -0.00 & -0.01 & 0.00 & -0.00 & -0.01 & -0.00 & -0.01 & -0.02 & \color{blue}{\textbf{0.01}} & -0.00 & -0.01 & -0.02 \\ \cmidrule{1-14}
$\text{TPR@5}$ & 5 & 5.6 & 7.8 & 8.1 & 4.5 & 4.8 & 6.7 & 8.6 & 5.6 & \color{red}{\textbf{2.6}} & 5.8 & 5.0 & 8.0 \\
& & -0.01 & -0.01 & -0.01 & -0.02 & -0.01 & -0.00 & -0.02 & -0.03 & \color{blue}{\textbf{0.01}} & -0.00 & -0.01 & -0.03 \\
& 10 & 5.0 & 7.4 & 7.7 & 4.4 & 4.8 & 6.8 & 9.8 & 5.2 & \color{red}{\textbf{2.7}} & 5.2 & 5.0 & 8.4 \\
& & -0.01 & -0.02 & -0.01 & -0.03 & -0.02 & -0.01 & -0.03 & -0.03 & \color{blue}{\textbf{0.01}} & -0.00 & -0.01 & -0.04 \\ 
\bottomrule
\end{tabular}

        \vspace*{0.05cm}
        \caption{semantic}
    \end{subtable}
    \caption{Comparison of the average AUC ranks and average change in the AUC with different ensemble sizes and criterion used for their selection for different types of data. Rows with values in blue show average improvement in AUC from baseline - no ensemble setting.}
    \label{tab:ensembles_sensitivity_grouped}
\end{table*}

\section{Implementation details}
\label{sec:appendix_implementation}
\subsection{Classical methods}
Due to the quadratic scaling of OC-SVM in the number of training samples, we have implemented the OC-SVM on image data as an ensemble of 10 OC-SVM models with the same hyperparameters but trained on 10 equal-sized subsets of training data. The resulting anomaly score of such an ensemble was obtained by averaging.

For the kNN model, we have used 3 anomaly scores that differed at the distance computation, as described in~\citep{harmeling2006outliers}. As oposed to the other methods, implementation of kNN required further downscaling of MVTec-AD images to $64\times64\times3$ in order to fit the computational envelope.

The original PIDForest~\citep{gopalanPIDForestAnomalyDetection2019} method's implementation provided scikit-learn model API, however that's where the convenience ended. We have encountered a number of errors both during training and prediction, some of which we have identified were caused by features with constant values, which we had to filter out from the splits, before proceeding. It's quite unfortunate when, what seems to be a sound method on paper, is handicapped by a subpar implementation, which is both slow and breaks at the first sight of the unknown. Though the authors have promised a better one, as of the time of writing, such an option was not available. As opposed to the original implementation we used negative sparsity as the anomaly score and parametrized its quantiles at three levels $\left\lbrace 0.10, 0.25, 0.50 \right\rbrace$.

\subsection{Autoencoders \& GANs}
The training of VAE, WAE and AAE models has been stopped prematurely if the loss on normal samples in the validation dataset has not improved for 200 batches. For GAN and fmGAN, patience of 50 batches was used and the discriminator loss was used for early stopping. We have not used batch normalization in any of these models, as it is recommended not to do so due to the additional stochasticity that it introduces in the training process.

In the construction of VAE, WAE, AAE, GAN and fmGAN models on image data,  convolution layers are used. First the number of convolution layers $n_c$ was sampled. To construct decoder/generator parts, the first $n_c$ elements of the channels, kernelsizes and scalings vectors in Tab.~\ref{tab:nn_hyperparameters} were used, e.g. for $n_c=2$, a decoder with $(16, 32)$ channels, $(3, 5)$ kernelsizes and $(1, 2)$ scaling was constructed, while the architecture of the encoder/discriminator was flipped. Downscaling in encoders/discriminators was done with maxpooling, upscaling in decoders/discriminators was done via transposed convolutions.

Each of the $k$ components of a VampPrior initialized with an average sample of the training data, perturbed with noise sampled from $\mathcal{N}(0,1)$ in each dimension.

The most challenging implementation was that of adVAE~\cite{wang2020advae} because no training loop or loss functions was provided in the code provided by the author. Additionally, the other implementation recommended by the author differs from the actual paper in terms of loss functions and some hyperparameters. In spite of these uncertainties, we have managed to implement a working version of the model that follows the paper as closely as possible. The only difference is the architecture of the encoder and decoder.

In the case of GANomaly and skip-GANomaly models for image data, we implemented the exact same architectures as in original papers~\cite{akcay2018ganomaly, akcay2019skip} and the only addition to those models is weight decay. This architecture, which is pretty much similar for both models, has one small disadvantage. The width of the input image must be equal to its height and that must be divisible by 32. This allows fully convolution encoders to compress input image into 1D latent vector resp. $(1, 1, dim(\vec{z}))$ tensor. Therefore we have to resize MNIST and FashionMNIST datasets in the preprocess phase from shape 28x28 to 32x32. 

To avoid overfitting of deep models we decided to use an early stopping procedure. The criterion for this procedure most often corresponds to the anomaly score formula or loss function, so it is different for every model. For example GANomaly's early stopping criterion is simple generator loss $L_G = w_1 L_{adversarial} + w_2 L_{contextual} + w_3 L_{latent}$~\citep{akcay2018ganomaly}, however for skip-GANomaly we use weighted average of generator loss $L_G$ and discriminator loss $L_D$ because both those terms are also part of anomaly score. Weights in $L_G$ are typically treated as hyperparameters which we sample. Exception is image version of GANomaly where we used fixed weights $w_1 = 1$, $w_2 = 50$ and  $w_3 = 1$ since those weights were empirically derived in original paper as optimal for the same datasets we use. 

During the implementation of fAnoGAN we were forced to make some adjustments to the original code. Firstly we switched Wasserstein GAN with gradient penalty (WGAN-GP~\cite{gulrajani2017improved}) to its predecessor WGAN with clipping weights~\cite{arjovsky2017wasserstein} because we were not able to compute the second order derivative needed to update weights of the model due to Julia limitations. Later we tried Pytorch implementation of WGAN-GP, but it performed on par with WGAN so we dropped it from results. 

Secondly, we replaced the original ResNet~\cite{he2016deep} generator and discriminator with much lighter architecture (in the same way as with GANomaly), that we also use for autoencoders. This helps to compare models more fairly, as their performance now depends mainly on the model concept, not the architecture itself. 
In addition, fAnoGAN is the only model in this category without an early stopping procedure, because its loss behavior is not standard and we had a hard time figuring out when to stop it.

\subsection{Normalizing flows}
While implementing RealNVP and MAF flows on tabular data we tried to follow as closely as possible the code that has accompanied~\citep{papamakariosMaskedAutoregressiveFlow2018}, however on a closer inspection, we found that there is a lot of variety in the literature, which we tried to include into hyperparameters of each of the methods. This variety comes mainly from the different techniques, that authors~\citep{dinh2016density, grathwohlFFJORDFreeformContinuous2018, kingmaGlowGenerativeFlow2018} employ to battle instabilities in the training of deep flows, though the exact nature of which is often not disclosed. In our case, the root of most problems was in scale conditioner's output saturating (yielding Inf) in the subsequent exponentiation. We have been able to isolate this to specific samples in the training datasets such as Statlog Shuttle, which unfortunately contained outliers with one of the features having disproportional values. The MAF flows seemed to be more affected by this due to their autoregressive structure, which propagated the discrepancy into other features in each subsequent flow.

Using batch normalization as suggested by~\citep{papamakariosMaskedAutoregressiveFlow2018} should supposedly improve both stability and target likelihoods, which we could confirm while experimenting on small problems, however it did not help us with the aforementioned issues. We've found that batch normalization with floating averages as used in~\citep{dinh2016density} performed poorly even on smaller problems, thus we used only statistics computed for each batch individually. While looking deeper into the performance of different hyperparameters, we have found that none of the models on KDD99 (10\%) with batch normalization trained properly yielding AUC's below $0.5$. As per the original implementation, during evaluation the statistics in batch normalization were initialized from the whole training dataset.

As suggested by \citep{dinh2016density} we have also experimented with initializing the flow to identity by zeroing the last layers of conditioner networks, however it also did not help. Furthermore in case of RealNVP flow we have tried to use "tanh normalization" together with learn-able location $\alpha_{s} \in \mathrm{R}$ and scale $\beta_{s} \in \mathrm{R}$ parameters
\begin{equation}
    \alpha_{s} \oplus \exp{\beta_{s}} \odot \tanh{\left(s\right)},
\label{eq:tanh_scaling}
\end{equation}
where $\vec{s}$ is the scale conditioner's output and $\oplus$, $\odot$ are element-wise operations. Using this scaling together with tanh activations, we were able to train even deeper and wider architectures.

One of the degrees of freedom is the choice of conditioner architecture. Where on images the consensus \citep{dinh2016density, kingmaGlowGenerativeFlow2018} seems to be to use one convolution neural network for both location and scale parameters and feeding the network positive and negative input. In the case of tabular flows, where the number of weights is not as closely controlled variable, the additional options are to employ either two networks for both scale and location (RealNVP implementation in\citep{papamakariosMaskedAutoregressiveFlow2018}), or one network whose output is fed through another pair of single-layer networks to provide the two outputs (MAF implementation in\citep{papamakariosMaskedAutoregressiveFlow2018}). We have opted for using two separate networks for both RealNVP/MAF flows to equal the situation.

Even the exact formulation transformation function $f$ from \eqref{eq:rv_transformation} has different forms in the chosen class of autoregressive affine flows~\citep{papamakariosNormalizingFlowsProbabilistic2019}, across different implementations. We have used the following definition of forward step $f^{-1}$ for RealNVP
\begin{align}
    \vec{z}_{\leq d} &= \vec{x}_{\leq d}, \nonumber \\
    \vec{z}_{>d} &= \exp{\left(-\frac{1}{2}\vec{s}\right)} \odot \left(\vec{t} - \vec{x}_{\leq d} \right),
\label{eq:rnvp_forward}
\end{align}
where $\vec{s}$, $\vec{t}$ are outputs of scale and location conditioners respectively when given the first half of the input $\vec{x}_{\leq d} = \vec{z}_{\leq d}$. In case of MAF flow the forward step had the following form
\begin{align}
    z_{i} &= \exp{\left(-\frac{1}{2} s_i \right)} \left( t_i - x_i \right), \;\; \forall i \in \left\lbrace 1,\dots, D, \right\rbrace
\label{eq:maf_forward}
\end{align}
where $s_i$, $t_i$ are the $i$-th elements of outputs of the autoregressive conditioners (MADE networks~\citep{germainMADEMaskedAutoencoder2015}) for scale and location parameters respectively, when given the whole input vector $\vec{x}$. It is important to stress the halving of the outputs of the scale conditioners, as this is another means with which to battle the aforementioned instabilities.

Lastly, we have used the same training loop for both flows with early stopping and patience at $200$, checking at every iteration, same as in~\citep{papamakariosMaskedAutoregressiveFlow2018}.

Adopting the code of SPTN flows was easier, as it has been already written in Julia and the only changes we have made were in the training loop. We have used early-stopping, though due to the high cost of evaluation of validation loss, we have opted for patience $20$ and checking interval $10$. Note also that on higher-dimensional datasets --- HAR, Letter Recognition, Isolet, Multiple Features, Arrhythmia --- the implementation did not allow us to compute all $5$ repetitions and therefore only the minimum of $3$ has been completed.

\subsection{Other reimplemented methods}
As the original article on the Deep Autoencoding Gaussian Mixture \citep{zong2018deep} model (DAGMM for short), did not contain any link to a reference implementation we have been forced to follow alternative ones. We have found in total 4 similar attempts at replicating results of the original article, all of which tailored specifically to the KDDCUP99 dataset. With our implementation we bring more flexibility by implementing encoder/decoder with the same architecture as the previously described variational autoencoders on tabular data, while having much lower latent dimension $\left\lbrace 1, 2 \right\rbrace$. This is crucial for this method's convergence as the latent representation together with cosine similarity and $L_2$ norm is used to fit parameters of a Gaussian mixture, whose covariant matrices have higher chance to degenerate to non positive-semi-definitive matrices, when the latent dimension is higher than $1$ on some datasets. There are two ways with which the authors and other practitioners tried to solve this problem: first is to penalize the inverse of diagonal entries (with the weight $\lambda_1$ of the term being on of the hyperparameters) and the second is simple addition of small $\epsilon$ to the diagonal. Each of these remedies may in part be the cause of overall poor performance of the model, because penalizing the diagonal terms leads to mixtures with high variance and thus potentially less sensitive to the detection task. Adding one size fits all $\epsilon$ is in this case non-systematic solution to an underlying ill posed application of this methods on particular data. We have been able to attain the performance in the original article, but the results with different models seed have high variance as has been reported in the other implementations. One other notable difference is that instead of Cholesky decomposition, we have used LU decomposition to compute covariant matrix's determinant and it's derivative, as it is the default in Julia's Flux/Zygote library. We have employed early stopping based on the training loss evaluated on normal validation data for with patience of 200 iterations. Though we have seen an attempt to extend this method to image dataset, we have not pursued this direction because the problems with convergence with higher latent dimensions of the underlying autoencoder goes directly against the requirements for autoencoders on higher dimensional image data.

Situation with the REPEN \citep{pangLearningRepresentationsUltrahighdimensional2018}) method, have been again easier as there exists original implementation, which we converted to Julia, while keeping the variable names and the structure itself almost intact. As with other methods, we took the liberty of extending the architecture of the input space projection to include two-layered neural networks with either relu or tanh non-linearity, which were mentioned in the article but not implemented. For its training we used fixed number of $10000$ iteration of ADADelta optimizer. During prediction on test and validation data the ensemble of 1NN detector has been fitted with random subsamples of the whole training set, but with fixed seed such that the results are reproducible. This procedure has not been mentioned in the original implementation, in which training set coincided with the test set. We believe that this method would work better on contaminated training data, as the initial sorting mechanism relies on the fact that one may meaningfully split the training samples into normal data and potentially anomalous.

\subsection{Evaluation}
Though we have been meticulous in the implementation, due to the scale of operation we could not cover all the edge cases and therefore some runs did not complete at all or produced corrupted results with invalid values. When a method produced scores with more than half of the values being NaN, such results were filtered out. Overall less than $0.5\%$ ($\sim11k$) of results from the tabular experiments have been affected, out of which the majority ($\sim10k$) have come from our kNN implementation on small dimensional datasets, where only low values of $k$ produced a valid result. Results on the image datasets did not contain any invalid values.

As a result, the experimental files for some combinations of hyperparameter-seed-model were missing and had to be handled in the later stages of evaluation. In our protocol we required at least 3 repetitions of each hyperparameter where repeated experiment were computed.

\subsection{Bayesian optimization}
\label{sec:appendix_bayes_impl}
There are two parts to the implementation of Bayesian optimization:
\begin{itemize}
    \item optimization itself and construction of parameter ranges,
    \item training orchestration.
\end{itemize}
The first part was mostly handled inside Python's scikit-optimise, however the second part has proven to be quite the challenge as we have been working already with most of the result precomputed. In spite of this we believe that we have implemented quite a clean solution that is based on binary files, in which we have stored all the previously sampled/trained hyperparameters and the objective value. Each run of an experiment loaded this file, which we called Bayesian cache and fit the process from beginning with fixed seed, such that the results are reproducible. One important detail is that if an experiment did not finish, we have kept an empty entry in the cache with objective value $0.0$, such that the underlying Gaussian process is given some sort of feedback. This solution allowed us to avoid long running jobs, which in some cases would take up the full $50$ days or end prematurely without any mean of continuation, unless meticulously handled.

\end{document}